\documentclass[twoside]{article}

\usepackage[accepted]{aistats2021}

\usepackage[round]{natbib}

\usepackage[utf8]{inputenc} 
\usepackage[T1]{fontenc}    

\usepackage{url}           
\usepackage{booktabs}     
\usepackage{amsfonts, amsmath, amssymb}      
\usepackage{nicefrac}      
\usepackage{microtype} 
\usepackage{hyperref}
\usepackage{subcaption}
\usepackage{xcolor}
\usepackage{multirow}
\usepackage{soul}

\definecolor{linkcolor}{RGB}{0,128,255}
\usepackage{hyperref}
\hypersetup{
    colorlinks=true,
    allcolors=linkcolor
}

\definecolor{dark-blue}{rgb}{0.15,0.15,0.4}

\hypersetup{
  colorlinks=true,
  linkcolor=dark-blue,
  citecolor=dark-blue,
  urlcolor=black,
}

\usepackage{graphicx}

\usepackage{subcaption}

\usepackage{bm}

\usepackage[ruled]{algorithm2e}
\usepackage{wrapfig}

\usepackage{multicol}

\usepackage{tikz}

\usepackage{xr}
\makeatletter
\newcommand*{\addFileDependency}[1]{
  \typeout{(#1)}
  \@addtofilelist{#1}
  \IfFileExists{#1}{}{\typeout{No file #1.}}
}
\makeatother

\newcommand*{\myexternaldocument}[1]{
    \externaldocument{#1}
    \addFileDependency{#1.tex}
    \addFileDependency{#1.aux}
}

\definecolor{britishracinggreen}{rgb}{0.0, 0.26, 0.15}
\definecolor{nyupurple}{rgb}{0.34, 0.02, 0.54}
 
\newcommand{\diag}{\text{diag}}

\newcommand{\cache}[1]{\textcolor{blue}{#1}}
\renewcommand{\vec}[1]{\mathbf{#1}}

\myexternaldocument{supplement}

\begin{document}

\runningauthor{Stanton, Maddox, Delbridge, Wilson}

\twocolumn[

\aistatstitle{Kernel Interpolation for Scalable Online Gaussian Processes}

\aistatsauthor{
Samuel Stanton$^{1,*}$ \\
\texttt{ss13641@nyu.edu} 
\And Wesley J. Maddox$^{1,*}$ \\
\texttt{wjm363@nyu.edu} 
\And Ian Delbridge$^2$ \\
\texttt{iad35@cornell.edu} 
\And Andrew Gordon Wilson$^{1}$ \\
\texttt{andrewgw@cims.nyu.edu}
}

\aistatsaddress{$^1$New York University, $^2$Cornell University \\ $^*$ Equal contribution.} ]

\begin{abstract}
	Gaussian processes (GPs) provide a gold standard for performance in online settings, such as sample-efficient control and black box optimization,
	where we need to update a posterior distribution as we acquire data in a sequential fashion.  However, updating a GP posterior to accommodate even a single new observation after having observed $n$ points incurs at least $\mathcal{O}(n)$ computations in the exact setting. We show how to use structured kernel interpolation to efficiently recycle computations for constant-time $\mathcal{O}(1)$ online updates with respect to the number of points $n$, while retaining exact inference. We demonstrate the promise of our approach in a range of online regression and classification settings, Bayesian optimization, and active sampling to reduce error in malaria incidence forecasting. Code is available at \url{https://github.com/wjmaddox/online_gp}.
\end{abstract}

\section{INTRODUCTION}

The ability to repeatedly adapt to new information is a defining feature of intelligent agents. Indeed, these \textit{online} or \textit{streaming} settings, 
where we observe data in an incremental fashion, are ubiquitous --- from real-time adaptation in robotics \citep{nguyen-tuong_local_2008} 
to click-through rate predictions for ads \citep{liu2017pbodl}. 

Bayesian inference is naturally suited to the online setting, where after each new observation, an old posterior becomes a new prior. 
However, these updates can be prohibitively slow. For Gaussian processes, if we have already observed $n$ data points, observing
even a single new point requires introducing a new row and column into an $n \times n$ covariance matrix, which can incur $\mathcal{O}(n^2)$ operations for the predictive distribution and $\mathcal{O}(n^3)$ operations for kernel hyperparameter updates.

Since Gaussian processes are now frequently applied in online settings, such as Bayesian
optimization \citep{yamashita2018crystal, letham2019constrained}, or model-based robotics \citep{xu2014gp, mukadam2016gaussian}, this scaling is particularly problematic. Moreover, despite the growing need for scalable online inference, recent research on this topic is scarce.

Existing work has typically focused on data sparsification schemes paired with low-rank kernel updates \citep[e.g.,][]{nguyen-tuong_local_2008}, 
or sparse variational posterior approximations \citep{cheng_incremental_2016,bui_streaming_2017}. Low-rank 
kernel updates are sensible but still costly, and data-sparsification can incur significant error. Variational approaches,
while promising, can provide miscalibrated uncertainty representations compared to exact inference \citep{jankowiak2019parametric, lazaro2009inter, bauer2016understanding}, and often involve careful tuning of many hyperparameters. 
In the online setting, these limitations are especially acute. Uncertainty representation can be particularly crucial for determining the balance of exploration and exploitation in
choosing new query points. Moreover, while tuning of hyperparameters and manual intervention may be feasible for a fixed dataset, it can become particularly burdensome
in the online setting if it must be repeated after we observe each new point.

\begin{figure*}[t!]
	\centering
	\begin{minipage}{\textwidth}
		\centering
		\includegraphics[width=0.3\textwidth]{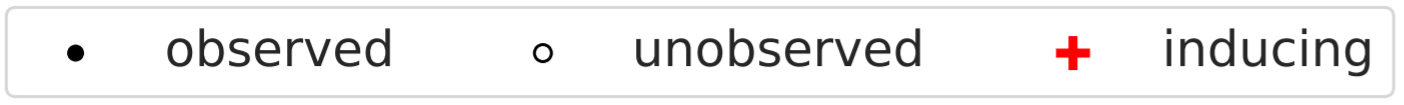}
	\end{minipage}
	\begin{subfigure}{0.49\textwidth}
		\centering
		\includegraphics[width=\textwidth]{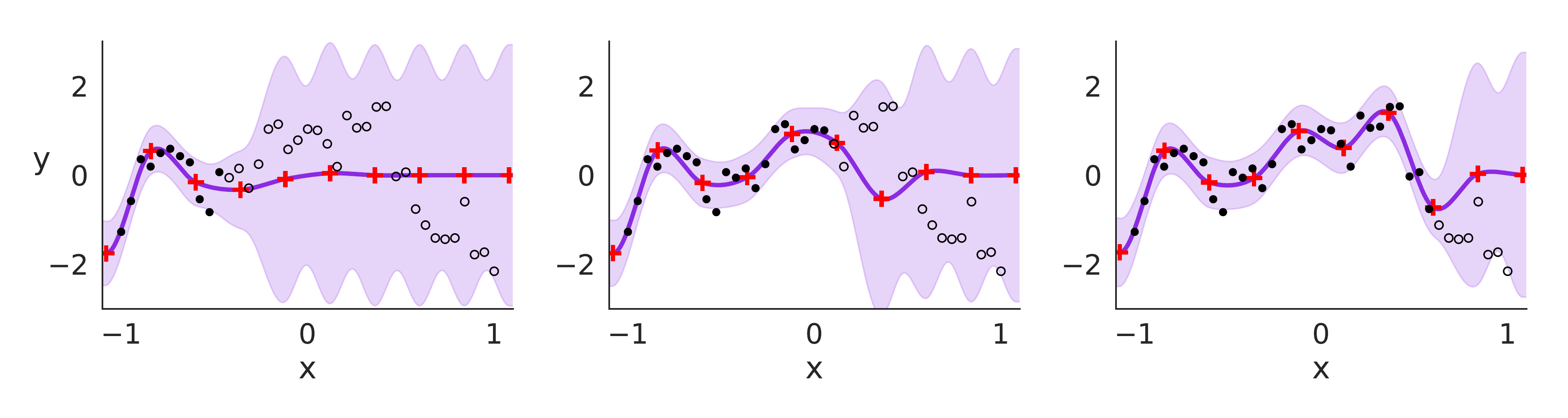}
		\caption{WISKI, time-ordered observations}
		\label{main:fig:wiski_stock_price_non_iid}
	\end{subfigure}
	\hfill
	\begin{subfigure}{0.49\textwidth}
		\centering
		\includegraphics[width=\textwidth]{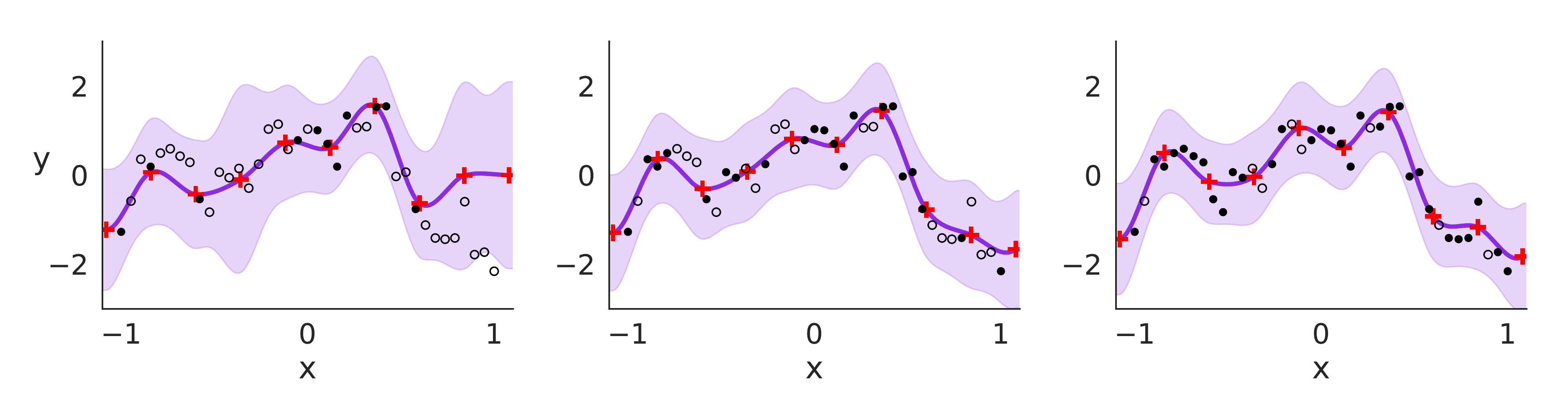}
		\caption{WISKI, randomly-ordered observations}
		\label{main:fig:wiski_stock_price_iid}
	\end{subfigure}
	\hfill
	\begin{subfigure}{0.49\textwidth}
		\centering
        \includegraphics[width=\textwidth]{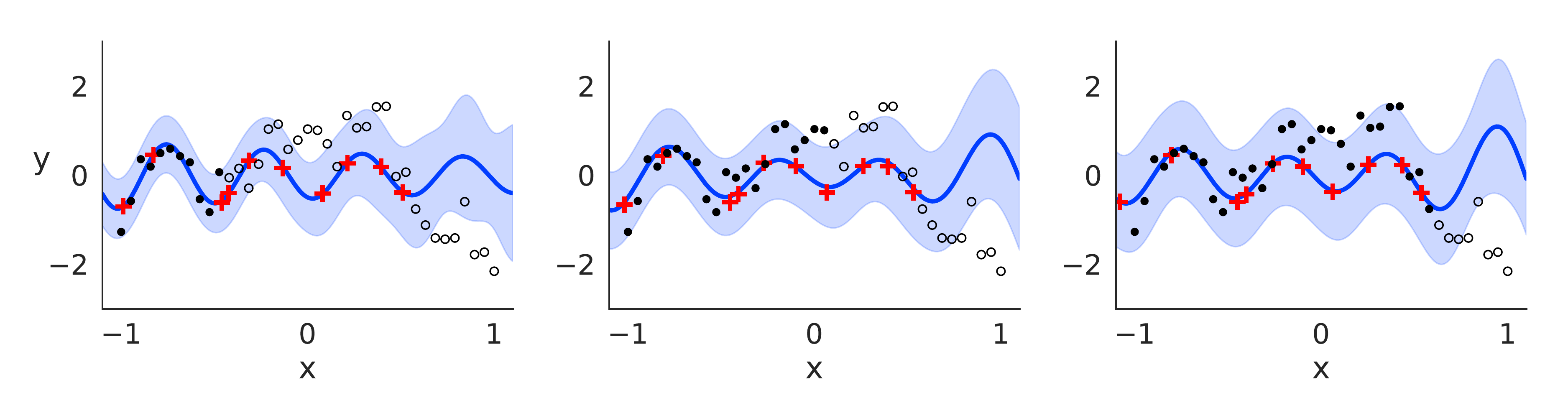}
		\caption{O-SVGP, time-ordered observations}
		\label{main:fig:svgp_stock_price_non_iid}
	\end{subfigure}
	\hfill 
	\begin{subfigure}{0.49\textwidth}
		\centering
        \includegraphics[width=\textwidth]{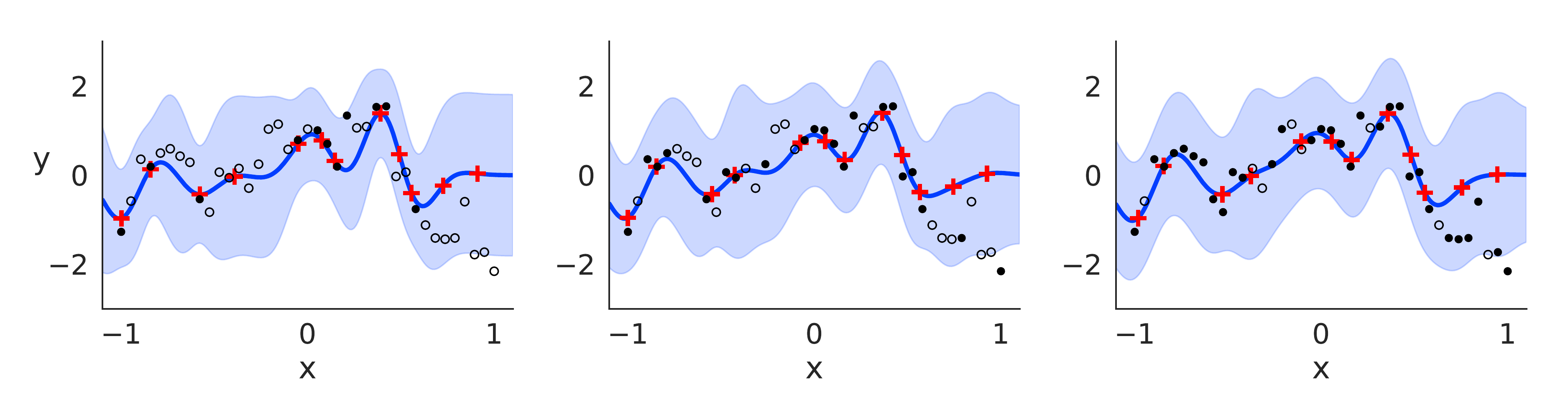}
		\caption{O-SVGP, randomly-ordered observations}
		\label{main:fig:svgp_stock_price_iid}
	\end{subfigure}

	\begin{subfigure}{0.49\textwidth}
		\centering
        \includegraphics[width=\textwidth]{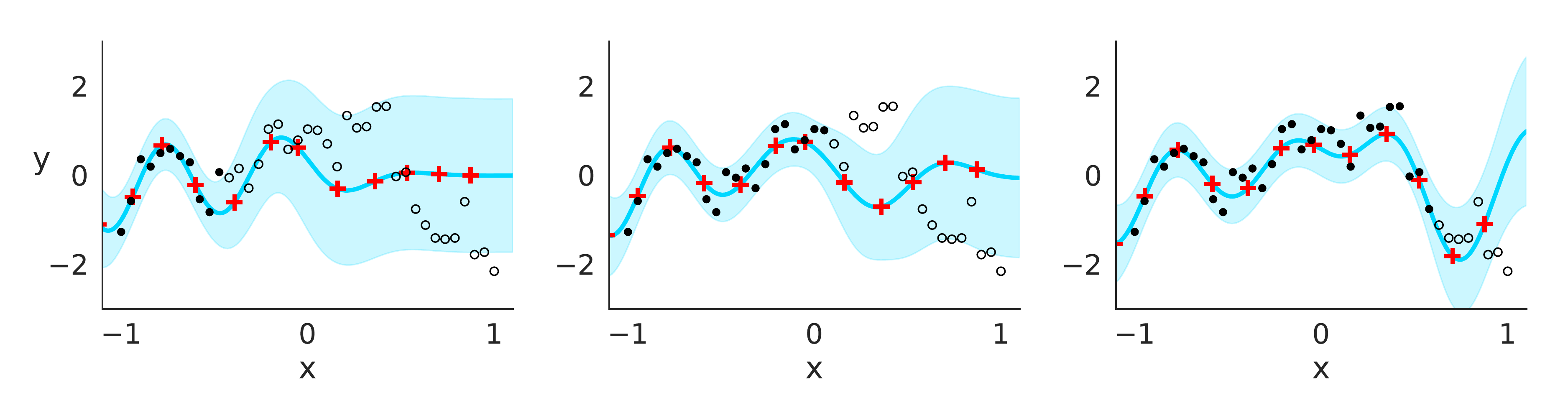}
		\caption{O-SGPR, time-ordered observations}
		\label{app:fig:sgprnonrs_stock_price_non_iid}
	\end{subfigure}
	\begin{subfigure}{0.49\textwidth}
		\centering
        \includegraphics[width=\textwidth]{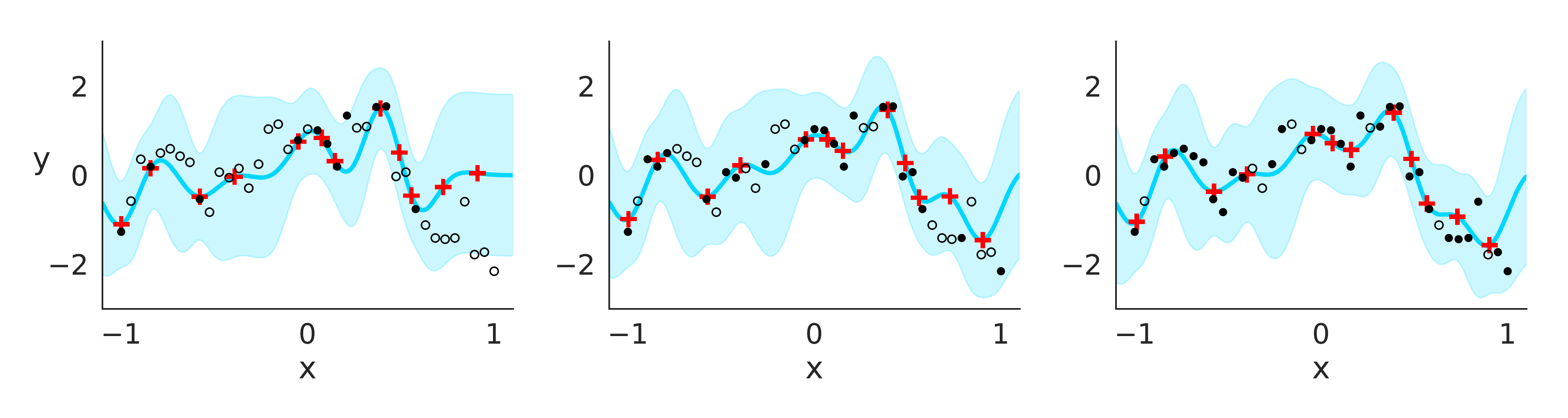}
		\caption{O-SGPR, randomly-ordered observations}
		\label{app:fig:sgpr_stock_price_iid}
	\end{subfigure}

	\caption{Online GP regression on exchange rate time series data ($N=40$). The shaded regions in each panel corresponds to a 95\% credible interval. In each subplot, the left subpanel shows the predictive distribution of the corresponding model after training in batch on an initial set of $10$ observations. The middle and right subpanels show the evolution of the predictive distribution after 10 and 20 online updates, respectively. 
	The \textbf{left} plots, \textbf{(a,c,e)}, show WISKI, O-SVGP, and O-SGPR using spectral mixture kernels \citep{wilson2013gaussian} trained on observations in a time-ordered fashion. O-SVGP heavily overfits to the initial data by interpolating the first batch of data points, and struggles to recover on the next batches. WISKI and O-SGPR perform well in this situation by picking up the signal on the first batches and updating the mean as the data comes in. The \textbf{right} plots, \textbf{(b,d,f)} show the methods trained on observations in a randomly ordered fashion. Here, O-SVGP is still very under-confident, while O-SGPR clumps its inducing points in the middle of the data. By comparison, WISKI learns more of the high frequency trend than either variational approach.
	}
	\label{main:fig:online_regression_updates}
\end{figure*}

Intuitively, we ought to be able to recycle computations to efficiently update our predictive distribution after observing an additional point, rather than 
starting training anew on $n+1$ points. However, it is extremely challenging to realize this intuition in practice, for if we observe a 
new point, we must compute its interaction with every previous point. In this paper, we show it is in fact possible to 
perform constant-time $\mathcal{O}(1)$ updates in $n$, and $\mathcal{O}(m^2)$ for $m$ inducing points, to the Gaussian process predictive distribution,
marginal likelihood, and its gradients, \emph{while retaining exact inference}. We achieve this scaling through 
a careful combination of caching, structured kernel interpolation (SKI) \citep{wilson_kernel_2015}, and reformulations involving the Woodbury identity.
We name our approach 
\textbf{W}oodbury \textbf{I}nversion with \textbf{SKI} (WISKI). We find that WISKI achieves promising results across a range of
online regression and classification problems, Bayesian optimization, and an active sampling problem for estimating malaria incidence
where fast online updates, exact inference for calibrated uncertainty, and fast test-time predictions are particularly crucial. 

\begin{figure}[t]
	\begin{subfigure}{0.23\textwidth}
		\centering
		\includegraphics[width=1.1\textwidth,clip,clip,trim=0cm 4cm 0cm 4cm]{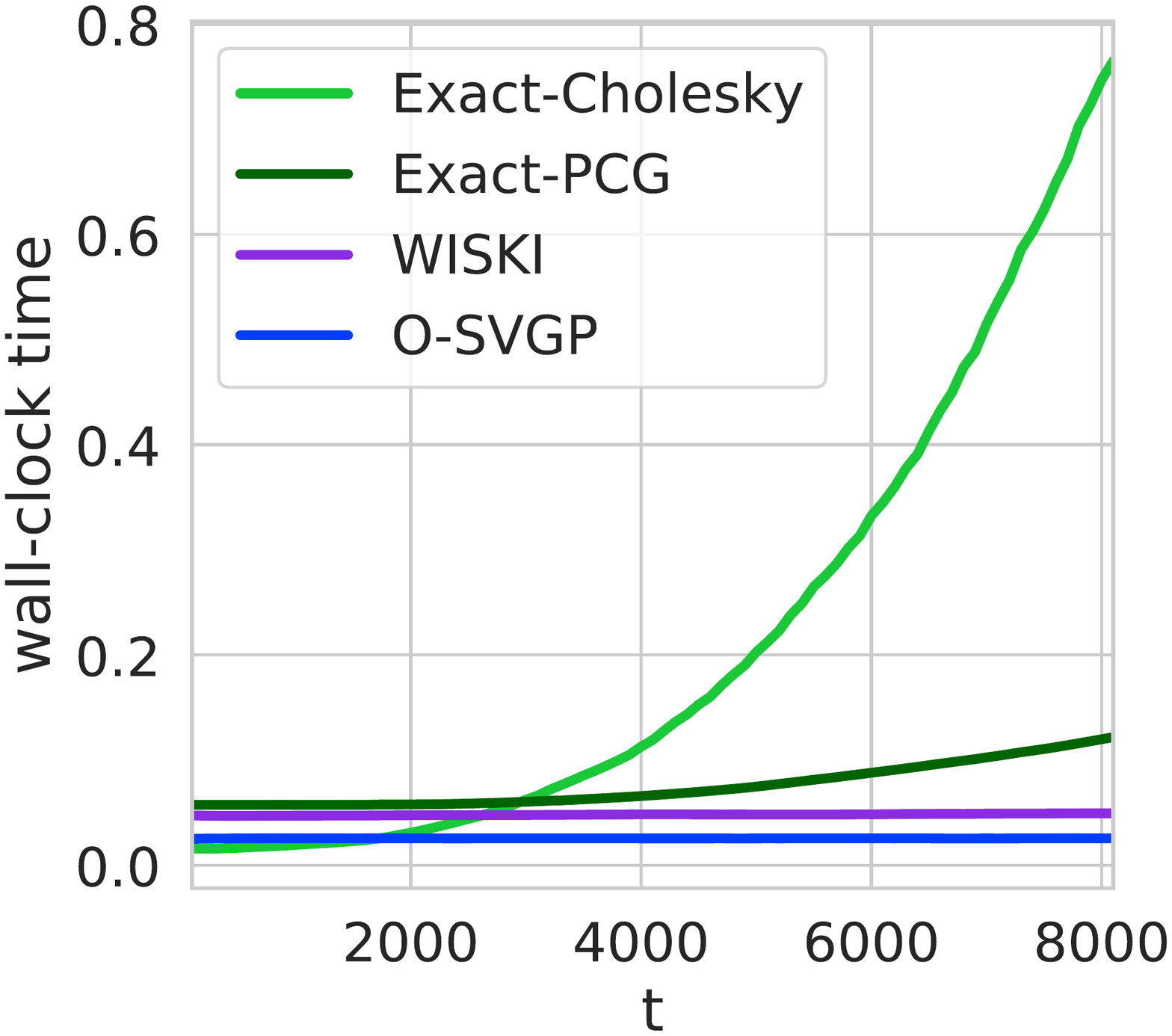}
	\end{subfigure}
	\hfill
	\begin{subfigure}{0.23\textwidth}
		\centering
		\includegraphics[width=1.1\textwidth,clip,clip,trim=0cm 4cm 0cm 4cm]{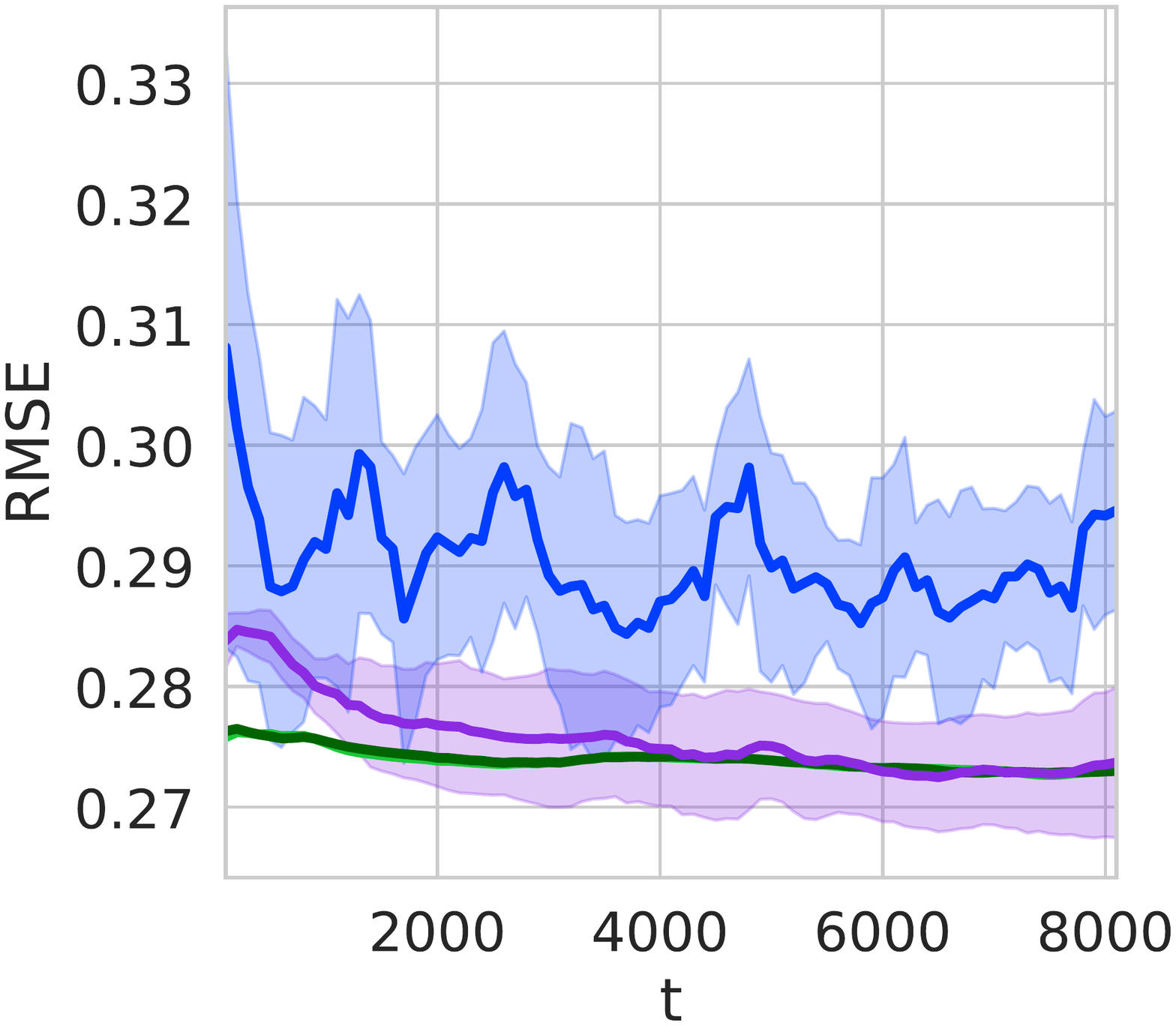}
	\end{subfigure}
	\caption{
	\textbf{Left:} Incorporating new observations becomes increasingly expensive for exact GPs (Exact-Cholesky), even when preconditioned conjugate gradients (Exact-PCG), as quantified in the left panel by the wall-clock time per iteration on the UCI Powerplant dataset. Variational GPs (O-SVGP) are an economical alternative by virtue of being constant time. WISKI has the constant-time profile of a variational method, but retains exact inference, is simple to train, and does not underfit. \textbf{Right:} RMSE on the UCI power plant dataset. Shown are mean and two standard deviations over $10$ trials. O-SVGP tends to overestimate noise and converges to a sub-optimal solution, while WISKI matches the performance of the exact methods trained in an incremental fashion.}
	\label{main:fig:powerplant_timing_comparison}
\end{figure}

As a motivating example, in Figure \ref{main:fig:online_regression_updates}, we fit GPs with spectral mixture kernels \citep{wilson2013gaussian} on British pound to USD foreign exchange data.\footnote{\url{https://raw.githubusercontent.com/trungngv/cogp/master/data/fx/fx2007-processed.csv}, fourth column. We rescaled the inputs to $[-1, 1]$ and standardized the responses.}
In this task, we observe points one at a time, after observing the first $10$ points in batch, and update the predictive distributions for WISKI, O-SVGP and O-SGPR \citep{bui_streaming_2017}, state-of-the-art streaming sparse variational GPs.
We illustrate snapshots after having observed $n=10, 20$, and $30$ points. We see that WISKI is able to more easily capture signal in the data, whereas O-SVGP tends to underfit and O-SGPR underfits on the random data setting. In addition to the general tendency of stochastic variational GP (SVGP) models to underfit the data and overestimate noise variance \citep{lazaro2009inter, bauer2016understanding}, the variational posterior of an O-SVGP is discouraged from adapting to surprising new observations (See Appendix \ref{supp:streamingvi_gps}). 
We also see that O-SVGP particularly struggles when we observe new points in a time-ordered fashion, which is a standard setup in the online setting. 

The initialization heuristics used to train SVGPs in the batch setting, such as initializing the inducing points with $k$-means or freezing the GP hyperparameters at the beginning of training, are not effective for O-SVGPs since the full dataset is not available. In order to obtain reasonable fits with O-SVGP on even this motivating example, we carefully tuned tempering parameters using generalized variational inference \citep{knoblauch2019generalized}, executed $6$ optimization steps for each new observation, and trained in batch on the first $10$ points. WISKI, by contrast, requires no tuning, only $1$ optimization step for each new observation, and does not require any batch training to find reasonable solutions.

These issues with O-SVGP\footnote{O-SGPR has different weaknesses, including numerical instability. We further consider O-SGPR in our larger study of incremental regression.} are particularly visible when we move beyond time series. In Figure \ref{main:fig:powerplant_timing_comparison}, we plot the incremental RMSE on a held out test set on the UCI powerplant dataset, while optimizing for only a single step as we observe new data points, finding that O-SVGP underfits and sub-optimal solution, while WISKI matches the performance of an exact GP also fit incrementally. 
The exact GP also uses pre-conditioned conjugate gradients \citep{gardner2018gpytorch} here.
However, WISKI and O-SVGP are both constant time (shown in the left panel), while using an exact GP with Cholesky factorization is cubic time, and using CG with the GP is quadratic time. 
Both are much slower than WISKI and O-SVGP after $t = 5000.$

\section{RELATED WORK}
\label{main:sec:related_work}

\subsection{Prior Approaches}
Despite its timeliness, there has not been much recent work on online learning with GPs. Older work considers sparse variational approximations to GPs in the streaming setting. \citet{csato2002sparse} proposed a variational sparse GP based algorithm in $\mathcal{O}(nm^2 + m^3)$ time, specifically for deployment in streaming tasks; however, it assumes that the hyperparameters are fixed. \citet{nguyen-tuong_local_2008} proposed local fits to GPs with weightings based on the distance of the test point to the local models. More recently, \citet{koppel_consistent_2019} extended the types of distances used for these types of models while using an iteratively constructed coreset of data points. 
\citet{evans2018scalable} proposed a structured eigenfunction based approach that requires one $\mathcal{O}(n)$ computation of the kernel and uses fixed kernel hyper-parameters but learns interpolation weights.
\citet{cheng_incremental_2016} proposed a variational stochastic functional gradient descent method in incremental setting with the same time complexity; however, like stochastic variational GPs \citep{hensman2013bigdatagp}, \citet{cheng_incremental_2016} assumes the number of data points the model will see is known and set before training begins.
\citet{hoang_unifying_15} proposed a similar variational natural gradient ascent approach, but assumed that the hyper-parameters are fixed during the training procedure, a major limitation for flexible kernel learning.

\subsection{Streaming SVGP and Streaming SGPR}

The current state-of-the-art for streaming Gaussian processes is the sparse variational O-SVGP approach of \citet{bui_streaming_2017} and its ``collapsed'' non-stochastic variant, O-SGPR, which does not use an explicit variational distribution, like its batch equivalent, SGPR \citep{titsias2009variational}. 

\paragraph{O-SVGP: } Unlike its predecessors, O-SVGP is fully compatible with online inference, since it has no requirements to choose the number of data points a priori, and it can update both model parameters and inducing point locations; however, it has the same time complexity as its predecessors: $\mathcal{O}(bm^2 + m^3),$ where $b$ is the size of the batch used to update the predictive distribution and model hyper-parameters. \citet{bui_streaming_2017}'s experiments primarily focused on large batch sizes --- practically $b = \mathcal{O}(n)$ --- rather than the pure streaming setting. 
A major limitation of variational methods in the streaming setting is that conditioning on new observations effectively requires the model parameters to be re-optimized to a minima after every new batch, increasing latency. 
In Appendix \ref{supp:streamingvi_gps} we include a detailed discussion of the requirements of the original O-SVGP algorithm, and provide a modified generalized variational update by downweighting the prior by a factor of $\beta < 1$ better adapted to the streaming setting to have a strong baseline for comparison.
We compare to the generalized O-SVGP implementation in our experiments as O-SVGP.

\paragraph{O-SGPR: } O-SGPR is also potentially promising but like O-SVGP falls prey to several key limitations.
First, O-SGPR relies on analytic marginalization and so can only be used for Gaussian likelihoods.
In Figure \ref{main:fig:online_regression_updates}, we implemented the O-SGPR bound in GPyTorch \citep{gardner2018gpytorch} and it has fair performance for both the random ordering and time ordering settings, though not as good as WISKI. However, this performance comes with two caveats.
First, we need to re-sample the inducing points to include some of the new data at each iteration, as is done in \citet{bui_streaming_2017}'s implementation.
Second, we found that even in double precision we needed to add a large amount of jitter $\epsilon = 0.01,$ while doing the required Cholesky decompositions (there is a matrix subtraction) to prevent numerical instability.

\section{BACKGROUND}\label{main:sec:background}

For a complete treatment on Gaussian Processes, see \citet{williams2006gaussian}. Here we briefly review the key ideas for efficient exact GPs, SKI, and the conditioning of GPs on new observations online. We note SKI provides scalable exact inference through creating an approximate kernel which admits fast computations.

\subsection{Exact GP Regression}
Starting with the regression setting, suppose $\mathbf{y} = f(x) + \varepsilon$, $f \sim \mathcal{GP}(0, k_\theta(x,x'))$, and $\varepsilon \sim \mathcal{N}(0, \sigma^2)$. 
Here, $k_\theta(x,x')$ is the kernel function with hyperparameters $\theta$, and $K_{AB} := k_\theta(A, B)$ is the covariance between two sets of data inputs $A$ and $B$. 
Given training data $\mathcal{D} = (X, \mathbf y)$, we can train the GP hyperparameters by maximizing the marginal log-likelihood,
\begin{align}
	\log p(\mathbf{y} | X, \theta) = &-\frac{1}{2}\mathbf{y}^\top (K_{XX} + \sigma^2 I )^{-1} \mathbf{y} \nonumber\\
	&- \frac{1}{2}\log|K_{XX} + \sigma^2 I | - \frac{n}{2} \log 2\pi. \label{main:eq:marginal_likelihood}
\end{align}
Conventionally, solving the linear system, $(K_{XX} + \sigma^2 I )^{-1} \mathbf{y},$ in Eq. \ref{main:eq:marginal_likelihood} costs $\mathcal{O}(n^3)$ operations. 
The posterior predictive distribution of a new test point $p(f(\mathbf x^*) | \mathbf x^*, \mathcal{D}, \theta) = \mathcal{N}(\mu_{f | \mathcal{D}}, \sigma^2_{f | \mathcal{D}})$, where
\begin{align}
	\mu_{f | \mathcal{D}}(\mathbf x^*) &= K_{\mathbf{x^*} X} (K_{XX} + \sigma^2 I )^{-1} \mathbf y \label{main:eq:exact_pred_mean}, \\
	\sigma^2_{f | \mathcal{D}}(\vec x^*) &= K_{\mathbf{x^*} \mathbf{x^*}} - K_{\mathbf{x^*} X} (K_{XX} + \sigma^2 I )^{-1}K_{X \mathbf{x^*}} \label{main:eq:exact_pred_var}
\end{align}
We build on previous work on scaling GP training and prediction by exploiting kernel structure and efficient GPU matrix vector multiply routines to quickly compute gradients (CG) of Eq. \ref{main:eq:marginal_likelihood} for training, and by caching terms in Eq. \ref{main:eq:exact_pred_mean} and Eq. \ref{main:eq:exact_pred_var} for fast prediction \citep{gardner2018gpytorch}. 
Conjugate gradient methods improve the asymptotic complexity of GP regression and to $\mathcal{O}(jn^2),$ where $j$ is the number of CG steps used.
These recent advances in GP inference have enabled exact GP regression on datasets of up to one million data points in the batch setting \citep{wang_exact_2019}.

\subsection{SKI and Lanczos Variance Estimates}
GPs are often \textit{sparsified} through the introduction of inducing points (also known as pseudo-inputs), which are small subset of fixed points \citep{snelson2006sparse}.
In particular, \citet{wilson_kernel_2015} proposed structured kernel interpolation (SKI) to approximate the kernel matrix as
$K_{XX} \approx \tilde{K}_{XX} = W K_{UU} W^\top,$ where $U$ represents the $m$ inducing points, and $W\in \mathbb{R}^{n \times m}$ is a sparse cubic interpolation matrix composed of $n$ vectors $\vec w_i \in \mathbb{R}^m$.
Each vector $\vec w_i$ is sparse, containing $4^d$ non-zero entries, where $d$ is the dimensionality of the input data.
SKI places the inducing points on a multi-dimensional grid. When $k_\theta$ is stationary and factorizes across dimensions, 
$K_{UU}$ can often be expressed as a Kronecker product of Toeplitz matrices, leading to fast multiplies. Overall multiplies with $\tilde{K}_{XX}$  take $\mathcal{O}(n + g(m))$ time, where $g(m) \approx m$  \citep{wilson_kernel_2015}, compared to the $\mathcal{O}(nm^2 + m^3)$ complexity associated with most inducing point methods \citep{quinonero2005unifying}.
In short, SKI provides scalable exact inference, through introducing an approximate kernel that admits fast computations. 

\citet{pleiss2018constant} propose to cache (i.e. to store in memory) all parts of the predictive mean and covariance that can be computed before prediction, enabling constant time predictive means and covariances. 
Directly substituting the SKI kernel matrix, $\tilde{K}_{XX},$ into Eq. \ref{main:eq:exact_pred_mean}, the predictive mean becomes
\begin{align*}
	\mu_{f | \mathcal{D}}\left(\mathbf{x}^{*}\right)=\vec w_{\mathbf{x}^{*}}^{\top}
	\textcolor{blue}{\underbrace{K_{U U} W^\top \left(W K_{U U} W^\top +\sigma^{2} I\right)^{-1} \mathbf{y}}_{\mathbf{a}}},
\end{align*}
where $\textcolor{blue}{\mathbf{a}}$ is the \emph{predictive mean cache}.\footnote{We refer to entities that can be computed, stored in memory, and used in subsequent computations as \emph{caches}. We use \textcolor{blue}{blue} font to identify which cached expressions.}
Similarly, Eq. \ref{main:eq:exact_pred_var} becomes
\begin{align*}
	\sigma^2_{f | \mathcal{D}}&\left(\mathbf{x}_{i}^{*}, \mathbf{x}_{j}^{*}\right) 
	= k(\mathbf{x}_{i}, \mathbf{x}_{j}^{*}) - \\
	& \vec w_{\mathbf{x}^{\star}}^{\top} 
	\textcolor{blue}{\underbrace{K_{U U} W^\top \left(\tilde{K}_{X X}+\sigma^{2} I\right)^{-1} W K_{U U}}_{C}} \vec w_{\mathbf{x}^{\star}},
\end{align*}
where $\textcolor{blue}{C \approx S S^\top},$ is the \emph{predictive covariance cache}. $\textcolor{blue}{S}$ is formed by computing a rank-$k$ root decomposition of $(\tilde{K}_{X X}+\sigma^{2} I)^{-1} \approx RR^\top$ and taking $\textcolor{blue}{S = K_{UU} W^\top R}$.
The complexity of the root decomposition is $\mathcal{O}(km^2)$, requiring $k \leq m$ iterations of the Lanczos algorithm \citep{lanczos1950iteration} and a subsequent eigendecomposition of the resulting $k \times k$ symmetric tridiagonal matrix.
Further details on Lanczos decomposition and the caching methods of \citep{pleiss2018constant} are in Appendix \ref{supp:sec:derivations}.

\subsection{Online Conditioning and Low-Rank Matrix Updates}
\label{main:subsec:new_observations}
GP models are conditioned on new observations through Gaussian marginalization \citep[][Chapter 2]{williams2006gaussian}.
Suppose we have past observations $\mathcal{D} = \{ (\mathbf x_i, y_i) \}_{i=1}^n$ used to make predictions via $p(y^* | \mathbf x^*, \mathcal{D}, \theta)$. We subsequently observe a new data point $(\mathbf x', y')$. For clarity, let $X = \mathbf x_{1:n}$, $X' = X \cup \{ \mathbf x' \}$. The new kernel matrix is
\begin{align}
	K_{X'X'} = 
	\left(\begin{matrix}
		K_{XX} & k(X, \mathbf x') \\
		k(\mathbf x', X) & k(\mathbf x', \mathbf x')
	\end{matrix}\right)
	\label{main:eq:new_kernel}
\end{align}
We would like to update our posterior predictions to incorporate the new data point without recomputing our caches that are not hyper-parameter dependent from scratch. If the hyperparameters are fixed, this can be a  $\mathcal{O}(n^2)$ 
low-rank update to the predictive covariance matrix (e.g. a Schur complement update or low rank Cholesky update to a decomposition of $(K_{XX} + \sigma^2I)^{-1}$).
If we additionally wish to update hyper-parameters, we must recompute the marginal log likelihood in Eq. \ref{main:eq:marginal_likelihood}, which costs $\mathcal{O}(n^3).$
Similarly, if we naively use the SKI approximations in Eq. \ref{main:eq:new_kernel} we additionally have an $\mathcal{O}(n)$ cost for both adding a new data point and to update the hyper-parameters afterwards.
Thus, as $n$ increases, training and prediction will slow down (Figure \ref{main:fig:powerplant_timing_comparison}).

\section{WISKI: ONLINE CONSTANT TIME SKI UPDATES}\label{main:sec:methods}

We now propose WISKI, which through a careful combination of caching, SKI, and the Woodbury identity, achieves constant time (in $n$) updates in the streaming setting, while retaining exact inference. To begin, we present two key identities that result from the application of the Woodbury matrix identity to the inverse of the updated SKI approximated kernel, $\tilde{K}_{X_t X_t},$ after having received $t$ data points. First, we can rewrite the SKI kernel inverse as
\begin{align}
	(\tilde{K}_{XX} + \sigma^2 I)^{-1} &= \frac{1}{\sigma^2} I -\frac{1}{\sigma^2} W M W^\top. \label{main:eq:ski_inverse} \\
	M :&= (\sigma^2 K^{-1}_{UU} + W^\top W)^{-1}. \label{main:eq:woodbury_inverse}
\end{align}
Second, after observing a new data point at time $t+1$, the inner matrix inverse term $M$ can be updated via a rank-one update, 
\begin{align}
	M^{-1}_{t+1} = M^{-1}_t + \vec w_{t+1} \vec w_{t+1}^\top,
	\label{main:eq:rank1_update}
\end{align}
where $\vec w_{t+1}$ is an interpolation vector for the $t+1$th data point.\footnote{We have dropped the dependence on $\mathbf{x}$ for simplicity of notation.} The exploitation of this rank-one update on a fixed size matrix by storing $W^\top W$ will form the basis of our work. 

Computing Eq. \ref{main:eq:rank1_update} as written requires explicit computation of $K_{UU}^{-1}$. 
In general, $K_{UU}$ will have significant structure as $U$ is a dense grid, which yields fast matrix inversion algorithms; however, the inverse will be very ill-conditioned because many kernel matrices on gridded data have (super-)exponentially decaying eigenvalues \citep{bach2002kernel}. We will instead focus on reformulating SKI into expressions that depend only on $K_{UU}$, $W$, and $\vec y$ with a constant $\mathcal{O}(m^2)$ memory footprint and can be computed in $\mathcal{O}(m^2)$ time.

\subsection{Computing the Marginal Log-Likelihood, Predictive Mean and Predictive Variance}
\label{main:subsec:woodbury_inverse}

Substituting Eq. \eqref{main:eq:ski_inverse} into Eqs. \eqref{main:eq:marginal_likelihood}, \eqref{main:eq:exact_pred_mean}, and \eqref{main:eq:exact_pred_var}, we obtain the following expressions for the marginal log-likelihood (MLL), predictive mean, and predictive variance\footnote{A similar result holds for fixed noise heteroscedastic likelihoods as well. See Appendix \ref{app:het_likelihoods} for further details.}:
\begin{align}
	&\log p(\mathbf{y} | X, \theta)
	=  -\frac{1}{2\sigma^2}(\cache{\mathbf{y}^\top \mathbf{y}} - \cache{\mathbf{y}^\top W} M \cache{W^\top \mathbf{y}}) - \nonumber \\
	&\frac{1}{2} \left(\log{|K_{UU}|} - \log{|M|} + (n - m) \log{\sigma^2}\right), \label{main:eq:intermediate_mll} \\
	&\mu_{f | \mathcal{D}}\left(\mathbf{x}^{*}\right)
	= \vec w_{\mathbf{x}^{*}}^{\top} M \cache{W^\top \mathbf{y}}, \\
	&\sigma^2_{f | \mathcal{D}}\left(\mathbf{x}_{i}^{*}, \mathbf{x}_{j}^{*}\right) 
	= \sigma^2 \vec w_{\mathbf{x_i}^{*}}^{\top} M \vec w_{\mathbf{x_j}^{*}}. \label{main:eq:woodbury_pred_covar}
\end{align}

For all derivations see Appendix \ref{supp:sec:derivations}. We begin by constructing a rank $r$ root decomposition of the matrix $W^\top W \approx \textcolor{blue}{L L^\top}$, along with the factorization of the (pseudo-)inverse, $\textcolor{blue}{J J^\top} \approx (W^\top W)^+$.
The root decomposition $\cache{LL^\top}$ can be a full Cholesky factorization ($r=m$) for relatively small $m$ (i.e. $m 
\leq 1000$) or an approximate Lanczos decomposition for larger $m$, at a one-time cost of $\mathcal{O}(m^2 r)$. Applying the Woodbury matrix identity to Eq. \eqref{main:eq:woodbury_inverse} and substituting $W^\top W \approx \textcolor{blue}{LL^\top}$, we have
\begin{align}
	M &= \sigma^{-2}K_{UU} - \sigma^{-2}K_{UU} \textcolor{blue}{L} Q^{-1} \textcolor{blue}{L^\top} \sigma^{-2}K_{UU}, \label{main:eq:computing-Q-directly} \\
	Q &:= I + \textcolor{blue}{L^\top} \sigma^{-2}K_{UU} \textcolor{blue}{L}.
\end{align}
$Q^{-1}\cache{L^\top}$ is a $r \times r$ system, so directly computing Eq. \eqref{main:eq:computing-Q-directly} requires $\mathcal{O}(r^2 m)$ time for the solve using conjugate gradients, $\mathcal{O}(r m \log m)$ time for the matrix multiplications with $K_{UU}$ if it has Toeplitz structure, and $\mathcal{O}(m^2)$ for the dense matrix additions, and $\mathcal{O}(k m)$ for the root decomposition of $W^\top W$, for a final total of $\mathcal{O}(r^2 m + k m \log m + m^2)$. 
However, we do not \emph{explicitly} store the matrix $M$ as doing so would require $m$ solves of a $r \times r$ system since $\cache{L} \in \mathbb{R}^{m \times r}$.

Eqs. \eqref{main:eq:intermediate_mll} - \eqref{main:eq:woodbury_pred_covar} involve computations of the form 
\begin{align*}
	M \vec v = \sigma^{-2}K_{UU} \vec v - \sigma^{-2}K_{UU} L {Q}^{-1} L \sigma^{-2}K_{UU} \vec v, 
\end{align*}
which can be computed using only a \emph{single} solve against the matrix ${Q}$ via first multiplying out $\vec a = L \sigma^{-2}K_{UU} \vec v,$ and then computing $\vec b = {Q}^{-1} \vec a$.
Applying the matrix determinant identity to $\log |M|$ results in a simplified expression in terms of $\log|Q|$. Taking $\vec v = \cache{W^\top \vec y}$, we obtain a practical expression for the MLL,
\begin{align}
	\log &p(\mathbf{y} | X, \theta) = -\frac{1}{2\sigma^2}\left(\cache{\mathbf{y}^\top \mathbf{y}} - \cache{\mathbf{y}^\top W} K_{UU} \cache{W^\top \mathbf{y}} + \right. \nonumber \\
	&\left. \vec a^\top Q^{-1} \vec a\right)-\frac{1}{2} \left(- \log{|Q|} + (n - m) \log{\sigma^2}\right)\label{main:eq:woodbury_mll}.
\end{align}
Computing $\vec a$ costs $\mathcal{O}(m \log m + rm),$ so computing the two quadratic forms are $\mathcal{O}(m \log m + m)$ and $\mathcal{O}(j r^2)$ respectively, assuming $j$ steps of conjugate gradients.
We use stochastic Lanczos quadrature to compute the log determinant of $|Q|$ which costs $\mathcal{O}(j r^2)$ \citep{gardner2018gpytorch}.
Overall, computation of the MLL becomes $\mathcal{O}(rm + m \log m + jr^2).$

The predictive mean is similarly computed by taking $\vec v = \cache{W^\top y}$, resulting in the expression
\begin{align}
	&\mu_{f | \mathcal{D}}\left(\mathbf{x}^{*}\right)
	= \vec w_{\mathbf{x}^{*}}^{\top} \left( \sigma^{-2}K_{UU}(\cache{W^\top \mathbf{y}} - \cache{L} \vec b) \right).
	\label{main:eq:woodbury_predictive_mean}
\end{align}
The only term that remains is the predictive variance, for which we take $\vec v = \vec w_{\vec x^*_j}$ and obtain
\begin{align}
	\sigma^2_{f | \mathcal{D}}\left(\mathbf{x}_{i}^{*}, \mathbf{x}_{j}^{*}\right) 
	&= \sigma^2 \vec w_{\mathbf{x_i}^{*}}^{\top} (K_{UU}(\vec w_{\vec x^*_j} - \cache{L} \vec b)). \label{main:eq:woodbury_predictive_variance}
\end{align}

\subsection{Conditioning on New Observations}
\label{main:subsec:wiski_conditioning}
When we observe a new data point $(\vec x_{t+1}, y_{t+1})$, we need to update $\cache{(W^\top \vec y)_t}$, $\cache{(\vec y^\top \vec y)_t}$, and $\cache{L_t L_t^\top} = (W^\top W)_t$. The update to the first two terms is simple:
\begin{align}
	\cache{(W^\top y)_{t+1}} &= \cache{(W^\top \vec y)_t} + y_{t+1} \vec w_{\vec x_{t+1}} \\
	\cache{(\vec y^\top \vec y)_{t+1}} &= \cache{(\vec y^\top \vec y)_t} + y_{t+1}^2
\end{align}

We can update $\cache{L_t}$ in $\mathcal{O}(mr + r)$ time by exploiting the rank-one structure of the expression 
\begin{align*}
	(W^\top W)_{t+1} = (W^\top W)_t + \vec w_{\vec x_{t+1}}\vec w_{\vec x_{t+1}}^\top.
\end{align*}
Recalling that $\cache{JJ^\top} = (W^\top W)^+$, let $\vec p = \cache{J_t}^\top \vec w_{\vec x_{t+1}}$. We compute the decomposition $BB^\top = I_r + \vec p \vec p^\top$ and obtain the expression for the updated root $\cache{L_{t+1}} = \cache{L_t} B$. Since $BB^\top$ is a decomposition of $I_r$ plus a rank-one correction, it can be computed in $\mathcal{O}(r)$ time. Since the updates to the first two caches are $\mathcal{O}(1)$ and $\mathcal{O}(m)$, respectively, the total complexity of conditioning on a new observation is $\mathcal{O}(mr^2 + r)$. Further details and an extended proof are given in Appendix \ref{supp:sec:derivations}.

\subsection{Updating Kernel Hyperparameters}
\label{main:subsec:online_hyper_updates}
\begin{figure*}[t]
	\centering
	\begin{subfigure}{0.19\textwidth}
		\includegraphics[width=\linewidth,clip,clip,trim=0cm 4cm 0cm 4cm]{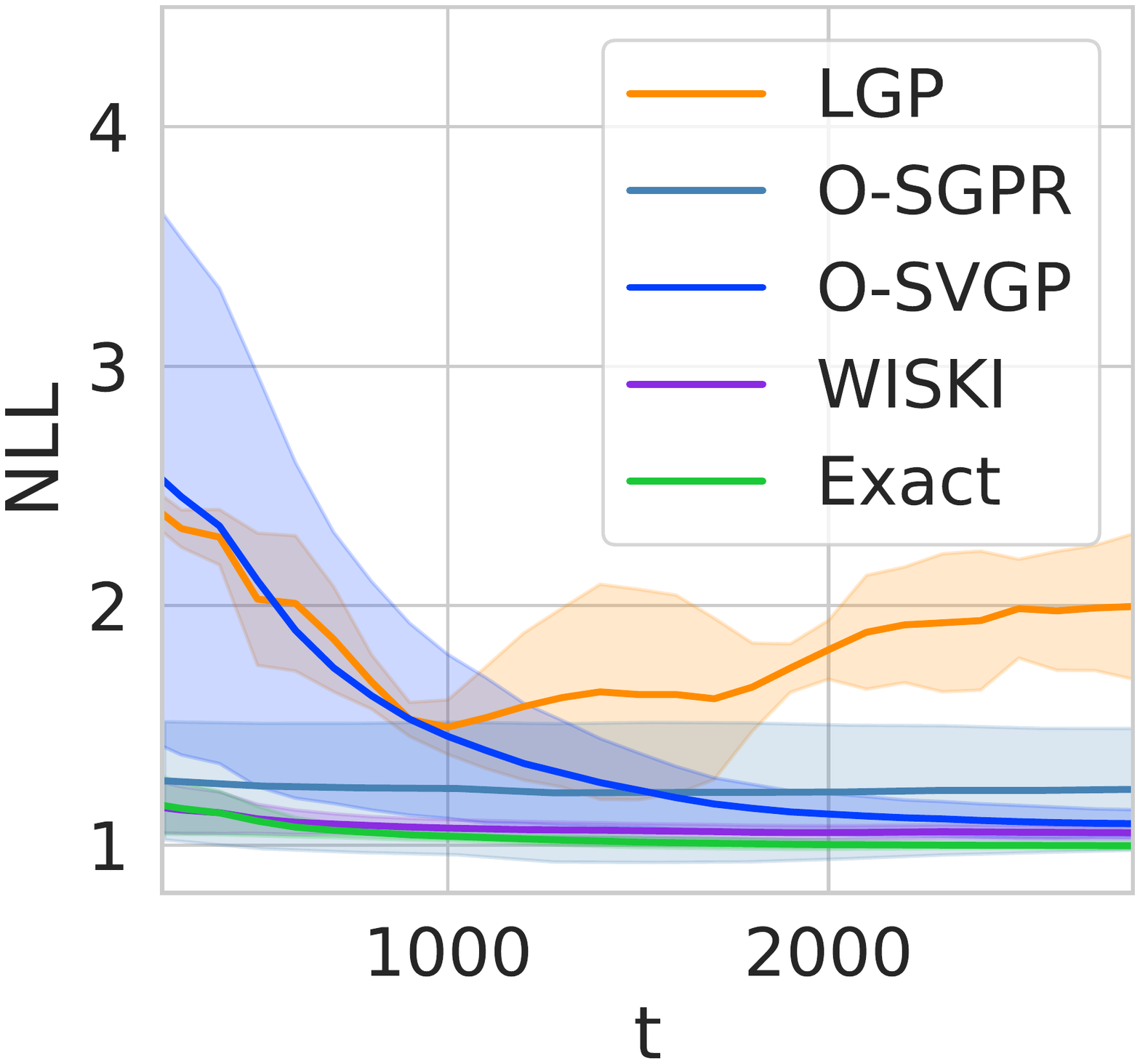}
	\end{subfigure}
	\hfill
	\begin{subfigure}{0.19\textwidth}
		\includegraphics[width=\linewidth,clip,clip,trim=0cm 4cm 0cm 4cm]{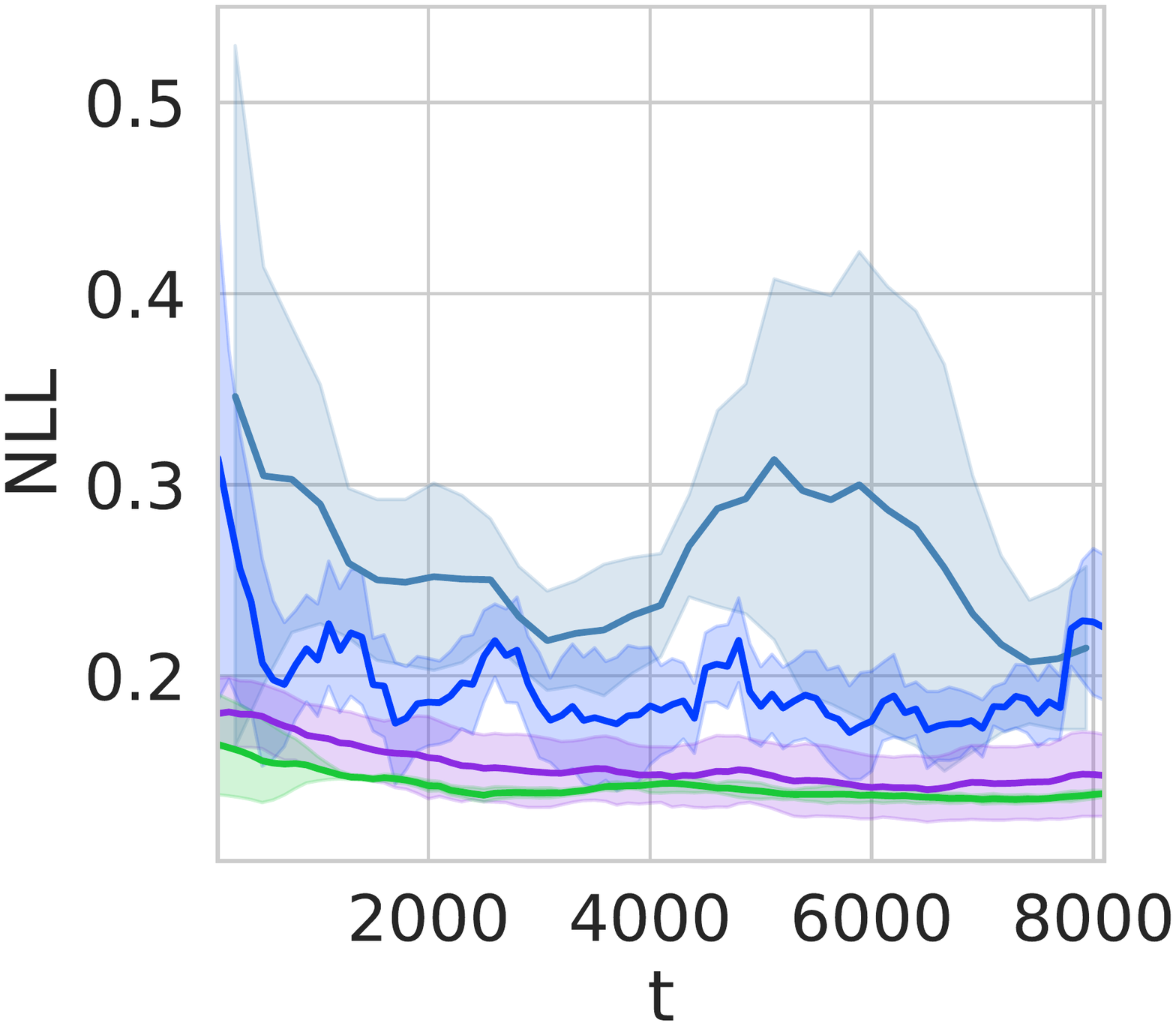}
	\end{subfigure}
	\hfill
	\begin{subfigure}{0.19\textwidth}
		\centering
		\includegraphics[width=\textwidth,clip,clip,trim=0cm 4cm 0cm 4cm]{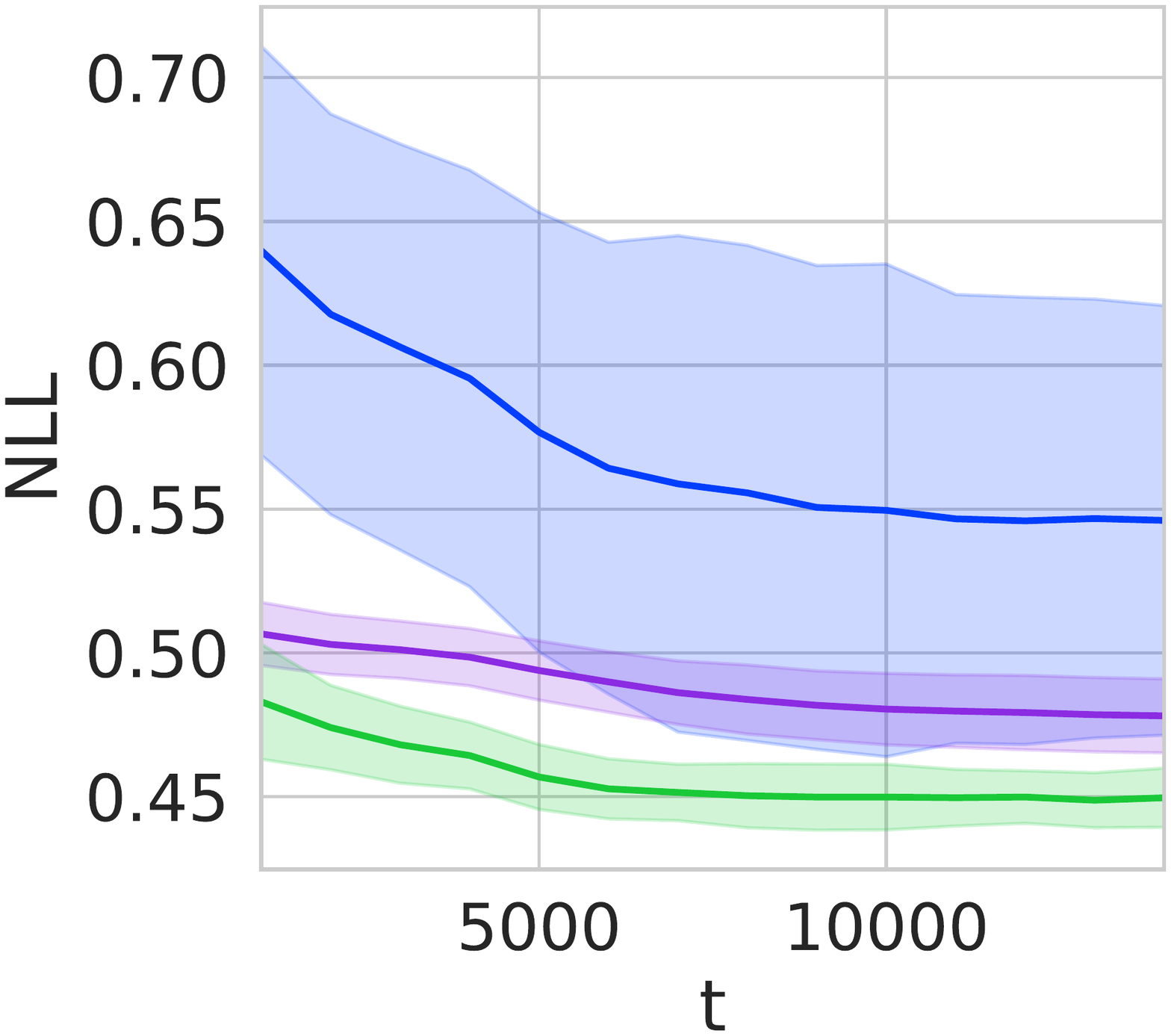}
	\end{subfigure}
	\hfill
	\begin{subfigure}{0.19\textwidth}
		\centering
		\includegraphics[width=\textwidth,clip,clip,trim=0cm 4cm 0cm 4cm]{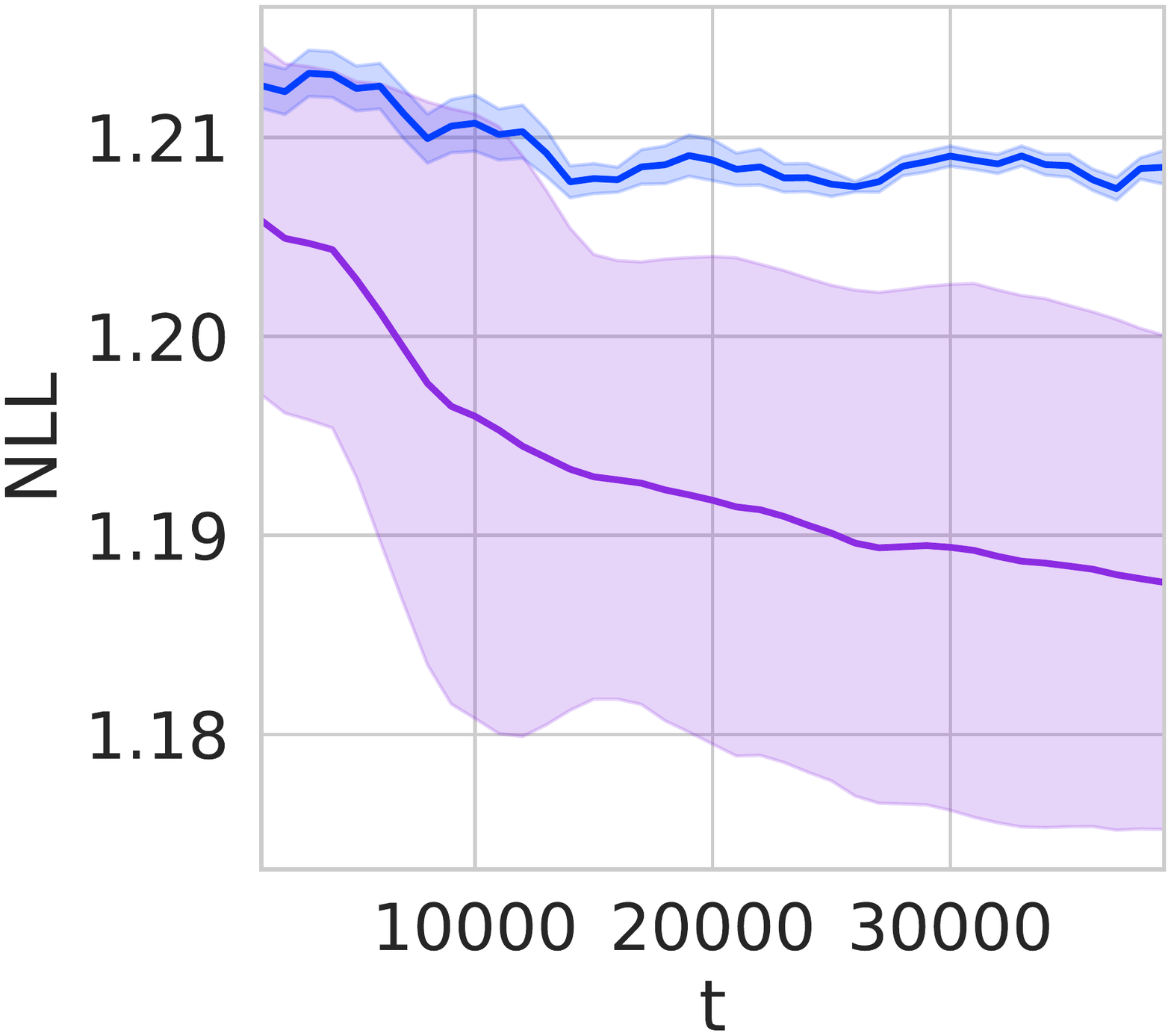}
	\end{subfigure}
	\hfill
	\begin{subfigure}{0.19\textwidth}
		\centering
		\includegraphics[width=\textwidth,clip,clip,trim=0cm 4cm 0cm 4cm]{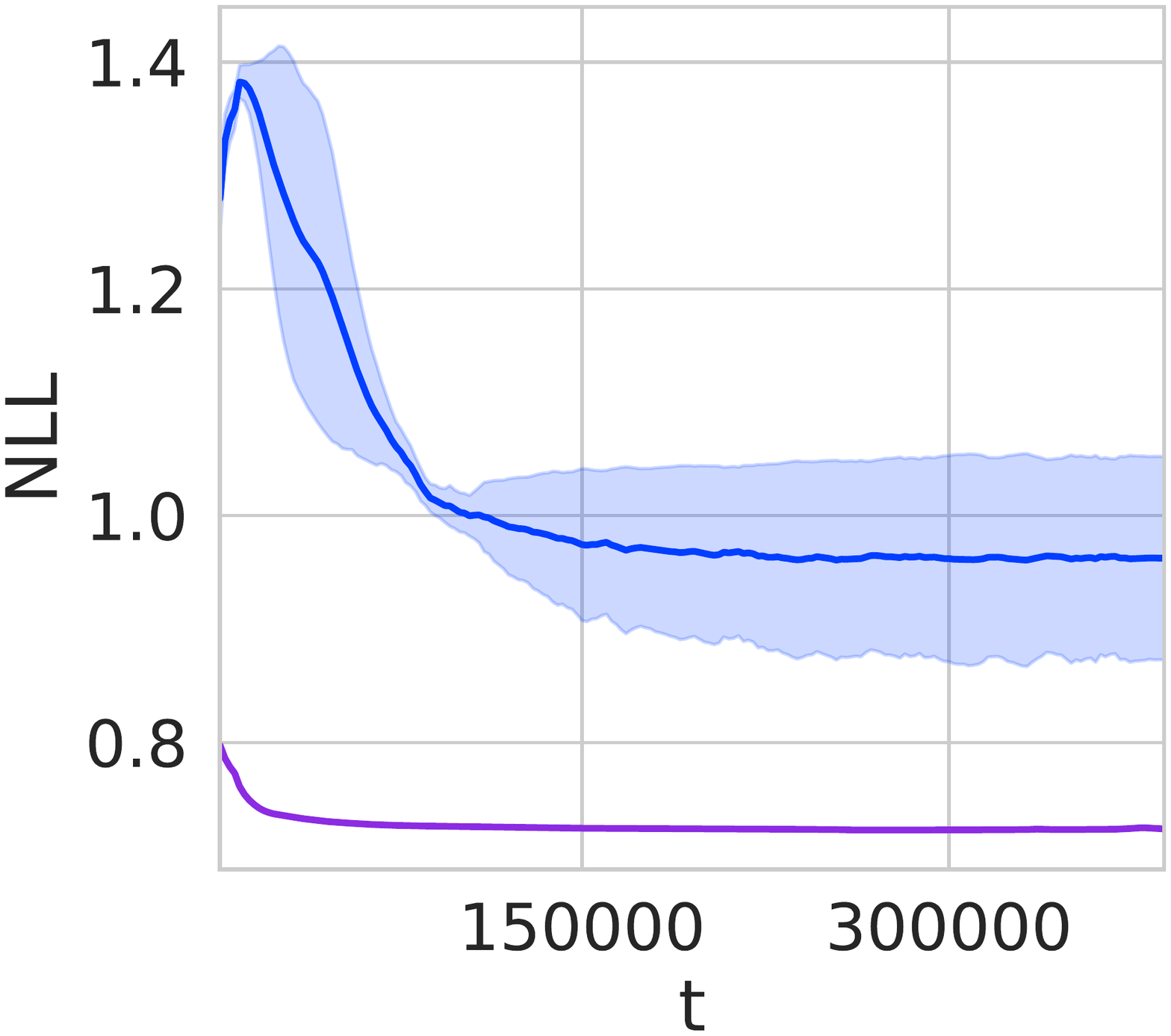}
	\end{subfigure}
	\centering
	\begin{subfigure}{0.19\textwidth}
		\includegraphics[width=\linewidth,clip,clip,trim=0cm 4cm 0cm 4cm]{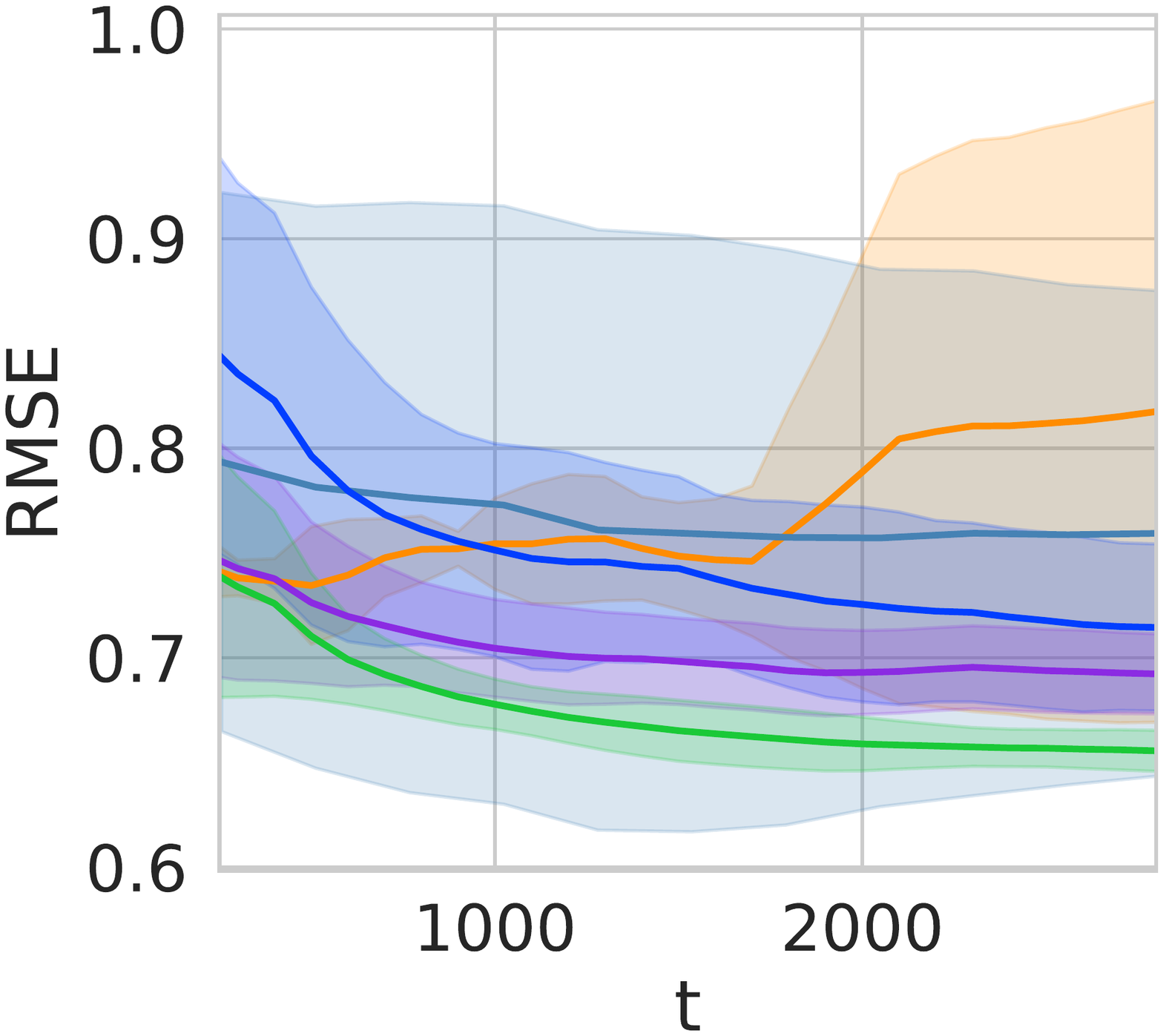}
		\caption{\scriptsize Skillcraft ($n=2855$)}
		\label{supp:fig:skillcraft_rmse}   
	\end{subfigure}
	\hfill
	\begin{subfigure}{0.19\textwidth}
		\includegraphics[width=\linewidth,clip,clip,trim=0cm 4cm 0cm 4cm]{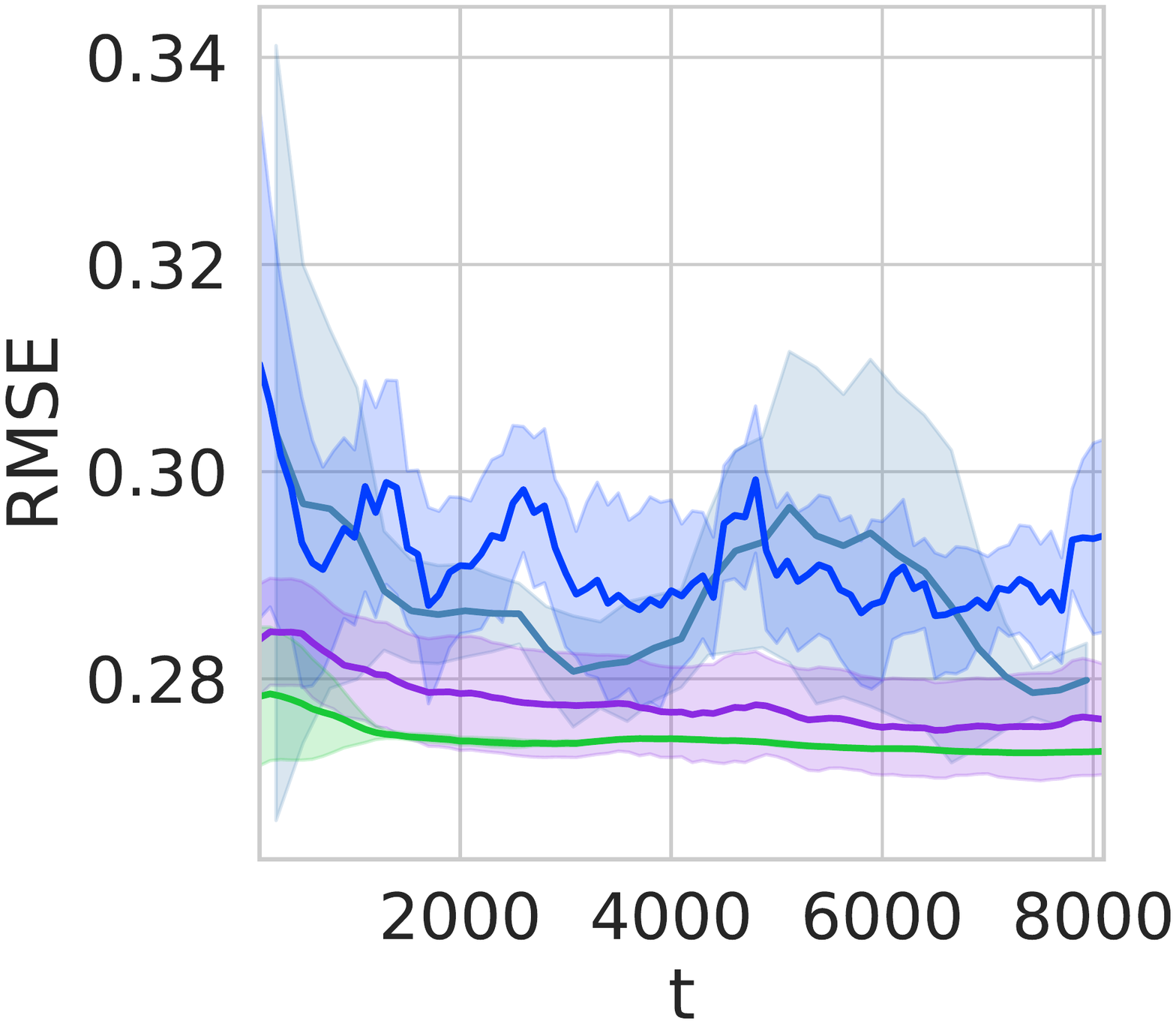}
		\caption{\scriptsize Powerplant ($n=8182$)}
		\label{supp:fig:powerplant_rmse}   
	\end{subfigure}
	\hfill
	\begin{subfigure}{0.19\textwidth}
		\centering
		\includegraphics[width=\textwidth,clip,clip,trim=0cm 4cm 0cm 4cm]{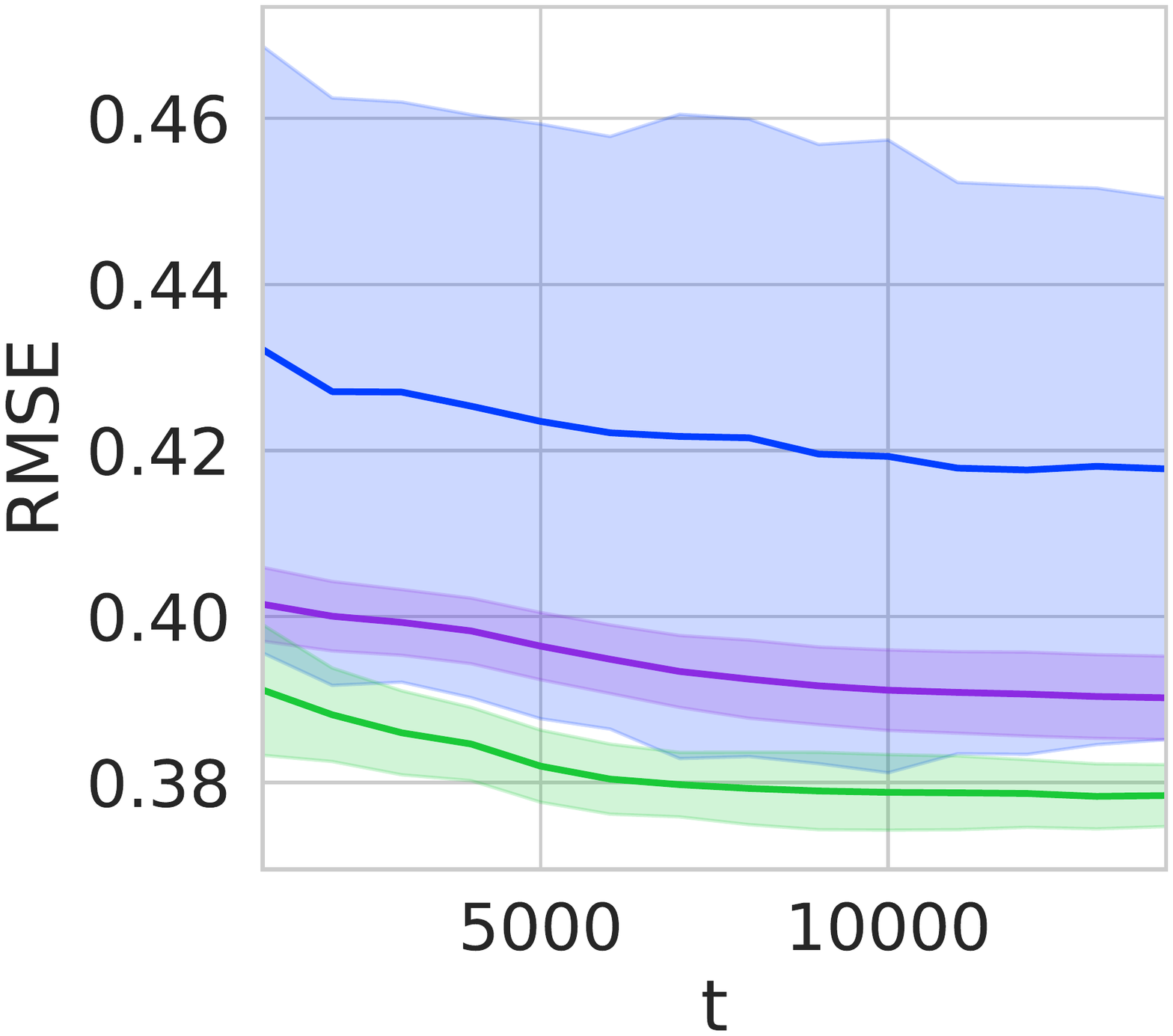}
		\caption{\scriptsize Elevators ($n=14193$)}
		\label{supp:fig:elevators_rmse}
	\end{subfigure}
	\hfill
	\begin{subfigure}{0.19\textwidth}
		\centering
		\includegraphics[width=\textwidth,clip,clip,trim=0cm 4cm 0cm 4cm]{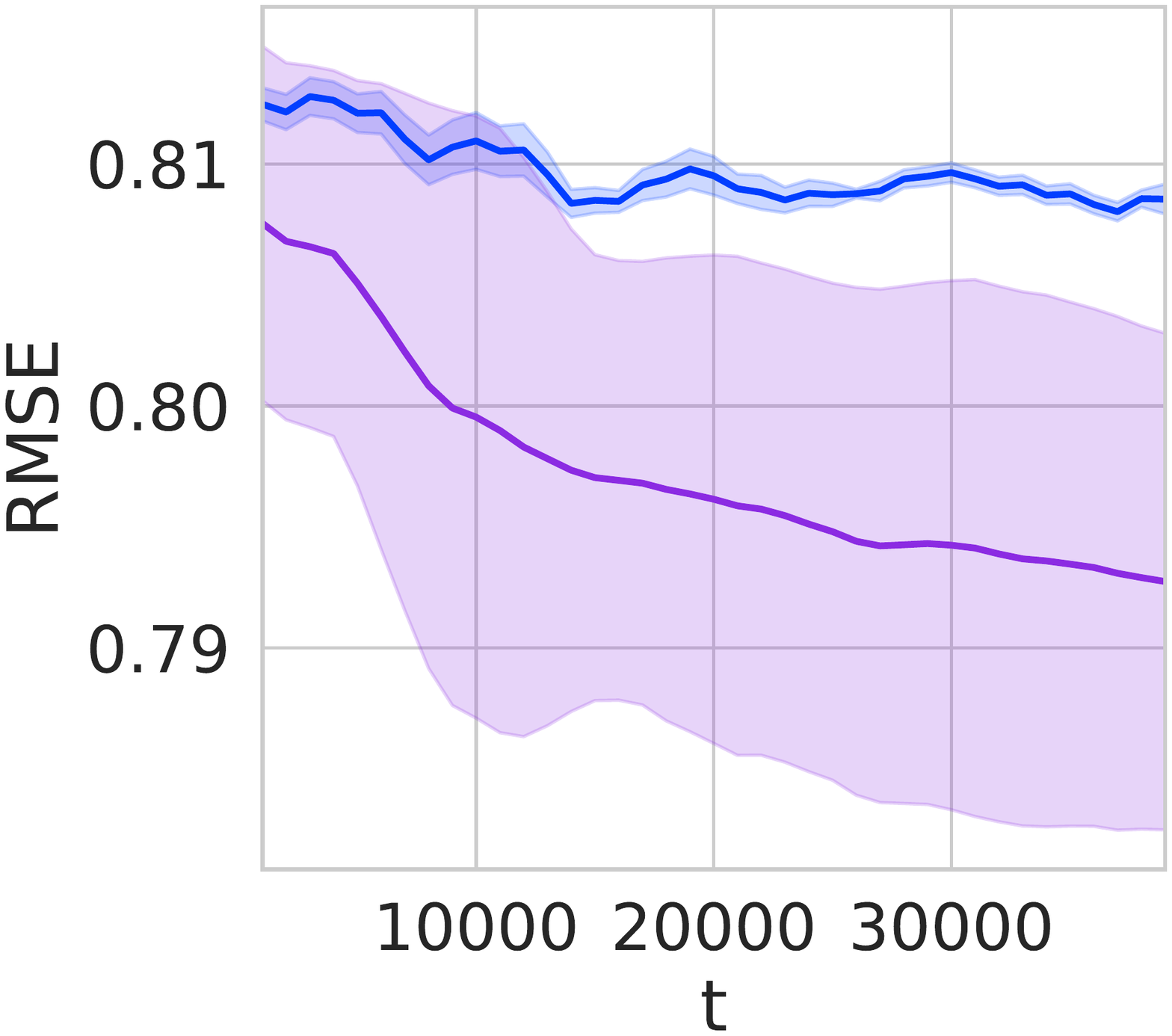}
		\caption{\scriptsize Protein ($n=39100$)}
		\label{supp:fig:protein_rmse}
	\end{subfigure}
	\hfill
	\begin{subfigure}{0.19\textwidth}
		\centering
		\includegraphics[width=\textwidth,clip,clip,trim=0cm 4cm 0cm 4cm]{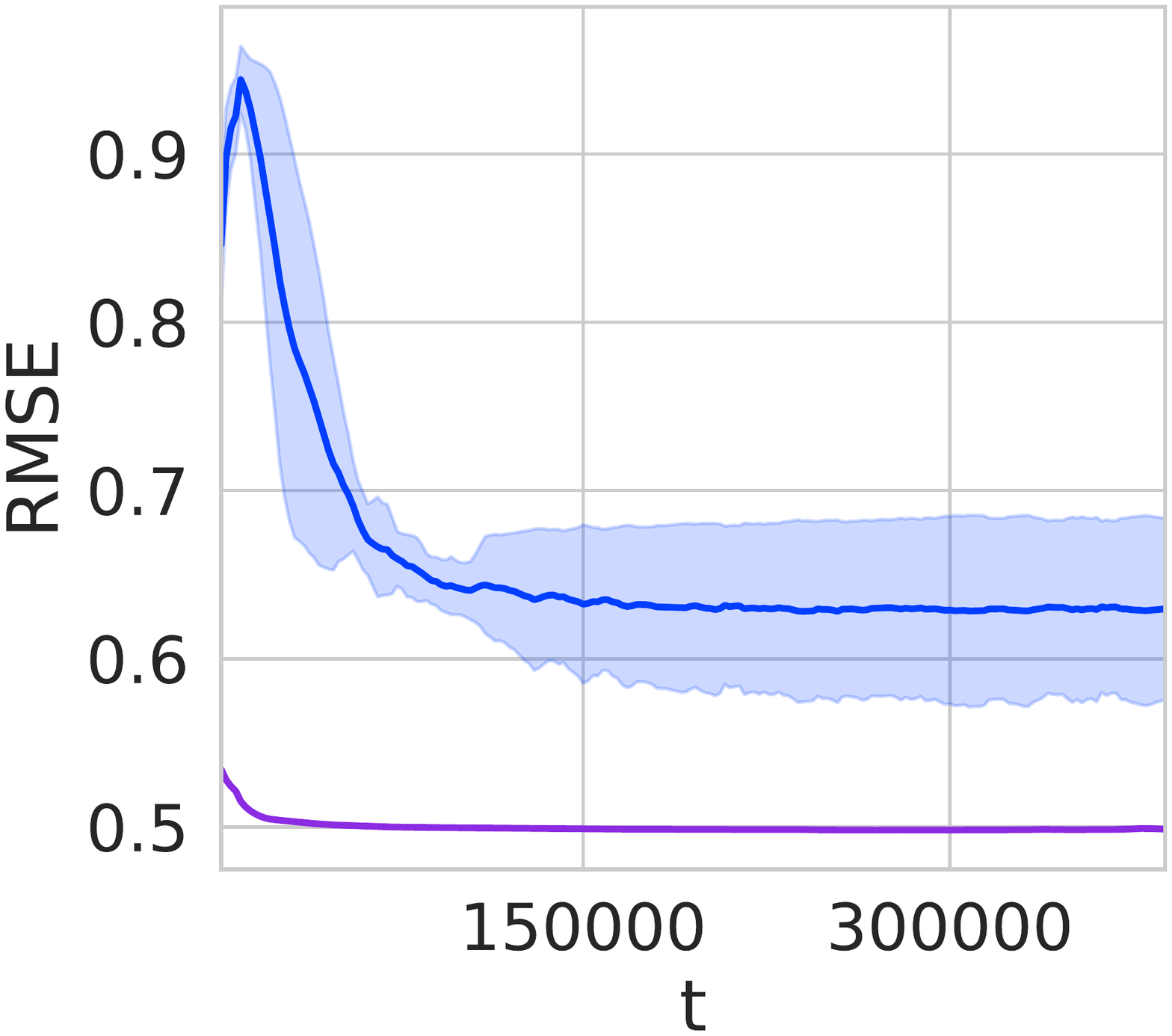}
		\caption{\scriptsize 3DRoad ($n=391000$)}
		\label{supp:fig:3droad_rmse}
	\end{subfigure}
	\caption{
	Online homoskedastic regression on UCI datasets. We compare to local GPs (LGP), O-SGPR, O-SVGP, and exact GPs. Due to memory constraints or numerical issues for other methods, only O-SVGP and WISKI were easily capable of running on the larger tasks (Protein and 3DRoad). WISKI has comparable accuracy to exact methods, with comparable runtime to scalable approximations like OSVGP.
	\textbf{Top:} Test set NLLs. 
	\textbf{Bottom:} Test set RMSEs.
	}
	\label{supp:fig:regression_rmse_comparison}
	\label{main:fig:regression_nll_comparison}
\end{figure*}
Conventionally, learning the kernel hyperparameters $\theta$ of a GP online presents two major challenges. First, the basic form of the gradient of the MLL naively costs at least $\mathcal{O}(n)$, even if scalable methods are employed. Second, after a parameter update, any cached terms that depend on the kernel matrix must be recomputed (e.g. a new factorization of $K_{XX}$). The reformulation of the MLL in Eq. \eqref{main:eq:woodbury_mll} addresses the first challenge, with a complexity of $\mathcal{O}(rm + m \log m + jr^2)$ (after computing the necessary caches). To address the second challenge, we observe that the combination of the SKI approximation to the kernel matrix and the Woodbury matrix identity in Section \ref{main:subsec:woodbury_inverse} has allowed us to reformulate GP inference entirely in terms of computations whose cost depends only on the number of inducing points and the rank of the matrix decompositions (which is at most $m$, and typically much less than $m$). As a result, we can recompute the necessary caches as needed without any increase in computational effort as $n$ increases.

The computational efficiency of SKI is a direct result of the grid structure imposed on the inducing points. The reduced computational complexity comes at the cost of memory complexity that is exponential in the dimension of the input. In practice, if the input data has more than three or four dimensions, the inputs must be projected into a low-dimensional space. The projection may be random \citep{delbridge2019randomly} or learned \citep{wilson_deep_2015}, depending on the requirements of the task. If the projection is learned, then the parameters $\phi$ of the projection operator $h$ are treated as additional kernel hyperparameters and trained through the marginal log-likelihood. 

In the batch setting the interpolation weights $W$ are updated after every optimization iteration to adapt to the new projected features $h(\vec x_{1:n} ; \phi)$. In the online setting, updating $W$ for every previous observation would be $\mathcal{O}(n)$. Since we cannot update the interpolation weights for old observations, the gradient for the projection parameters at time $t$ can be rewritten as follows:
\begin{align}
	\nabla_{\phi} \mathcal{L}(\phi) &= \nabla_{\phi} \frac{1}{2} \bigg( (\vec y^\top W)_t M_{t-1} (W^\top \vec y)_t \nonumber \\
	 - \frac{1}{1 + \vec v^\top \vec w} & \left( \vec v_t^\top (W^\top \vec y)_t \right)^2 - \log(1 + \vec v_t \vec w_t)\bigg) , \label{main:eq:partial_mll} \\
	\vec w_t &= w(h(\vec x_t ; \phi)), \hspace{4mm} \vec v_t = M_{t-1} \vec w_t, \nonumber \\
	(W^\top \vec y)_t &= \cache{(W^\top \vec y)_{t-1}} + y_t \vec w_t. \nonumber
\end{align}
The gradient in Eq. \eqref{main:eq:partial_mll}
will move the projection parameters $\phi$ in a direction that maximizes the marginal likelihood, assuming $\vec w_{1:t-1}$ are fixed. 
That is, only projections on new data are updated, while the old projections remain fixed.
In contrast to the batch setting, where $\phi$ is jointly optimized with $\theta$, the online update is sequential; whenever a new observation is received $\phi$ is updated through Eq. \eqref{main:eq:partial_mll}, then the GP is conditioned on $(h(\vec x_t ; \phi_t), y_t)$, and finally $\theta$ is updated through Eq. \eqref{main:eq:marginal_likelihood}.
See Appendix \ref{supp:sec:online_dkl} for the full derivation.

\section{EXPERIMENTS}
\label{main:sec:experiments}

To evaluate WISKI, we first consider online regression and binary classification. We then demonstrate how WISKI can be used to accelerate Bayesian optimization, a fundamentally online algorithm often applied to experiment design and hyperparameter tuning \citep{frazier2018tutorial}. Finally we consider an active learning problem for measuring malaria incidence, and show that the scalability of WISKI enables much longer horizons than a conventional GP. 

We compare against exact GPs (no kernel approximations), O-SGPR, O-SVGP \citep{bui_streaming_2017}, sparse variational methods that represents the current gold standard for scalable online Gaussian processes, and local GPs (LGPs) \citep{nguyen-tuong_local_2008}.
All experimental details (hyper-parameters, data preparation, etc.) are given in Appendix \ref{supp:sec:regression}, where we also include ablation studies on the $\beta$ parameter for O-SVGP as well as the the number of inducing points (as we had to modify it to achieve good results in the incremental setting for O-SVGP), $m,$ for WISKI and O-SVGP. Unless stated otherwise, shaded regions in the plots correspond to $\overline{\mu} \pm 2\overline{\sigma}$, estimated from 10 trials.

\subsection{Regression}
\label{main:sec:regression_experiments}

We first consider online regression on several datasets from the UCI repository \citep{Dua:2019}. In each trial we split the dataset into a 90\%/10\% train/test split. We scaled the raw features to the unit hypercube $[-1, 1]^d$ and normalized the targets to have zero-mean and unit variance. Each model learned a linear projection from $\mathbb{R}^d$ to $\mathbb{R}^2$ that was transformed via a batch-norm operation and the non-linear $\mathrm{tanh}$ activation to ensure that the features were constrained to $[-1, 1]^2$. Each model learned an RBF-ARD kernel on the transformed features, except on the \emph{3DRoad} dataset, which did not require dimensionality reduction. We used the same number of inducing points for WISKI, O-SGPR, and O-SVGP and set $n_{\mathrm{max}} = m$ for local GPs. The models were pretrained on 5\% of the training examples, and then trained online for the remaining 95\%. When adding a new data point, we update with a single optimization step for each method, such that the runtime is similar for the scalable methods. Since O-SVGP can be sensitive to the number of gradient updates per timestep, in Figure \ref{supp:fig:num_update_steps_ablation} in the Appendix we provide results for an ablation. 

We show the test NLL and RMSE for each dataset in Figure~ \ref{main:fig:regression_nll_comparison}. For the two largest datasets we only report results for WISKI and O-SVGP. We found that O-SGPR was fairly unstable numerically, even after using an exceptionally large jitter value (0.01) and switching from single to double precision. The exact baseline and the WISKI model overfit less to the initial examples than O-SGPR or O-SVGP. Note that O-SVGP is a much stronger baseline in this experiment since the observations are independently observed than would typically be the case for correlated time-series data, as seen in Figure~\ref{main:fig:online_regression_updates}.

\begin{figure}[t]
	\centering
	\begin{subfigure}{0.23\textwidth}
		\centering
		\includegraphics[width=1.05\linewidth,clip,clip,trim=0cm 4cm 0cm 4cm]{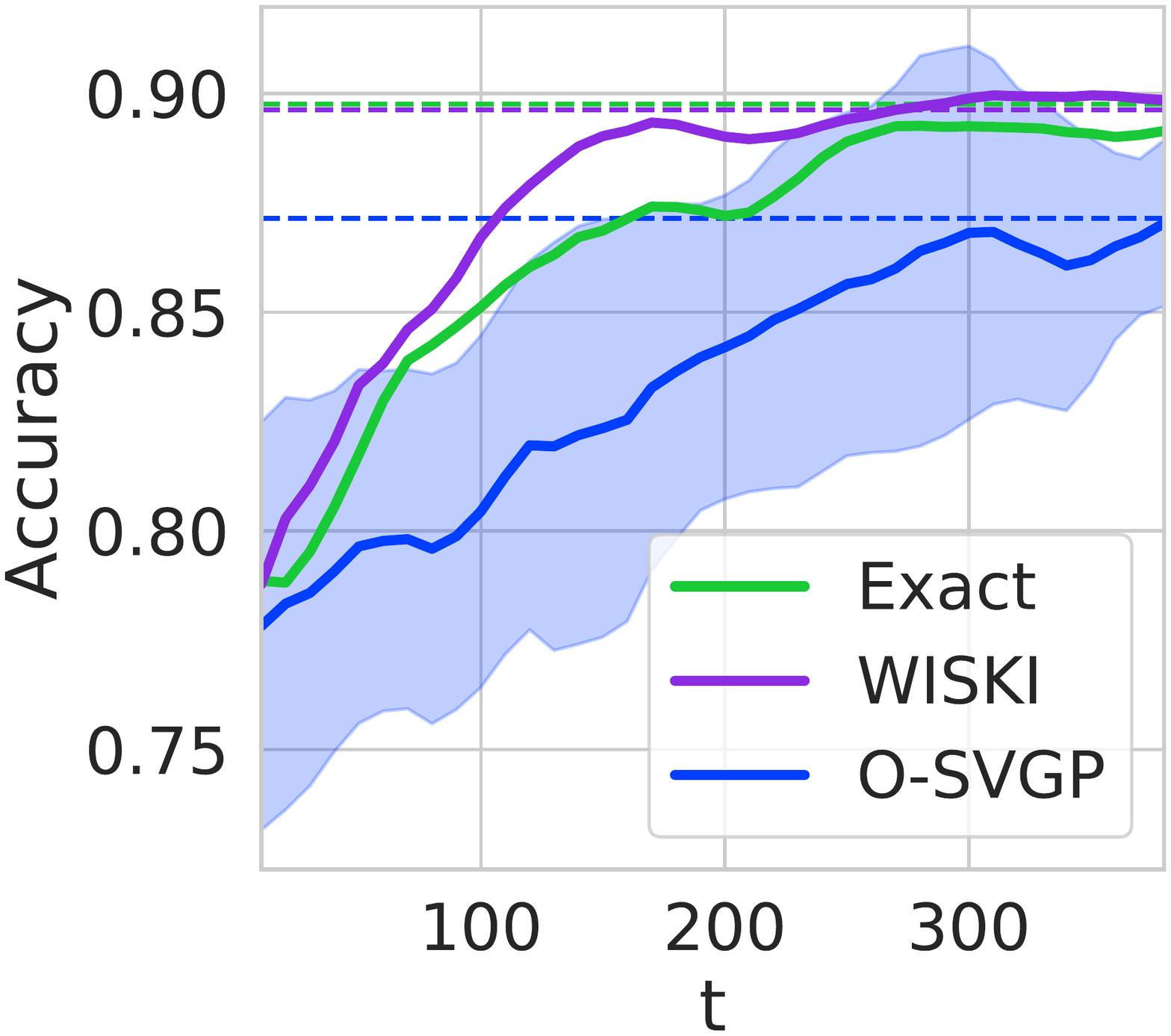}
		\caption{Banana ($n=400$)}
		\label{main:fig:banana_acc}   
	\end{subfigure}
	\hfill
	\begin{subfigure}{0.23\textwidth}
		\centering
		\includegraphics[width=1.05\linewidth,clip,clip,trim=0cm 4cm 0cm 4cm]{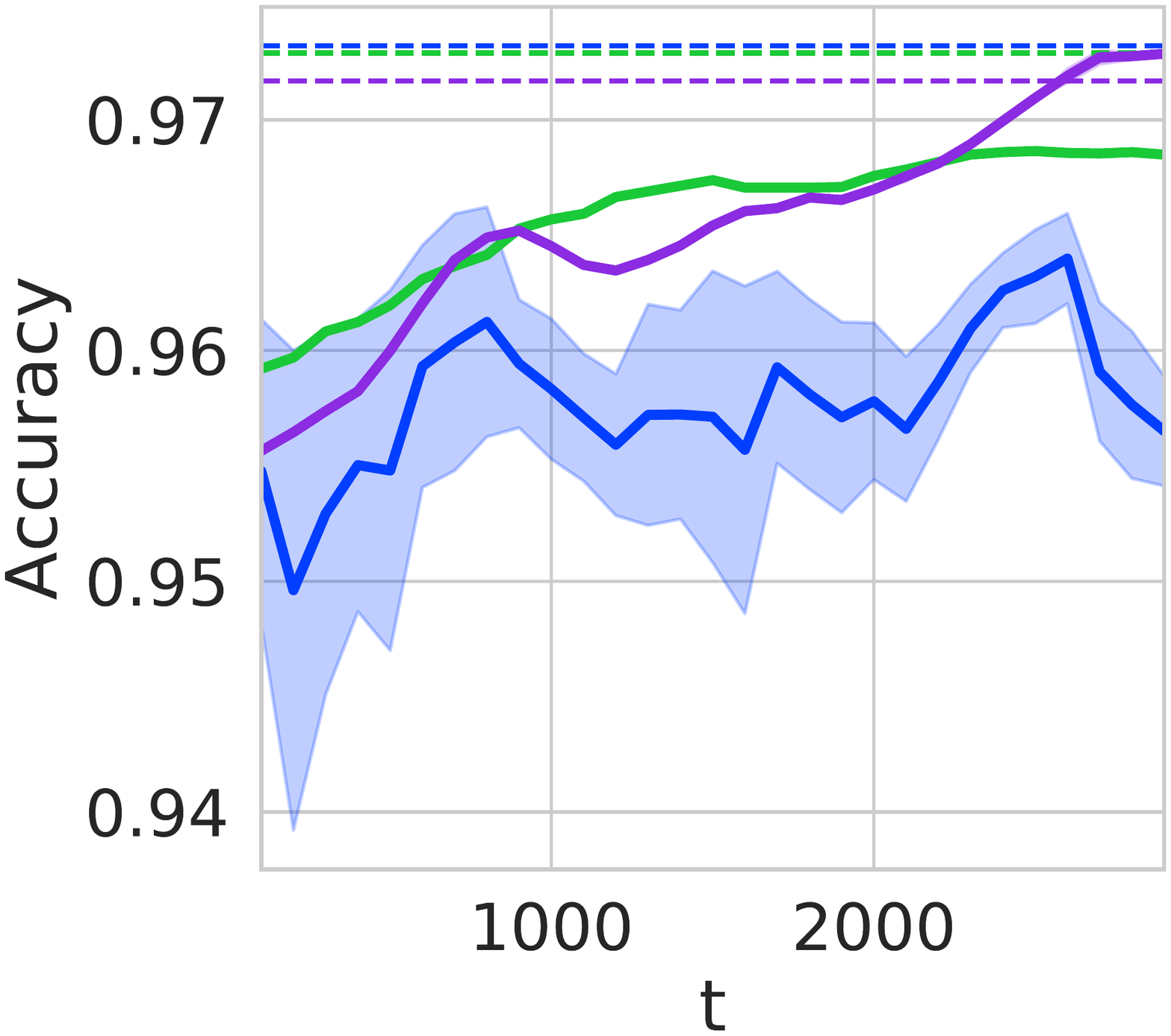}
		\caption{SVM Guide 1 ($n=3000$)}
		\label{main:fig:svm_guide_1_acc}   
	\end{subfigure}
	\caption{A comparison of Dirichlet-based exact and WISKI GP classifiers against an O-SVGP with a binomial likelihood. The exact and WISKI models overfit less to the initial data and ultimately match the performance of their hindsight counterparts trained on the full dataset, shown as a dotted line.}
	\label{main:fig:classification_accuracy_comparison}
\end{figure}

\subsection{Classification}
\label{main:sec:classification_experiments}

We extend WISKI to classification though the Dirichlet-based GP (GPD) classification formulation of \citet{milios2018dirichlet}, 
which reformulates classification as a regression problem with a fixed noise Gaussian likelihood. Empirically the approach has been found to be competitive with the conventional softmax likelihood formulation. In Figure \ref{main:fig:classification_accuracy_comparison} we compare exact and WISKI GPD classifiers to O-SVGP with binomial likelihood
on two binary classification tasks, Banana\footnote{\url{https://raw.githubusercontent.com/thangbui/streaming_sparse_gp/master/data}} and SVM Guide 1 \citep{CC01a}. Banana has 2D features, and SVM Guide has 4D features, so we did not need to learn a projection. As in the UCI regression tasks, WISKI and O-SVGP both had 256 inducing point, each model used an RBF-ARD kernel, and the classifiers were pretrained on 5\% of the training examples and trained online on the remaining 95\%. In both cases the WISKI classifier outperformed the O-SVGP baseline, and matched the accuracy of the the exact baseline. 

\subsection{Bayesian Optimization}
\label{main:sec:bayes_opt_experiments}

\begin{figure*}[t!]
	\centering
	\begin{subfigure}{0.44\textwidth}
	\includegraphics[angle=90,height=4cm,clip,clip,trim=2.5cm 0cm 2.5cm 0cm]{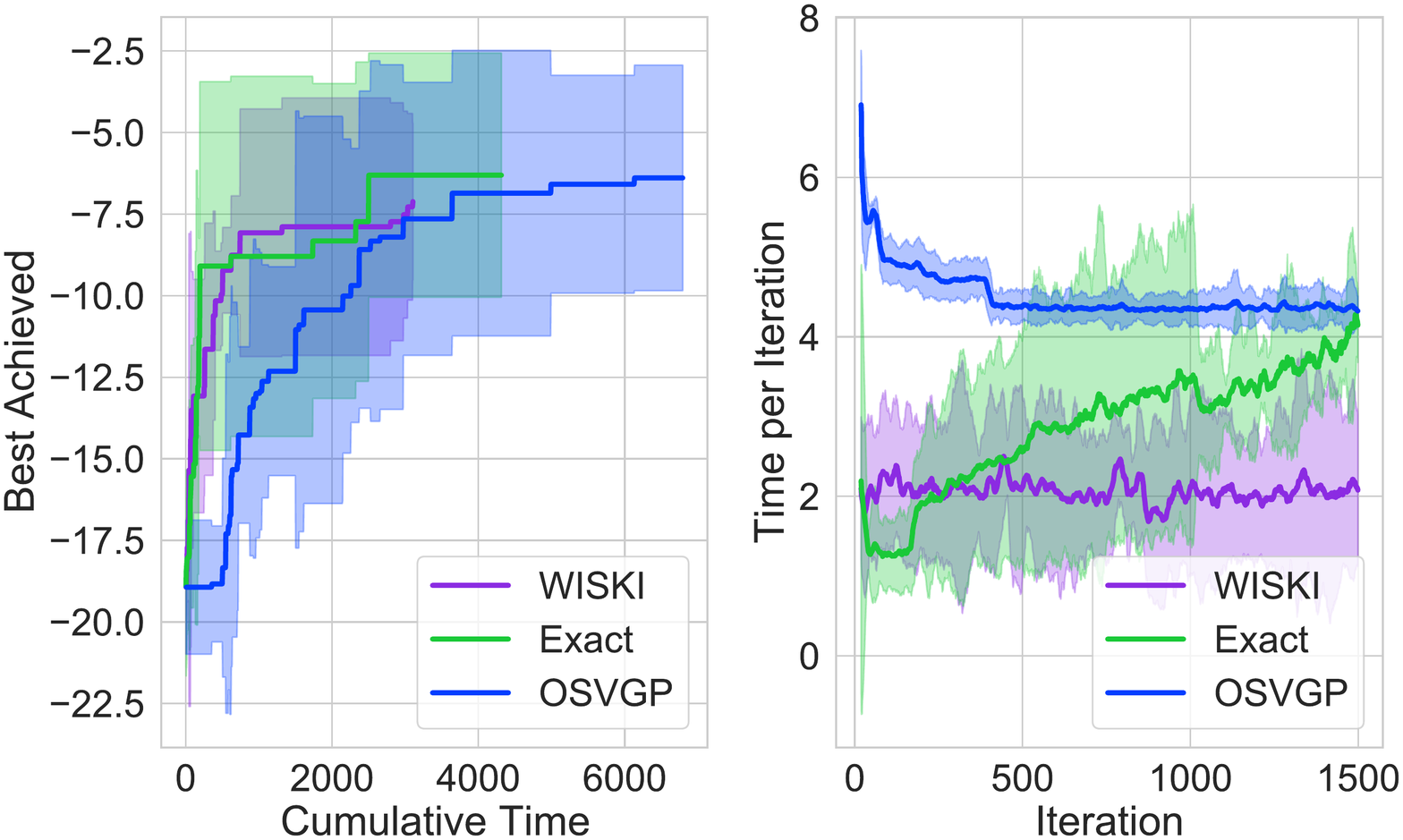}
		\caption{Bayesian optimization.}\label{main:fig:bo}
	\end{subfigure}
	\hspace{-0.75cm}
	\begin{subfigure}{0.24\textwidth}
	\includegraphics[height=4cm,clip,clip,trim=0cm 3.2cm 0cm 3.2cm]{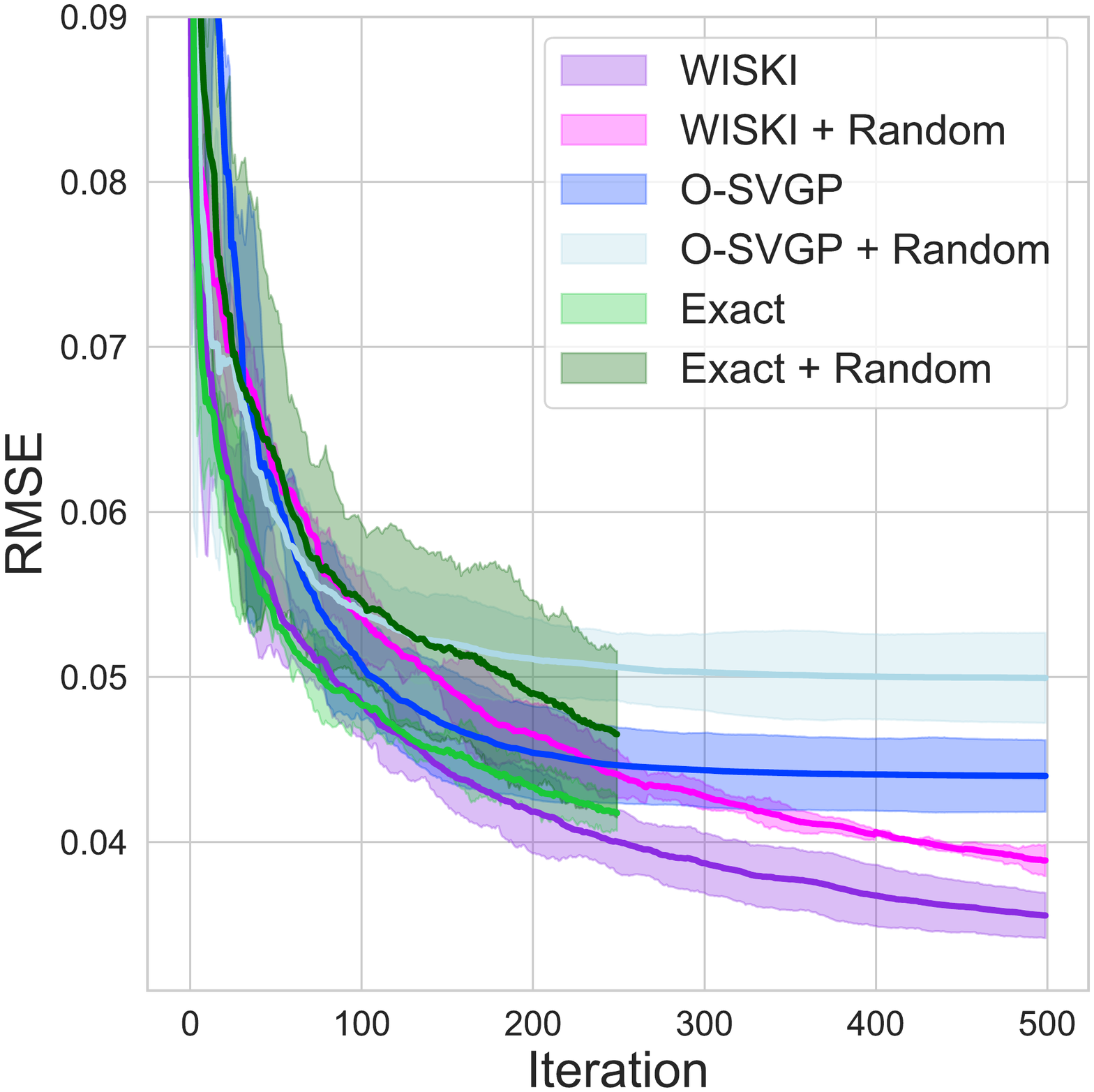}
		\caption{Active learning, RMSE.}\label{main:fig:malaria_active_learning}
	\end{subfigure}
	\hspace{-0.1cm}
	\begin{subfigure}{0.3\textwidth}
			\includegraphics[angle=90,height=4cm]{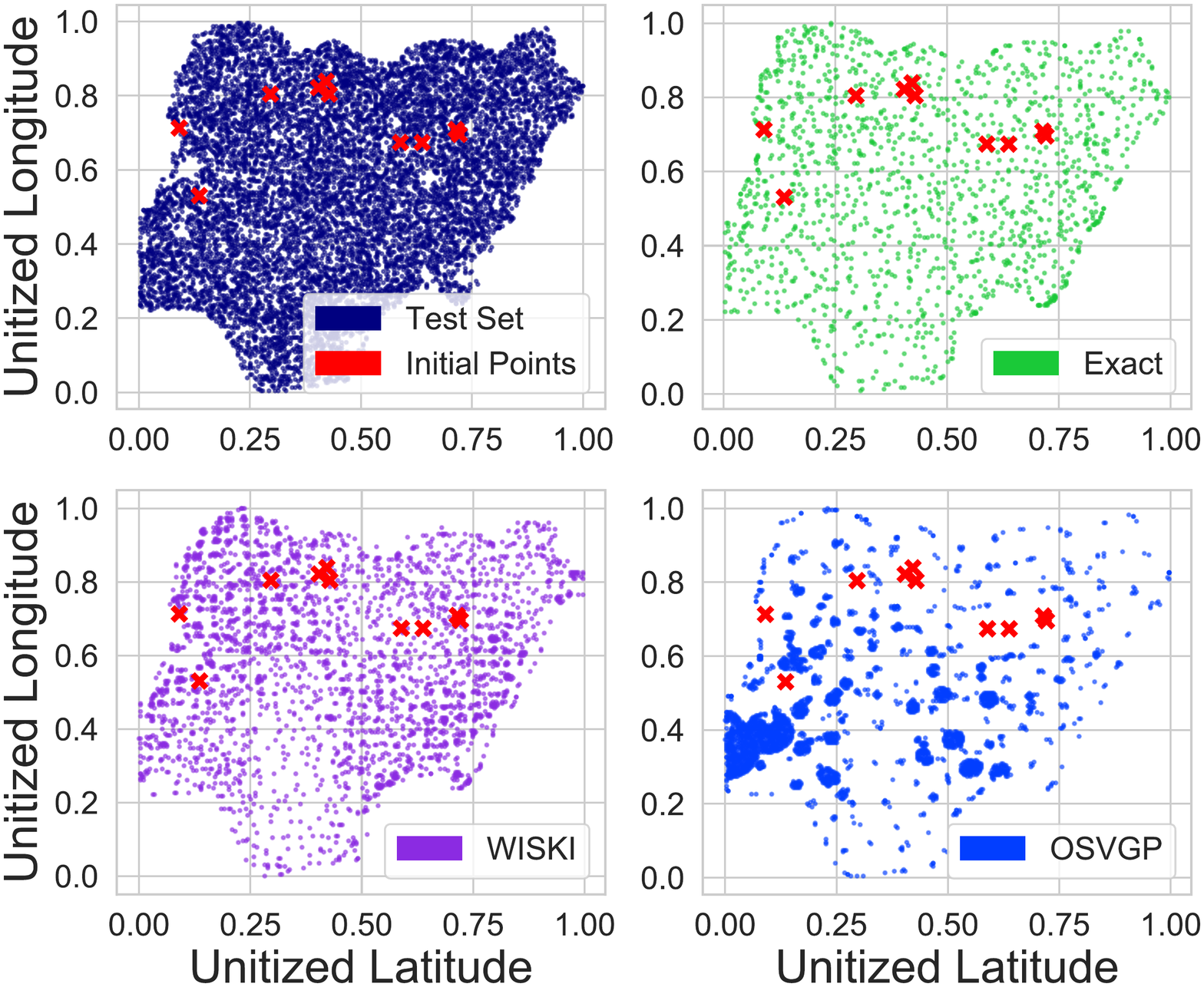}
		\caption{Active learning, fantasy points.}\label{main:fig:malaria_queried_points}
	\end{subfigure}
	\caption{\textbf{(a):} Objective value as a function of cumulative time and time per iteration on the Levy test problem with noise standard deviation $10.0$, while performing Bayesian optimization with EI acquisition for $1500$ steps with a batch size of $3$ so that by the end $4500$ data points have been acquired. WISKI allows rapid updates of the posterior surrogate objective out to thousands of observations, while preserving the rapid convergence rate and asymptotic optimality of the exact GP. 
	\textbf{(b):} RMSE on the test set after choosing new points either randomly or with qnIPV (for WISKI and exact GPs) or by the maximal posterior variance (for O-SVGP). WISKI is able to continue improving the downstream error throughout the entire experiment matching the performance of the exact GP, while O-SVGP's performance flatlines. We also compare against the RMSE of models that have data points randomly selected (shown with Random in the legend). \textbf{(c):} The test set (navy), as well as points chosen for all three methods with initial points (red). WISKI and the exact GP query the entire support, while O-SVGP queries clump together.}
\end{figure*}

In Bayesian optimization (BO) one optimizes a black-box function by iteratively conditioning a surrogate model on observed data and choosing new observations by optimizing an acquisition function based on the model posterior \citep{frazier2018tutorial}. Thus, BO requires efficient posterior predictions, updates to caches as new data are observed, and hyperparameter updates. While BO has historically been applied only to expensive-to-query objective functions, we demonstrate here that large-scale Bayesian optimization is possible with WISKI. Our BO experiments are conducted as follows: we choose $5$ initial observations using random sampling, then iteratively optimize a batched version of upper confidence bound (UCB) (with $q = 3$) using BoTorch \citep{balandat2019botorch} and compute an online update to each GP model, before re-fitting the model.
Accurate model fits are critical to high performance; therefore, we wish to use as many inducing points for WISKI and O-SVGP as possible. 
For both methods, we $1000$ inducing points. We show the results over four trials plotting mean and two standard deviations in Figure \ref{main:fig:bo} for the Ackley benchmarks.
On both problems, WISKI is significantly faster than the exact GP and O-SVGP, while achieving comparable performance on Levy.
We show results over a wider range of test functions in Appendix \ref{supp:bo}, along with the best achieved point plotted against the number of steps and the average time per step.

\subsection{Active Learning}
\label{main:sec:active_learning_experiments}

Finally, we apply WISKI to an active learning problem inspired by \citet{balandat2019botorch}. We consider data describing the infection rate of \emph{Plasmodium falciparum} (a parasite known to cause malaria)\footnote{Downloaded from the Malaria Global Atlas.} in 2017. We wish to choose spatial locations to query malaria incidence in order to make the best possible predictions on withheld samples.
To selectively choose points, we minimize the negative integrated posterior variance \citep[NIPV,][]{seo2000gaussian}, defined as 
\begin{align*}
	\text{NIPV}(x) := -\int_{\mathcal{X}} \mathbb{E}(\mathbb{V}(f(x) | \mathcal{D}_{\bm{x}} ) | \mathcal{D}) dx.
\end{align*}
Optimizing this acquisition function amounts to finding the batch of data points $\vec x_{1:q}$, the fantasy points, which when added into the GP model will reduce the variance on the domain of the model the most. 
Here, we randomly sample $10,000$ data points in Nigeria to serve as a test set that we wish to reduce variance on and select $q = 6$ data points at a time from a held-out training set (to act as a simulator) at a time; the inner expectation drops out because the posterior variance only depends on the fantasy points and the currently observed data, and not any fantasized responses.
Stochastic variational models do not have a straightforward mechanism for fantasizing (i.e. re-computing the posterior variance after updating a new data point conditional on the fantasy points), so we instead query the test set predictive variance and then choose the training points closest to the six test set points with maximum predictive variance.

As both mean and variances are available for the given locations, we model the data with a fixed noise Gaussian process with scaled Matern $0.5$ kernels, beginning with an initial set of $10$ data points, and iterating out for $500$ iterations for WISKI and O-SVGP and $250$ iterations for an exact GP model (the limit of data points that a single GPU could handle due to the large amount of test points). We show the results of the experiment in Figure \ref{main:fig:malaria_active_learning} across three trials, finding that all of the models reduce the RMSE considerably from the initial fits; however, O-SVGP and its random counterpart stagnate in RMSE after about 250 trials, while both the exact GP and WISKI continue improving throughout the entire experiment.
Closer examination of the points queried by all of the three methods in Figure \ref{main:fig:malaria_queried_points}, we find that the points queried by O-SVGP tend to clump together, locally reducing variance, while WISKI and the exact GP choose points throughout the entire support of the test set, choosing points which better reduce global variance.

\section{CONCLUSION}
We have shown how to achieve constant-time online updates with Gaussian processes while retaining exact inference. Our approach, WISKI, achieves comparable performance to Gaussian processes with exact kernels, and comparable speed to state-of-the-art streaming Gaussian processes based on variational inference. Despite the present day need for scalable online probabilistic inference, recent research into online Gaussian processes has been relatively scarce.
We hope that our work is a step towards making streaming Bayesian inference more widely applicable in cases when both speed and accuracy are crucial for online decision making.

\subsection*{Acknowledgements}
WJM, SS, AGW are supported by an Amazon Research Award, NSF I-DISRE 193471, NIH R01 DA048764-01A1, NSF IIS-1910266, and NSF 1922658 NRT-HDR: FUTURE Foundations, Translation, and Responsibility for Data Science.
WJM was additionally supported by an NSF Graduate Research Fellowship under Grant No. DGE-1839302.
SS is additionally supported by the United States Department of Defense through the National Defense Science \& Engineering Graduate (NDSEG) Fellowship Program. We'd like to thank Max Balandat, Jacob Gardner,  and Greg Benton for helpful comments.

\bibliographystyle{apalike}
\bibliography{refs}

\appendix
\onecolumn
\aistatstitle{Kernel Interpolation for Scalable Online Gaussian Processes: \\Supplemental Material}
\setcounter{figure}{0}
\setcounter{equation}{0}
\renewcommand\thefigure{A.\arabic{figure}}
\renewcommand\theequation{A.\arabic{equation}}

\section{FULL DERIVATIONS}
\label{supp:sec:derivations}
\subsection{Efficient Computation of GP Predictive Distributions}

In this section we provide a brief summary of a major contribution of \citet{pleiss2018constant}. Since our cached approach to online inference was partially inspired by the approach of \citet{pleiss2018constant}, it is helpful to first understand how predictive means and variances are efficiently computed in the batch setting.

The Lanczos algorithm is a Krylov subspace method that can be used as a subroutine to solve linear systems (i.e. the conjugate gradients algorithm) or to solve large eigenvalue problems \citep{golub2012matrix}. Given a square matrix $A \in \mathbb{R}^{n \times n}$ and initial vector $\mathbf b \in \mathbb{R}^n$, the $d$-rank Krylov subspace is defined as $\mathcal{K}_d(A, \mathbf b) := \text{span}\{\mathbf b, A \mathbf b, A^2 \mathbf b, \dots, A^{d-1}\mathbf b\}$. The Lanczos algorithm is an iterative method that produces (after $d$ iterations) an orthogonal basis $Q_d \in \mathbb{R}^{n \times d}$ and symmetric tridiagonal matrix $T_d \in \mathbb{R}^{d \times d}$ such that $\mathbf q_i \in \mathcal{K}_d(A, \mathbf b)$ and $T_d = Q_d^\top A Q_d$. \footnote{$Q_d$ is conventionally used to denote the $d$-rank Lanczos basis.} If we take $A \approx Q_d T_d Q_d^\top$ and compute the eigendecomposition $T_d = V_d \Lambda_d V_d^\top$, we can write $A \approx SS^\top$, where $S = Q_d V_d \Lambda_d^{1/2}$. Note that if $d = n$ then the decomposition is exact to numerical precision. Hence the computational cost of a root decomposition via Lanczos is $\mathcal{O}(dn^2 + d^2)$ \citep{trefethen1997numerical}. 

The predictive mean caches are straightforward, since they are just the solution $\cache{\mathbf a} = (K_{XX} + \sigma^2 I)^{-1} \mathbf y$ which can be stored regardless of the method used to solve the system (i.e. preconditioned CG, Cholesky factorization, \textit{e.t.c.}). Once computed, in the exact inference setting the predictive mean is given by
\begin{align*}
	\mu_{f | \mathcal{D}}(\mathbf{x}^*) = k(\mathbf{x}^*, X)\cache{\mathbf a}.
\end{align*}

For inference with SKI we take $\cache{\mathbf a} = K_{UU}W^\top(W K_{UU} W^\top + \sigma^2 I)^{-1} \mathbf y$ and
\begin{align*}
	\mu_{f | \mathcal{D}}(\mathbf{x}^*) = w(\mathbf{x}^*)^\top\cache{\mathbf a}.
\end{align*}

For exact inference, the predictive covariance caching procedure of \citet{pleiss2018constant} begins with the root decomposition 
\begin{align}
	K_{XX} + \sigma^2 I = (Q_d V_d \Lambda_d^{1/2})(\Lambda_d^{1/2}V_d^\top Q_d^\top). \label{supp:eq:exact_root_decomp}
\end{align} Since predictive variances require a root decomposition of $(K_{XX} + \sigma^2 I)^{-1}$, they store $\cache{S} = Q_d^\top V_d^\top \Lambda_d^{-1/2}$ and obtain predictive variances as follows:
\begin{align}
	\sigma^2_{f | \mathcal{D}}\left(\mathbf{x}^* \right) = k(\mathbf{x}^*, \mathbf{x}^*) - k(\mathbf{x}^*, X)\cache{S}\cache{S}^\top k(X, \mathbf{x}^*). \label{supp:eq:love_pred_var}
\end{align}

For inference with SKI the procedure is much the same, except the root decomposition in Eq. \ref{supp:eq:exact_root_decomp} is replaced with that of the SKI kernel matrix,
\begin{align}
	W K_{UU}W^\top + \sigma^2 I = (Q_d V_d \Lambda_d^{1/2})(\Lambda_d^{1/2}V_d^\top Q_d^\top),
\end{align}
$\cache{S} = K_{UU}W^\top Q_d^\top V_d^\top \Lambda_d^{-1/2}$, and Eq. \ref{supp:eq:love_pred_var} is modified to 
\begin{align}
	\sigma^2_{f | \mathcal{D}}\left(\mathbf{x}^{(i)}, \mathbf{x}^{(j)} \right) = k(\mathbf{x}^{(i)}, \mathbf{x}^{(j)}) - w(\mathbf{x}^{(i)})^\top \cache{S}\cache{S}^\top w(\mathbf{x}^{(j)}).
\end{align}

\subsection{Deriving the WISKI Predictive Mean and Variance}
In this section we derive the Woodbury Inverse SKI predictive distributions. In contrast to \citet{pleiss2018constant}, the WISKI predictive mean and covariance can be formulated in terms of quantities that can be cached in $\mathcal{O}(m^2)$ space and updated with new observations in constant time. Recall the form of the Woodbury SKI inverse, $M := \textcolor{red}{(\sigma^2 K_{UU}^{-1} + W^\top W)^{-1}}$. For both the predictive mean and variance we begin with the standard SKI form, and show how to derive the WISKI form.

\paragraph{Predictive Mean}
\begin{align*}
	\mu_{f | \mathcal{D}}(\mathbf{x}^*) &= w(\mathbf x^*)^\top K_{UU} W^\top (W K_{UU} W^\top + \sigma^2 I)^{-1} \mathbf y, \\
	&= w(\mathbf x^*)^\top \textcolor{red}{(I + \sigma^{-2}K_{UU} W^\top W)^{-1} (\sigma^{-2} K_{UU})} \cache{W^\top \mathbf y}, \\
	&= w(\mathbf x^*)^\top \textcolor{red}{\left((\sigma^{-2}K_{UU})(\sigma^2 K_{UU}^{-1} + W^\top W)\right)^{-1} (\sigma^{-2}K_{UU})} \cache{W^\top \mathbf y}, \\
	&= w(\mathbf x^*)^\top \textcolor{red}{(\sigma^2 K_{UU}^{-1} + W^\top W)^{-1} K_{UU}^{-1} K_{UU}} \cache{W^\top \mathbf y}, \\
	&= w(\mathbf x^*)^\top M \cache{W^\top \mathbf y}. \\
	&= w(\mathbf x^*)^\top (\sigma^{-2} K_{UU}\cache{W^\top \mathbf{y}} - \sigma^{-2} K_{UU} \cache{L} (I + \sigma^{-2} \cache{L}^\top K_{UU} \cache{L})^{-1} \cache{L}^\top K_{UU} \cache{W^\top \mathbf{y}}.
\end{align*}
The second line follows from the push-through identity (a special case of the Woodbury matrix identity). 

\paragraph{Predictive Covariance}
The low-rank SKI predictive covariance of a GP is given (elementwise) by
\begin{align*}
	\sigma^2_{f | \mathcal{D}}\left(\mathbf{x}^{(i)}, \mathbf{x}^{(j)} \right) &= w(\mathbf{x}^{(i)})^\top \textcolor{red}{K_{UU}} w(\mathbf{x}^{(j)}) - w(\mathbf{x}^{(i)})^\top \textcolor{red}{K_{UU} W^\top (W K_{UU} W^\top + \sigma^2 I)^{-1} W K_{UU}} w(\mathbf{x}^{(j)}) \\
	&= \sigma^2 w(\mathbf{x}^{(i)})^\top \textcolor{red}{\left (\sigma^{-2}K_{UU} - (\sigma^{-2}K_{UU}) W^\top (W (\sigma^{-2} K_{UU}) W^\top + I)^{-1} W (\sigma^{-2} K_{UU}) \right)} w(\mathbf{x}^{(j)}) \\ 
	&= \sigma^2 w(\mathbf{x}^{(i)})^\top \textcolor{red}{(\sigma^2 K_{UU}^{-1} + W^\top W)^{-1}} w(\mathbf{x}^{(j)}), \\
	&= \sigma^2 w(\mathbf{x}^{(i)})^\top M w(\mathbf{x}^{(j)}) \\
	&= w(\mathbf{x}^{(i)})^\top \left(K_{UU} - K_{UU} \cache{L} (I + \sigma^{-2} \cache{L}^\top K_{UU} \cache{L})^{-1} \cache{L}^\top K_{UU} \right) w(\mathbf{x}^{(j)})
\end{align*}
The third line immediately follows from an application of the Woodbury matrix identity to $\textcolor{red}{(\sigma^2 K_{UU}^{-1} + W^\top W)^{-1}}$. 
Following \citet{pleiss2018constant}, we may compute two root decompositions of the form $BB^\top = K_{UU}$ and $DD^\top \approx (I + \sigma^{-2} \cache{L}^\top K_{UU} \cache{L})^{-1}$ to speed up predictive variance computation, as this yields efficient diagonal computation:
\begin{align*}
	\sigma^2_{f | \mathcal{D}}\left(\mathbf{x}^{(i)}, \mathbf{x}^{(j)} \right) &= w(\mathbf{x}^{(i)})^\top \left(BB^\top - BB^\top \cache{L} DD^\top \cache{L}^\top BB^\top \right) w(\mathbf{x}^{(j)}).
\end{align*}
As we noted in the main text, at time $t+1$, $\cache{Q_t}$ can be updated with a new observation in constant time via a Sherman-Morrison update, and $(W^\top \mathbf y)_{t+1} = \cache{(W^\top \mathbf y)_t} + y_{t+1}w(\mathbf x_{t+1})$.

\subsection{Conditioning on New Observations}

\paragraph{Updating the Marginal Likelihood}
For fixed $n,$ the Woodbury version of the log likelihood in WISKI is constant in $n$ after an initial $\mathcal{O}(n)$ precomputation of $W^\top W,$ as the only terms that ever get updated are the scalar $\sigma^2$ and the $m \times m$ matrix $K_{UU}^{-1};$
this is in and of itself an advance over the computation speeds of other Gaussian process models, including SKI \citep{wilson_kernel_2015}.

\paragraph{Updating $W^\top W$.}

To update $W^\top W$ as we see new data points, we follow the general strategy of \citet{gill1974methods} by performing rank-one updates to root decompositions:
\begin{align*}
	\tilde{A} = A + zz^\top 
	&= L(I + pp^\top)L^\top, \hspace{2cm} p = L^{-\top} z, \\
	&= L B B^\top L^\top = \tilde{L} \tilde{L}^\top, \hspace{1.3cm} BB^\top = I + pp^\top, \tilde{L} = L B
\end{align*}
In our setting, the update is given by 
\begin{align*}
	(W^\top W)_{t+1} = (W^\top W)_t + \vec w_{\vec x_{t+1}}\vec w_{\vec x_{t+1}}^\top.
\end{align*}
For full generality, we will assume $q$ new points come at once, making $\vec w_{\vec x_{t+1}} \in \mathbb{R}^{m \times q}$.
Recall that $\cache{JJ^\top} = (W^\top W)^+$, let $\vec p = \cache{J_t}^\top \vec w_{\vec x_{t+1}}$, which is the product of a $r \times m$ matrix and a $m \times q$ matrix, which costs $\mathcal{O}(mrq).$
To compute the decomposition $BB^\top = I_r + \vec p \vec p^\top$ in a numerically stable fashion, we compute the SVD of $\vec p = USV^\top$ and use it to update the root decomposition:
\begin{align*}
	I_r + \vec p \vec p^\top = I_r + U S V^\top V S U^\top = U \diag((S_{ii}^2 + 1); \pmb{1}_{r-q}) U^\top = U \diag(\sqrt{S_{ii}^2 + 1}, \pmb{1}_{r-q}) \diag(\sqrt{S_{ii}^2 + 1}, \pmb{1}_{r-q}) U^\top.
\end{align*}
The SVD of this matrix costs $\mathcal{O}(q^2 r),$ assuming $q < r$ ($q=1$ for most applications). 
The inner root is $B = U \diag(\sqrt{S_{ii}^2 + 1}, \pmb{1}_{r-q}),$ and a final matrix multiplication costing $\mathcal{O}(mr^2)$ obtains the expression for the updated root
$\cache{L_{t+1}} = \cache{L_t} B.$
The updated inverse root is obtained similarly by 
$\cache{J_{t+1}} = \cache{J_t} U \diag(1./\sqrt{S_{ii}^2 + 1}, \pmb{1}_{r-q})$
The overall computation cost is then $\mathcal{O}(mrq + q^2 r + mr^2).$

\subsection{Online SKI and Deep Kernel Learning}
\label{supp:sec:online_dkl}

When combining deep kernel learning (DKL) and SKI, the interpolation weight vectors $\vec w_i = w(\vec x_i, \vec u_1, \dots, \vec u_m)$ become $\vec w_i = w(h(\vec x_i; \phi), \vec u_1, \dots, \vec u_m)$, where $h(\cdot ; \phi)$ is a feature map parameterized by $\phi$. One implication of this change is that if the parameters of $h$ change, then the interpolation weights must change as well. In the batch setting, the features and associated interpolation weights can be recomputed after every optimization iteration, since the cost of doing so is negligible compared to the cost of computing the MLL. The online setting does not admit the recomputation of past features and interpolation weights, because doing so would require $\mathcal{O}(n)$ work. Hence at any time $t$ we must consider the previous features and interpolation weights $(\vec h_1, \vec w_1), \dots, (\vec h_{t-1}, \vec w_{t-1})$ to be fixed. As a result, when computing the gradient of the MLL w.r.t. $\phi$, we need only consider the terms that depend on $\vec w_t$. 

\textbf{Claim:} \\ 
\begin{align}
	\nabla_{\vec w_t} \log p(\vec y_t | \vec x_{1:t}, \theta) = \nabla_{\vec w_t} \frac{1}{2\sigma^2} \bigg( \vec y_t^\top W_t M_{t-1} W_t^\top \vec y_t
	- \frac{1}{1 + \vec v_t^\top \vec w_t}\left( \vec v_t^\top W_t^\top \vec y_t \right)^2\bigg) - \frac{1}{2}\log(1 + \vec v_t \vec w_t), \label{supp:eq:partial_mll}
\end{align} 
where
\begin{align*}
	\vec w_t = w(h(\vec x_t ; \phi), \vec u_{1:m}), \hspace{8mm} \vec v_t = M_{t-1} \vec w_t. 
\end{align*}

\textbf{Proof:}\\
\begin{align*}
	\log p(\vec y_t | \vec x_{1:t}, \theta)
	=  -\frac{1}{2\sigma^2}(\vec y_t \vec y_t^\top - \vec y_t^\top W_t M_t W_t^\top \vec y_t) -
	\frac{1}{2} \left(\log{|K_{UU}|} - \log{|M_t|} + (n - m) \log{\sigma^2}\right) - \frac{t}{2}\log 2\pi.
\end{align*}

Recalling $M_t := (K_{uu}^{-1} + W_t^\top W_t)^{-1} = (K_{uu}^{-1} + W_{t-1}^\top W_{t-1} + \vec w_t \vec w_t^\top)^{-1}$, by the Sherman-Morrison identity we have
\begin{align}
	M_t &= M_{t-1} - \frac{1}{1 + \vec v_t^\top \vec w_t} \vec v_t \vec v_t^\top. \label{supp:eq:sherman_morrison}
\end{align}

Since $M_{t-1}$ is constant w.r.t. $\vec w_t$, we can substitute Eq. \eqref{supp:eq:sherman_morrison} into the MLL and differentiate w.r.t. $\vec w_t$ to obtain

\begin{align*}
	\nabla_{\vec w_t} \log p(\vec y_t | \vec x_{1:t}, \theta) = \nabla_{\vec w_t} \frac{1}{2\sigma^2} \vec y_t^\top W_t (M_{t-1} - \frac{1}{1 + \vec v_t^\top w_t}v_t v_t^\top) W_t^\top \vec y_t + \frac{1}{2} \log|M_{t-1} - \frac{1}{1 + \vec v_t^\top w_t} \vec v_t \vec v_t^\top|
\end{align*}

The quadratic term straightforwardly simplifies to the first two terms in Eq. \eqref{supp:eq:partial_mll}. The final term results from an application of the matrix-determinant identity, once again dropping any terms with no dependence on $\vec w_t$,

\begin{align*}
	\log|M_{t-1} - \frac{1}{1 + \vec v^\top \vec w} \vec v \vec v^\top| &= \log|M_{t -1}| - \log\left( 1 + \frac{1}{1 + \vec v^\top \vec w} \vec v^\top M_{t-1}^{-1}\vec v \right) \\
	&= \log|M_{t -1}| - \log\left( 1 + \frac{1}{1 + \vec v^\top \vec w} \vec v^\top \vec w \right) \\
	&= \log|M_{t -1}| - \log\left( 1 + \vec v^\top \vec w \right). \hskip0.25\textwidth \blacksquare
\end{align*}

\subsection{Heteroscedastic Fixed Gaussian Noise Likelihoods and Dirichlet Classification}\label{app:het_likelihoods}

For a \emph{fixed} noise term, the Woodbury identity still holds and we can still perform the updates in constant time.
For fixed Gaussian noise, the term training covariance becomes 
\begin{align*}
	K_{XX} &\approx K_{SKI} = W K_{UU} W + D, \\
	K_{SKI}^{-1} &= D^{-1} - D^{-1} W(K_{UU}^{-1} + W^\top D^{-1} W)^{-1} W^\top D^{-1}.
\end{align*}
Plugging the second line into Eq. \ref{main:eq:woodbury_mll} tells us immediately that we need to store $\cache{\mathbf y D^{-1} \mathbf y}$ instead of $\cache{\mathbf y \mathbf y},$ $\cache{W^\top D^{-1} W}$ instead of $\cache{W^\top W},$ $\cache{W^\top D^{-1} \mathbf y}$ instead of $\cache{W^\top \mathbf y}.$
The rest of the online algorithm proceeds in the same manner as at each step, we update these caches with new vectors.

The heteroscedastic fixed noise regression approach naturally allows us to perform GP inference as in \citet{milios2018dirichlet}.
Given a one-hot encoding of the class probabilities, e.g. $y = e_c$ where $c$ is the class number, they derive an approximate likelihood so that the transformed regression targets are 
\begin{align*}
	\tilde{y}_i = \log \alpha_i - \tilde{\sigma}_i^2 / 2, \hspace{3cm} \tilde{\sigma}^2_i = \log(1 + 1/\alpha_i),
\end{align*}
where $\alpha_i = I_{y_i=1} + \alpha_{\epsilon},$ where $\alpha_{\epsilon}$ is a tuning parameter. We use $\alpha_{\epsilon} = 0.01$ in our classification experiments. 
As there are $C$ classes, we must model each class regression target; \citet{milios2018dirichlet} use independent GPs to model each class as we do.
The likelihood at each data point over each class target is then $p(\tilde{y}_i | \mathbf f) = \mathcal{N}(f_i, \tilde{\sigma}_i^2),$ which is simply a heteroscedastic fixed noise Gaussian likelihood.
Posterior predictions are given by computing the arg max of the posterior mean, while posterior class probabilities can be computed by sampling over the posterior distribution and using a softmax (Equation 8 of \citet{milios2018dirichlet}).

\section{CHALLENGES OF STREAMING VARIATIONAL INFERENCE FOR GPS}
\label{supp:streamingvi_gps}

\begin{figure}[t]
	\centering
	\includegraphics[angle=90,width=\textwidth,clip,clip,trim=6cm 0cm 6cm 0cm]{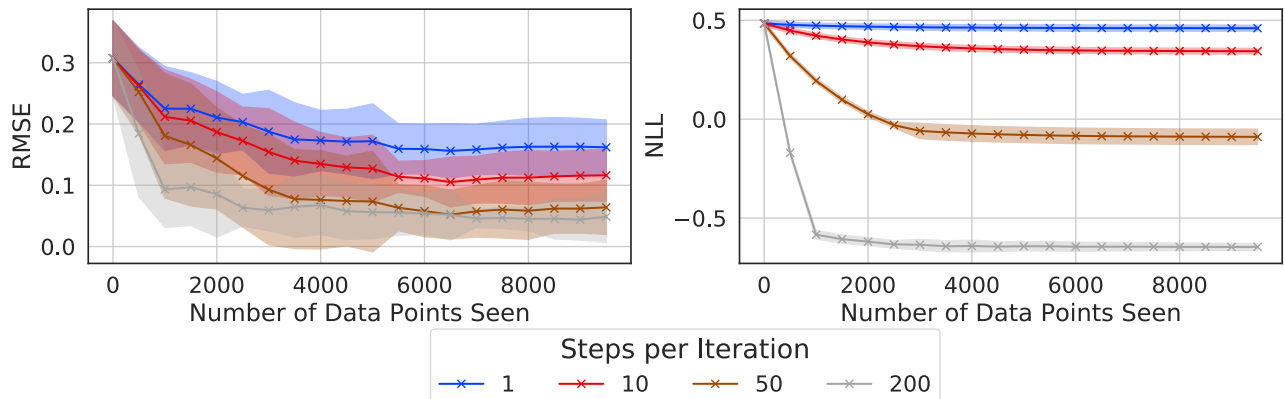}
	\caption{(\textbf{Left:}) MSEs through the course of the dataset stream for up to $10,000$ data points coming in batches of $500$ data points for online SVGP. We varied the number of optimization steps per batch, finding that at least $10$ steps were required to achieve good performance. The data points are drawn from a synthetic sine function corrupted by Gaussian noise. (\textbf{Right:}) NLLs over the course of the dataset stream; again, we see that many optimization steps are needed to decrease the NLL on the test set over the course of the stream.}
	\label{supp:fig:osvgp_steps}
\end{figure}

\begin{figure}
    \centering
    \begin{subfigure}{0.32\textwidth}
        \includegraphics[width=\textwidth]{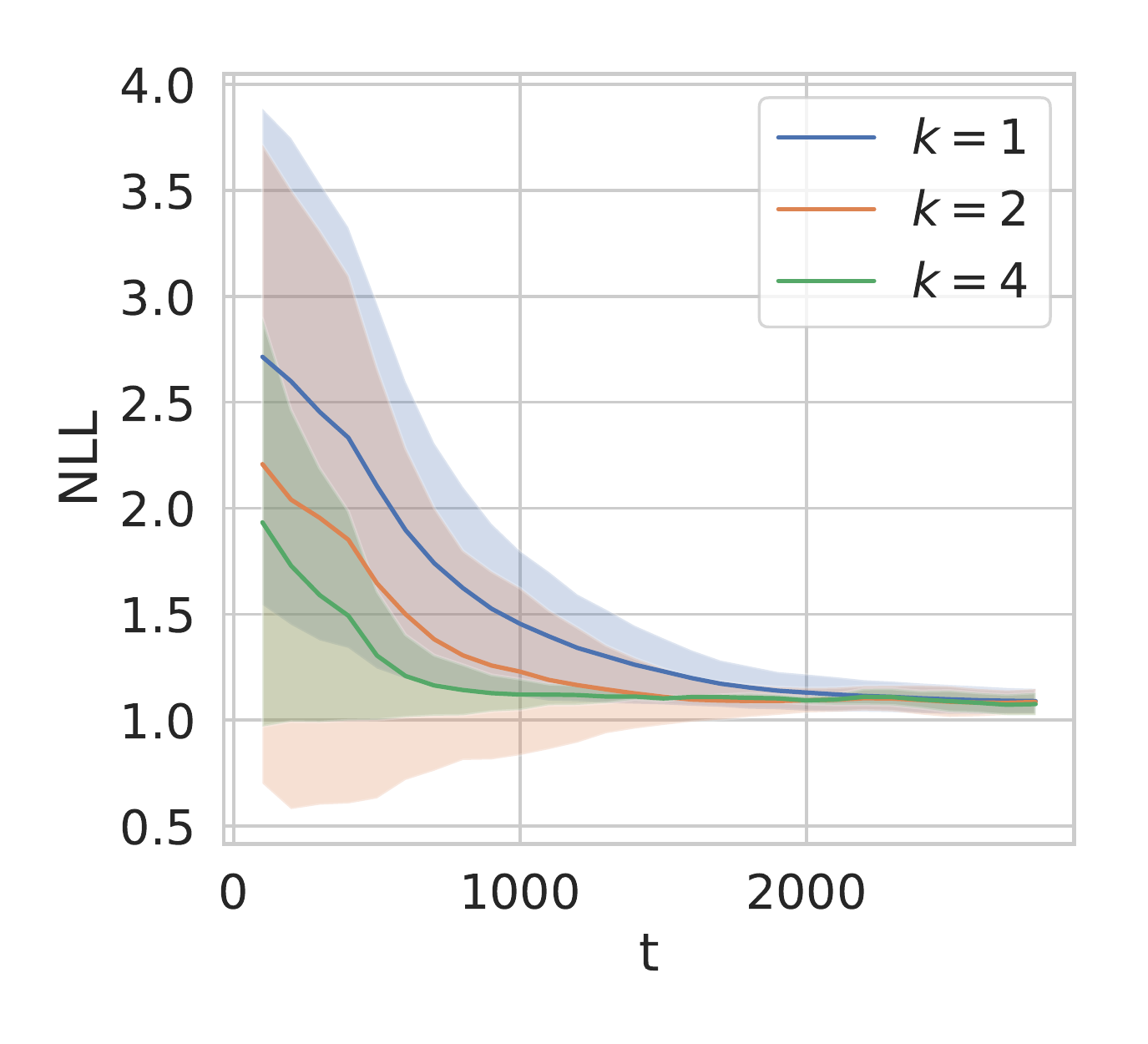}
        \caption{Skillcraft}
        \label{supp:fig:num_update_steps_ablation_skillcraft}
    \end{subfigure}
    \hfill
    \begin{subfigure}{0.32\textwidth}
        \includegraphics[width=\textwidth]{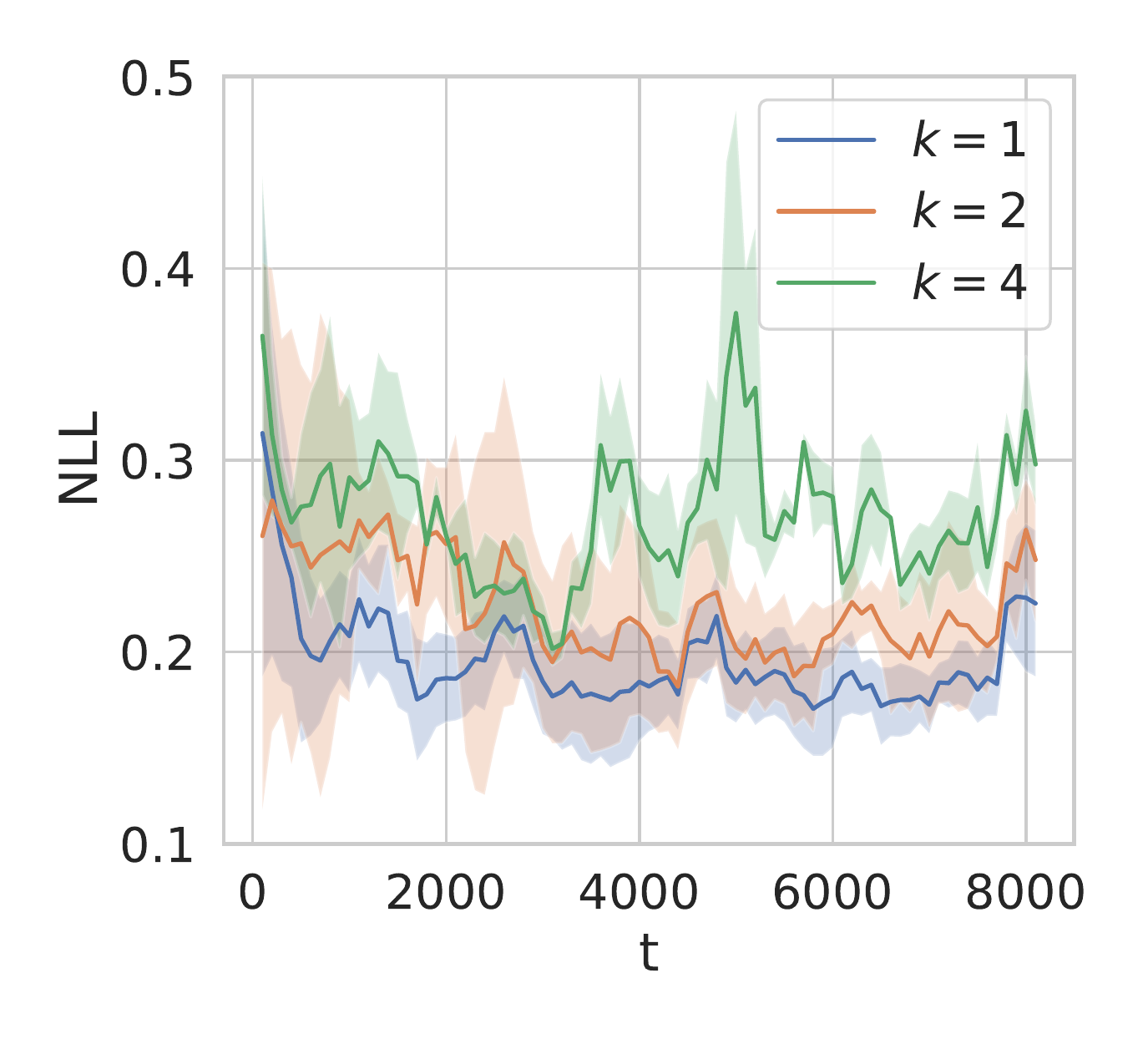}
        \caption{Powerplant}
        \label{supp:fig:num_update_steps_ablation_powerplant}
    \end{subfigure}
    \hfill
    \begin{subfigure}{0.32\textwidth}
        \includegraphics[width=\textwidth]{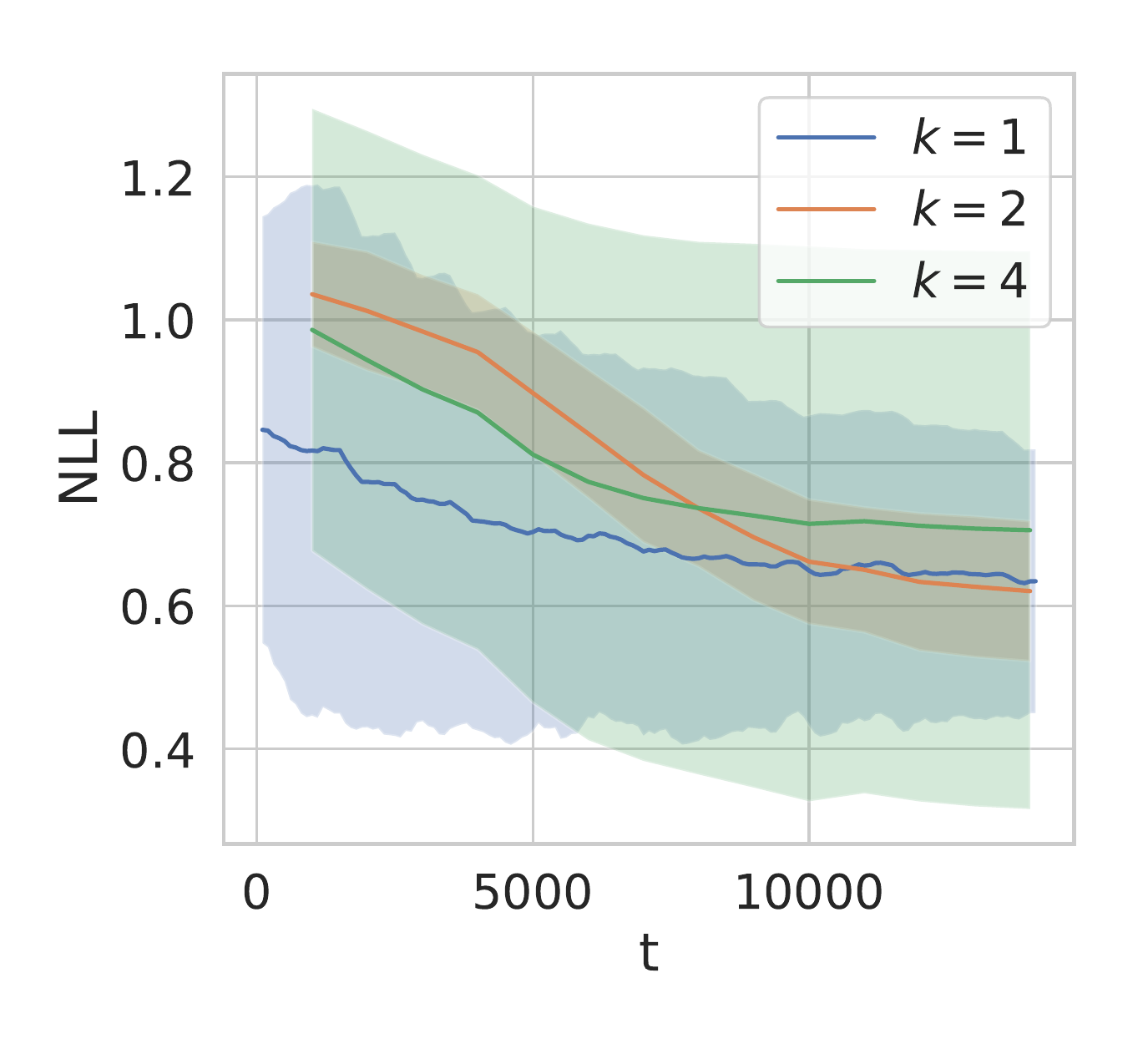}
        \caption{Elevators}
        \label{supp:fig:num_update_steps_ablation_elevators}
    \end{subfigure}
    \caption{Here we ablate the number of O-SVGP gradient updates per timestep in the context of UCI regression. Notably we see that the value we chose for our comparison in the main text ($k=1$) performs well in comparison to larger values of $k$. In contrast to the results in Figure \ref{supp:fig:osvgp_steps}, there is relatively little benefit to increasing the number of gradient update steps per batch when the batch size is very small (in our case, 1).}
    \label{supp:fig:num_update_steps_ablation}
\end{figure}
\subsection{A Closer Look at O-SVGP}

In this paper, we focus primarily on the online SVGP objective of \citet{bui_streaming_2017}, ignoring for the moment their $\alpha$-divergence objective that is used in some of their models --- which can itself be viewed as a type of generalized variational inference \citep{knoblauch2019generalized}.

Recalling the online uncollapsed bound of \citet{bui_streaming_2017} and adapting their notation --- copying directly from their appendix, the objective becomes 
\begin{align}
	\mathcal{F}\left(q_{\text {new }}(f)\right) &= \int  \mathrm{d} f q_{\text {new }}(f)\left[\log \frac{p\left(\mathbf{a} | \theta_{\text {old }}\right) q_{\text {new }}(\mathbf{b})}{p\left(\mathbf{b} | \theta_{\text {new }}\right) q_{\text {old }}(\mathbf{a}) p\left(\mathbf{y}_{\text {new }} | f\right)}\right]  
	\label{supp:eq:osvgp} \\
	= -\mathbb{E}_{q_{\text {new }}(f)}( \log &p\left(\mathbf{y}_{\text {new }} | f\right)) + \mathrm{KL}\left(q(\mathbf{b}) \| p\left(\mathbf{b} | \theta_{\text {new }}\right)\right)  +\mathrm{KL}\left(q_{\text {new }}(\mathbf{a}) \| q_{\text {old }}(\vec a)\right)-\mathrm{KL}\left(q_{\text {new }}(\mathbf{a}) \| p\left(\mathbf{a} | \theta_{\text {old }}\right)\right),
	\nonumber
\end{align}
where $\vec a$ is the old set of inducing points, $\vec b$ is the new set of inducing points, $q(.)$ is the variational posterior on a set of points, $\theta_{new}$ are the current hyperparameters to the GP, and $\theta_{old}$ are the hyperparameters for the GP at the previous iteration.
In Eq. \ref{supp:eq:osvgp}, the first two terms are the standard SVI-GP objective (e.g from \citep{hensman2013bigdatagp}), while the second two terms add to the standard objective allowing SVI to be applied to the streaming setting.
Mini-batching can be achieved without knowing the number of data points a priori; however, this achievement comes at the expense of having to compute two new terms in the loss.

Since the bound in Eq. \ref{supp:eq:osvgp} is uncollapsed, it must be optimized to a global maximum at every timestep to ensure both the GP hyperparameters and the variational parameters are at their optimal values. As noted in \citet{bui_streaming_2017}, this optimization is extremely difficult in the streaming setting for two main reasons. 1) Observations may arrive in a non-iid fashion and violate the assumptions of SVI, and 2) each observation batch is seen once and discarded, preventing multiple passes through the full dataset, as is standard practice for SVI. Even in the batch setting, optimizing the SVI objective to a global maximum is notoriously difficult due to the proliferation of local maxima. This property makes a fair timing comparison with our approach somewhat difficult in that multiple gradient steps per data point will necessarily be slower than WISKI, which does not have any variational parameters to optimize, and thus can learn with fewer optimization steps per observation.
We implemented Eq \ref{supp:eq:osvgp} in GPyTorch \citep{gardner2018gpytorch}, but additionally attempted to use the authors' provided implementation of O-SVGP\footnote{\url{https://github.com/thangbui/streaming_sparse_gp}} finding similar results --- many gradient steps are required to reduce the loss.
As a demonstration, we varied the number of steps of optimization per batch in Figure \ref{supp:fig:osvgp_steps} with a large batch size $300$ (the same as \citet{bui_streaming_2017}'s own experiments) in the online regression setting on synthetic sinusoidal data, finding that at least $10$ optimization steps were needed to decrease the RMSE in a reasonable manner even on this simpler problem.

\begin{figure}[t]
	\centering
	\begin{subfigure}{0.23\textwidth}
		\includegraphics[width=\linewidth,clip,clip,trim=0cm 5cm 0cm 5cm]{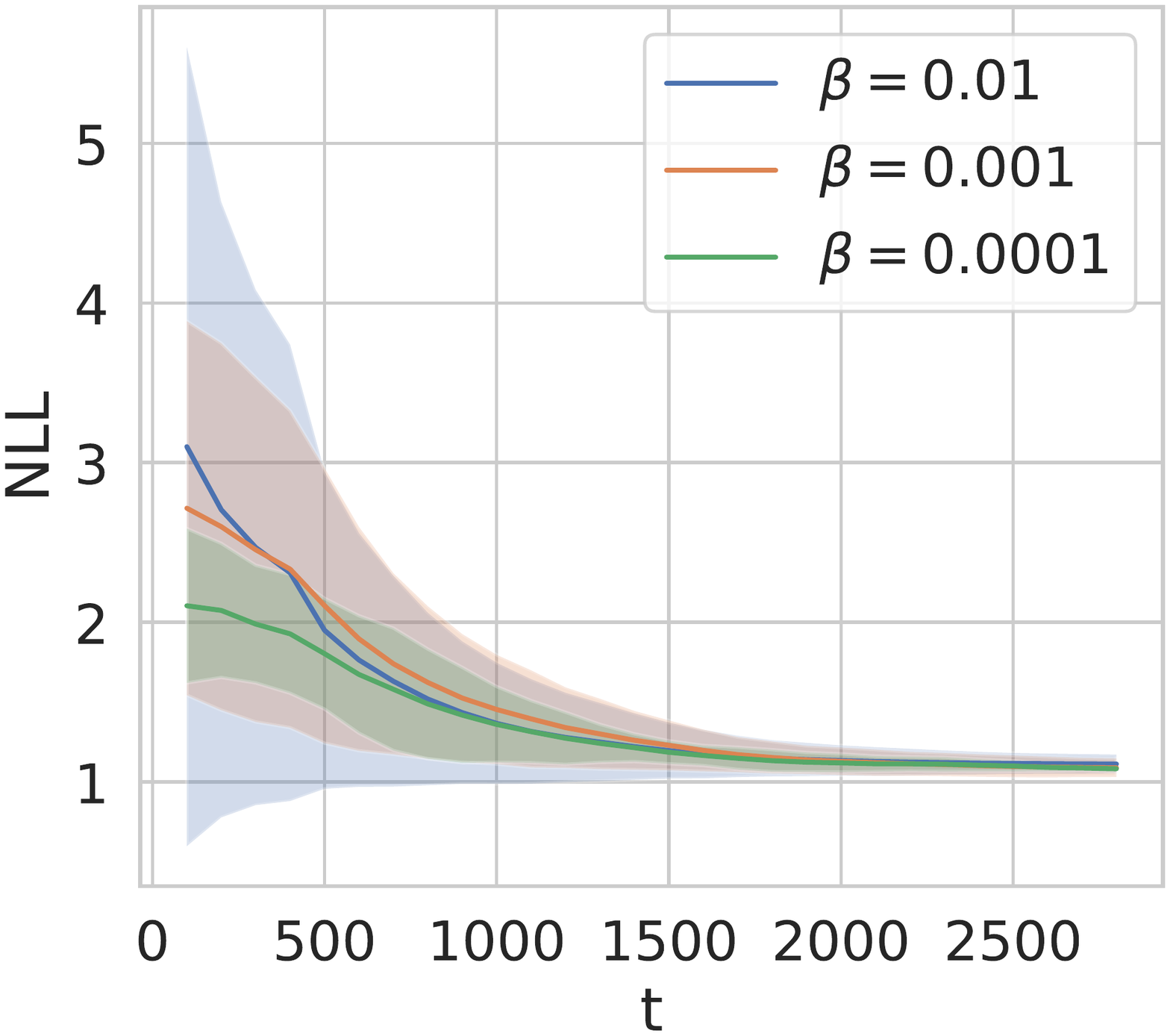}
		\caption{Skillcraft}
		\label{supp:fig:beta_ablation_skillcraft}   
	\end{subfigure}
	\hfill
	\begin{subfigure}{0.23\textwidth}
		\includegraphics[width=\linewidth,clip,clip,trim=0cm 5cm 0cm 5cm]{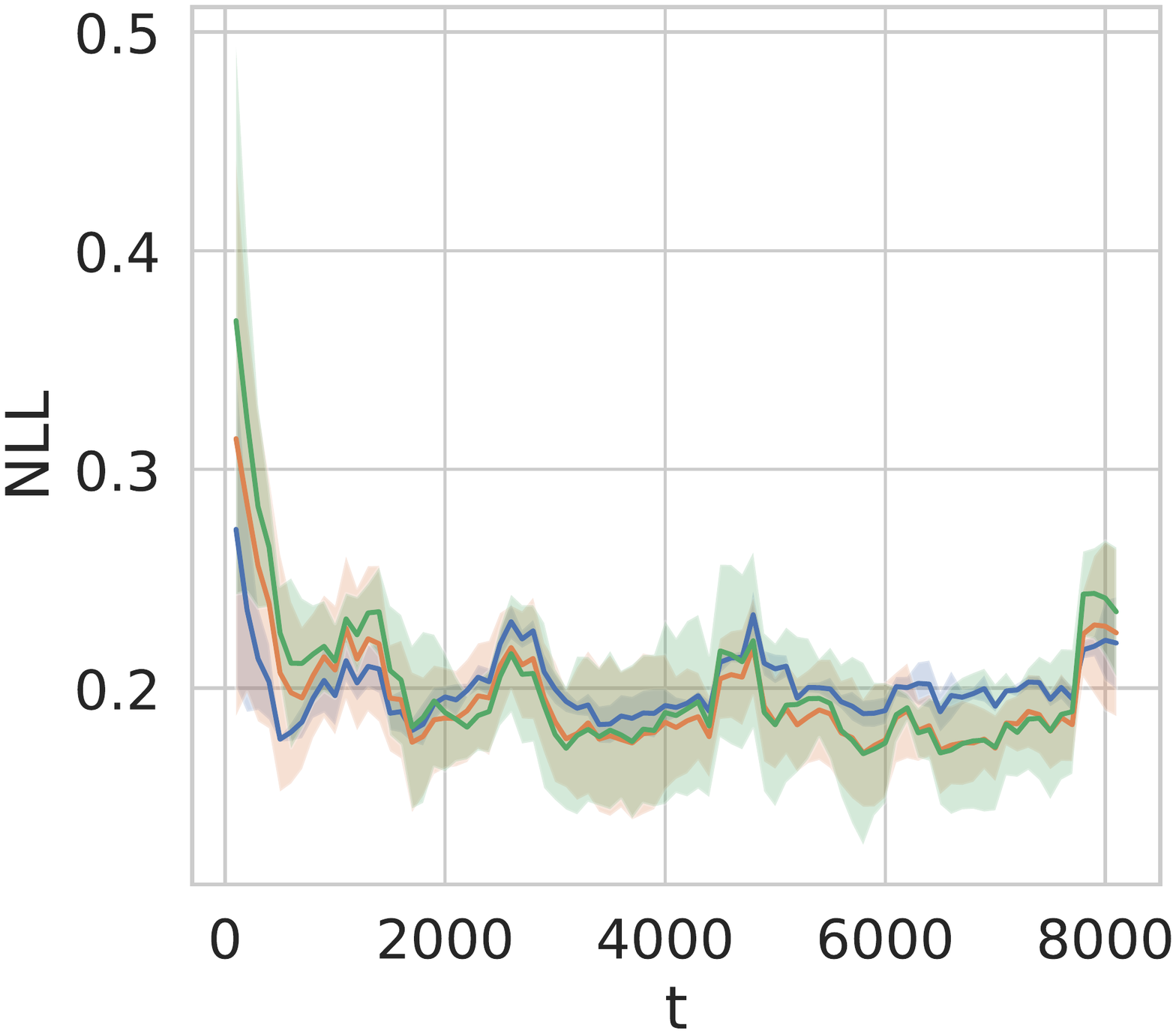}
		\caption{Powerplant}
		\label{supp:fig:beta_ablation_powerplant}   
	\end{subfigure}
	\hfill
	\begin{subfigure}{0.23\textwidth}
		\centering
		\includegraphics[width=\textwidth,clip,clip,trim=0cm 5cm 0cm 5cm]{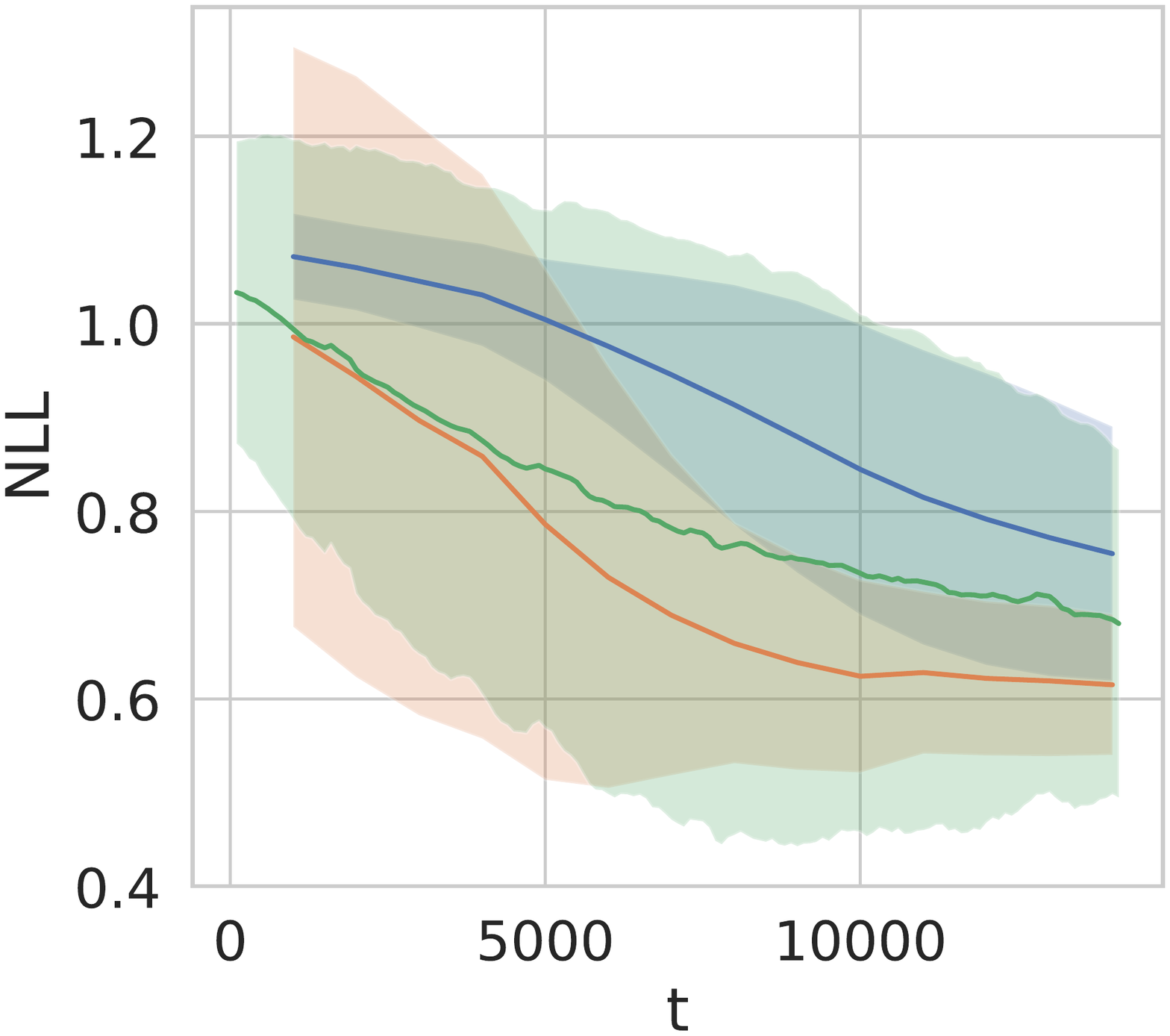}
		\caption{Elevators}
		\label{supp:fig:beta_ablation_elevators}
	\end{subfigure}
	\hfill
	\begin{subfigure}{0.23\textwidth}
		\centering
		\includegraphics[width=\textwidth,clip,clip,trim=0cm 5cm 0cm 5cm]{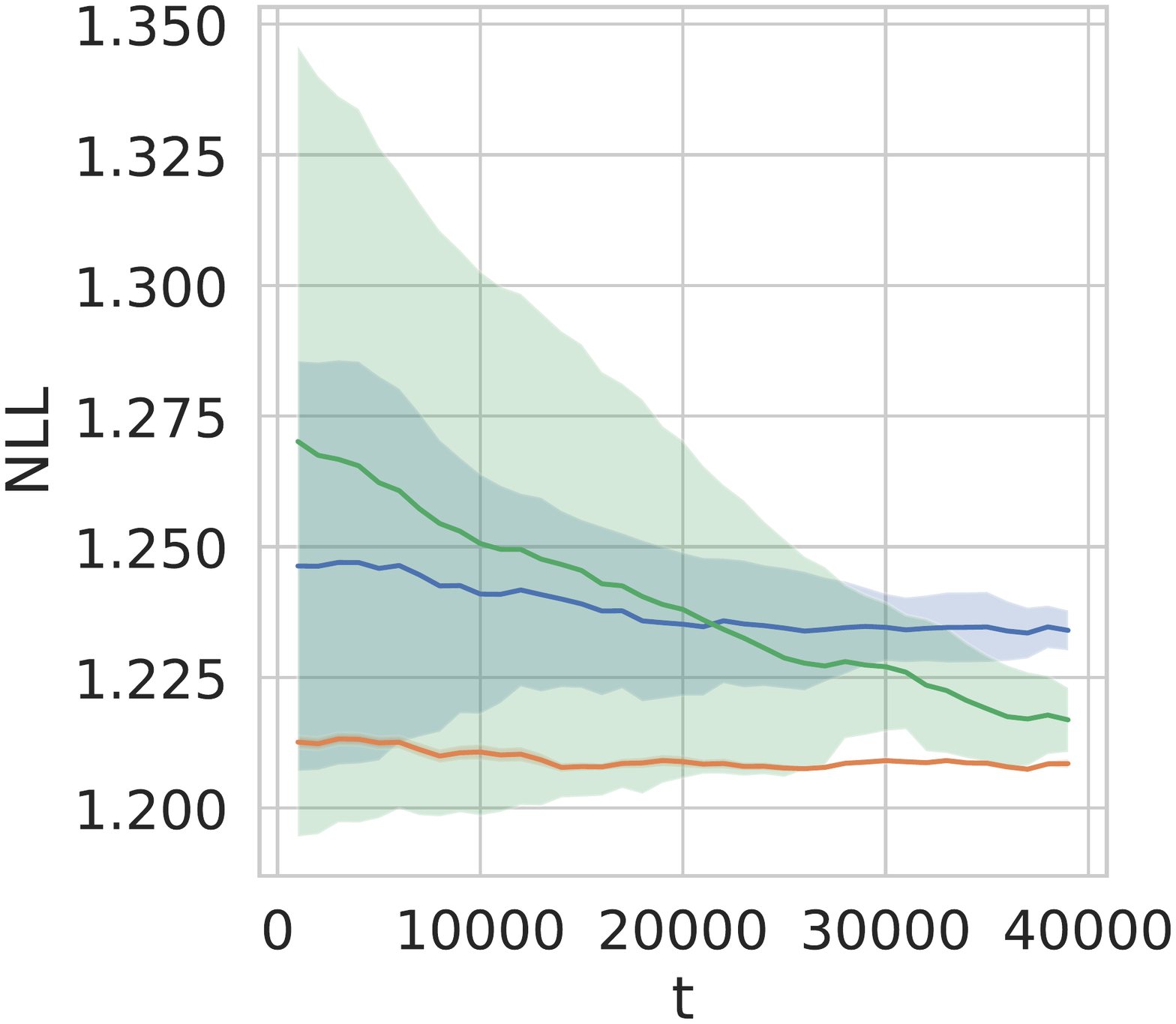}
		\caption{Protein}
		\label{supp:fig:beta_ablation_protein}
	\end{subfigure}
	\caption{Here we ablate the $\beta$ hyperparameter in the GVI loss for O-SVGP on the UCI datasets considered in this paper. While there is not a clear winner, we find that $\beta =$ 1e-3 works well for all datasets. }
	\label{supp:fig:svgp_beta_ablation}
\end{figure}

To remedy the convergence issue (and to create a fair comparison with \emph{one} gradient step per batch of data), we down-weighted the KL divergence terms, producing a generalized variational objective (equivalent to taking the likelihood to a power $1/\beta$) \citep{knoblauch2019generalized}.
Eq \ref{supp:eq:osvgp} loss becomes:
\begin{align}
	\mathcal{F}
	&= -\mathbb{E}_{q_{\text {new }}(f)}( \log p\left(\mathbf{y}_{\text {new }} | f\right)) + \beta \mathrm{KL}\left(q(\mathbf{b}) \| p\left(\mathbf{b} | \theta_{\text {new }}\right)\right) \nonumber \\
	& \hspace{4mm} +\beta \mathrm{KL}\left(q_{\text {new }}(\mathbf{a}) \| q_{\text {old }}(a)\right)- \beta \mathrm{KL}\left(q_{\text {new }}(\mathbf{a}) \| p\left(\mathbf{a} | \theta_{\text {old }}\right)\right), \label{supp:eq:osvgp_generalized}
\end{align}
where all terms are as before except for $\beta < 1$.
We found that $\beta << 1$ produces more reasonable results for a batch size of $1.$
Ablations for varying this hyperparameter are shown in Figure \ref{supp:fig:svgp_beta_ablation}.
Using generalized variational inference does not change the complexity of the streaming objective, which remains $\mathcal{O}(Bm^2 + m^3);$ the $\mathcal{O}(m^3)$ term stays the same due to the log determinant term in the KL objective.

Finally, we vary the number of inducing points in the O-SVGP bound in Figure \ref{supp:fig:num_inducing_ablation}, finding that O-SVGP is quite sensitive to the number of inducing points.

\section{EXPERIMENTAL DETAILS}
\label{supp:sec:regression}

\subsection{Regression and Classification}

\begin{algorithm}
	\SetAlgoLined
	\caption{Online Learning with WISKI \label{supp:alg:online_learning}}
	\KwIn{Kernel function $k$, inducing grid $U$, initial data $\mathbf x_{1:n}, \mathbf y_{n}$, GP parameters $\theta$, feature map parameters $\phi$, learning rate $\eta$.}
	Initialize $K_{UU}, \cache{L}, \cache{W^\top \mathbf{y}_{n}}, \cache{\mathbf y_{n}^\top \mathbf y_{n}}$. \\
	\For{$t=n + 1, n + 2, \dots$}{
		Receive $\mathbf x_t$. \\
		Predict $\hat p(y_t | \mathbf x_{1:t}, \mathbf y_{t-1}, \theta, \phi)$ (Eq. \ref{main:eq:woodbury_predictive_mean}). \\
		Observe $y_t$, compute $\vec w_t$. \\
		Update caches $\cache{L_t}, \cache{W_t^\top \vec y_t}, \cache{\vec y_t^\top \mathbf y_t}$. \\
		$\phi \leftarrow \phi - \eta \nabla_\phi \mathcal{L}(\phi)$ (Eq. \ref{main:eq:partial_mll}). \\
		$\theta \leftarrow \theta - \eta \nabla_\theta \mathcal{L}(\theta)$ (Eq. \ref{main:eq:woodbury_mll}). \\
	}
\end{algorithm}

\begin{figure*}[t]
	\centering
	\begin{subfigure}{0.23\textwidth}
		\includegraphics[width=\linewidth,clip,clip,trim=0cm 5cm 0cm 5cm]{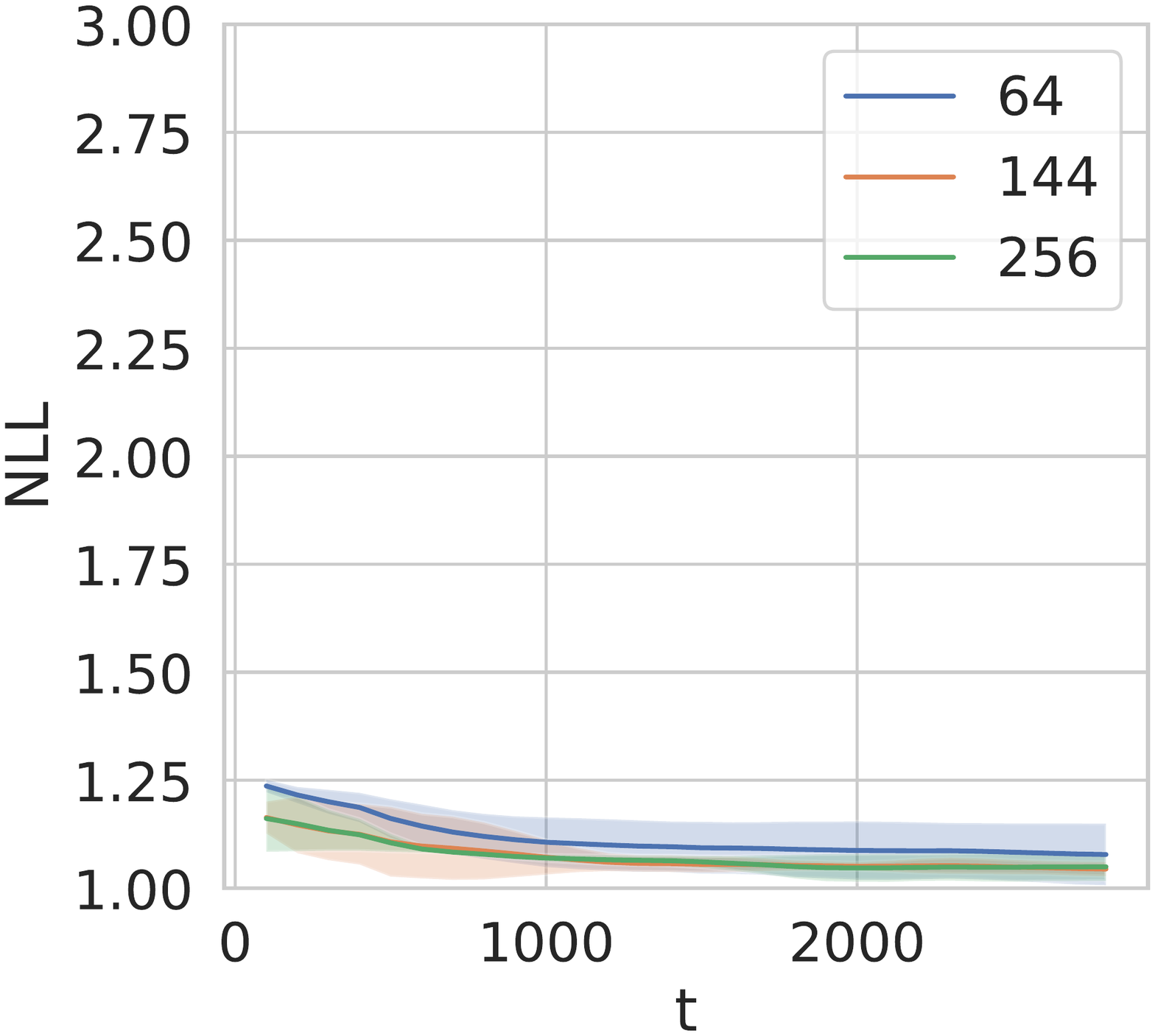}
		\caption{WISKI, Skillcraft}
		\label{supp:fig:wiski_skillcraft_num_inducing_ablation}   
	\end{subfigure}
	\hfill
	\begin{subfigure}{0.23\textwidth}
		\includegraphics[width=\linewidth,clip,clip,trim=0cm 5cm 0cm 5cm]{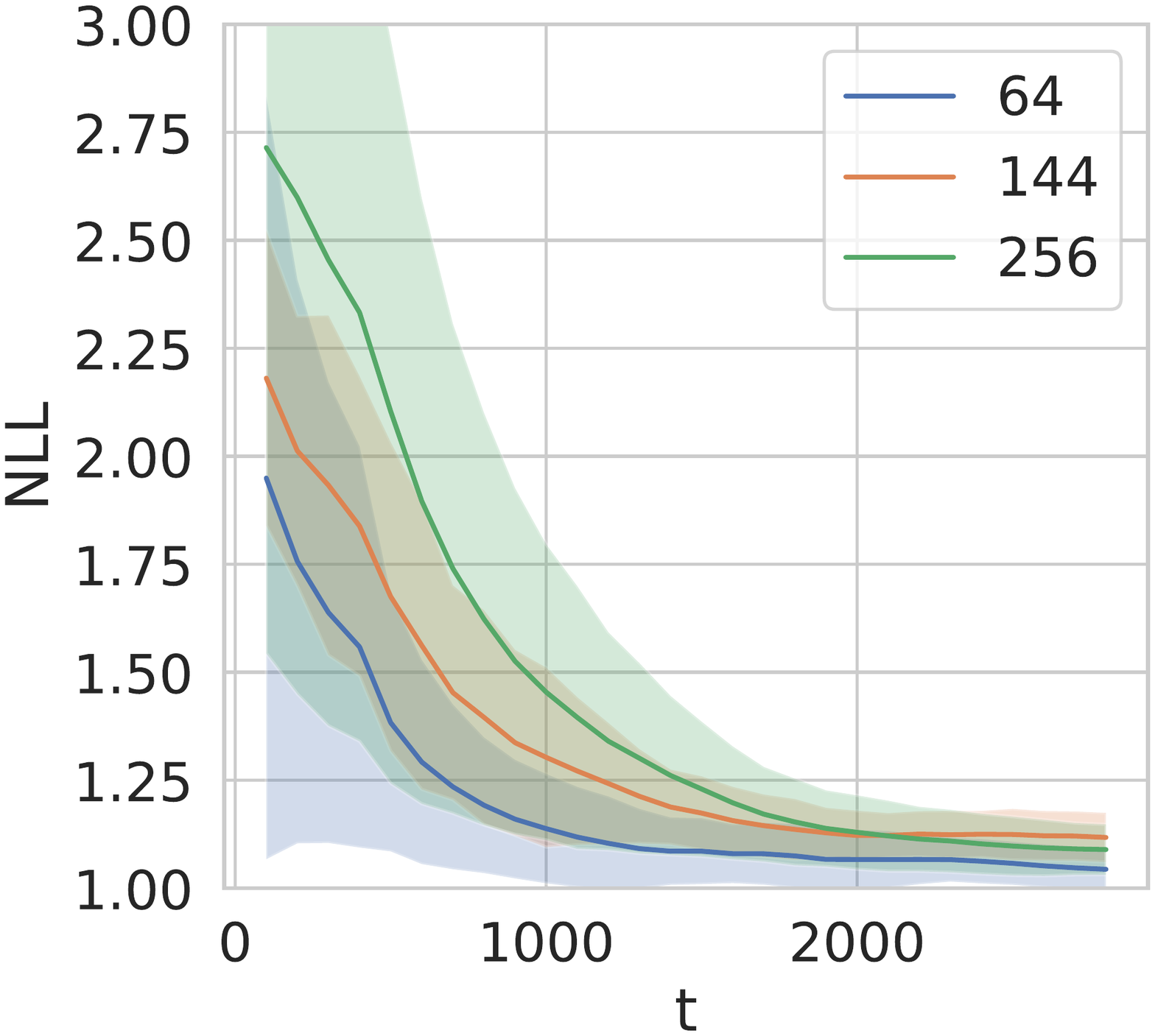}
		\caption{O-SVGP, Skillcraft}
		\label{supp:fig:svgp_skillcraft_num_inducing_ablation}   
	\end{subfigure}
	\hfill
	\begin{subfigure}{0.23\textwidth}
		\centering
		\includegraphics[width=\textwidth,clip,clip,trim=0cm 5cm 0cm 5cm]{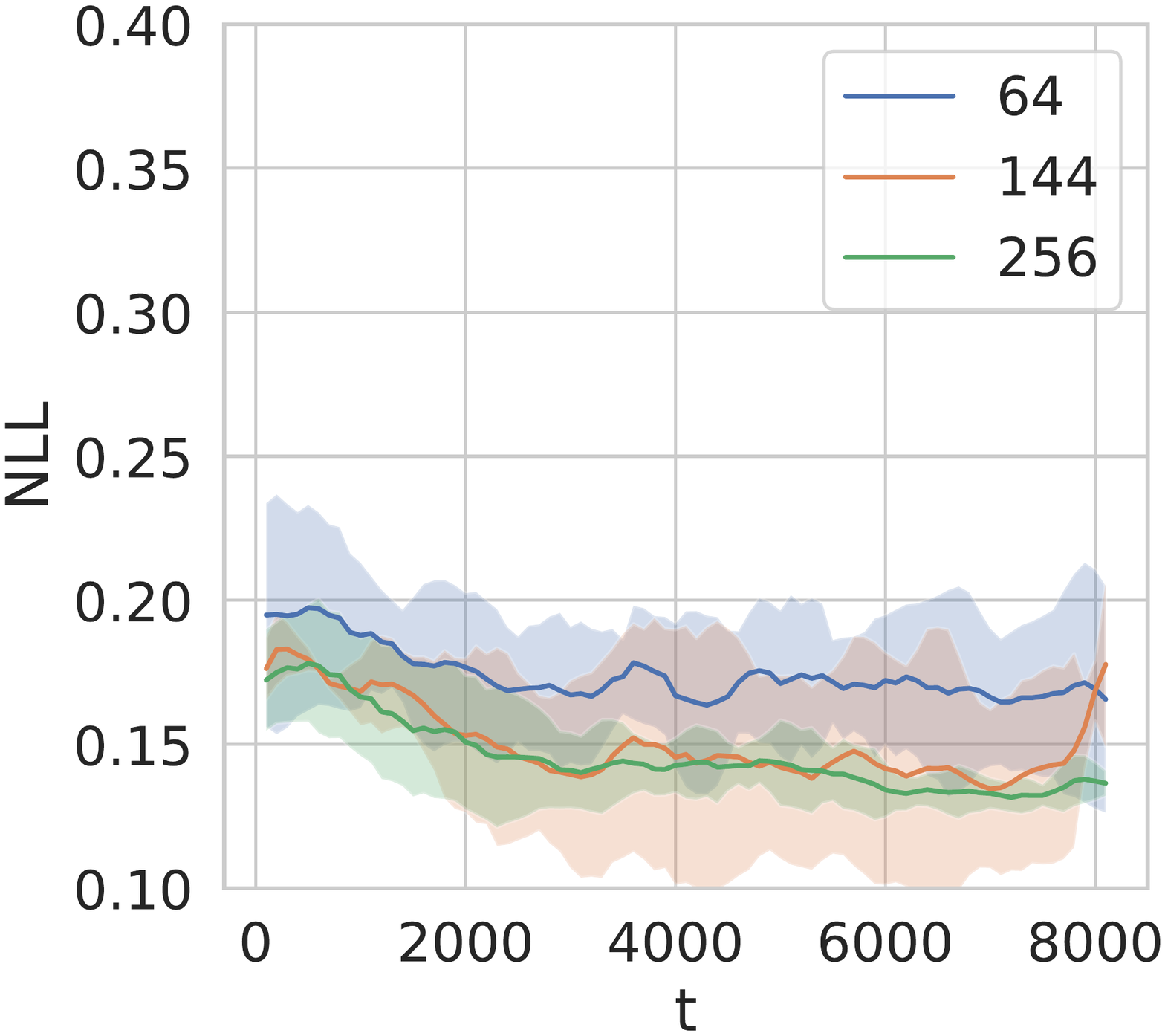}
		\caption{WISKI, Powerplant}
		\label{supp:fig:wiski_powerplant_num_inducing_ablation}
	\end{subfigure}
	\hfill
	\begin{subfigure}{0.23\textwidth}
		\centering
		\includegraphics[width=\textwidth,clip,clip,trim=0cm 5cm 0cm 5cm]{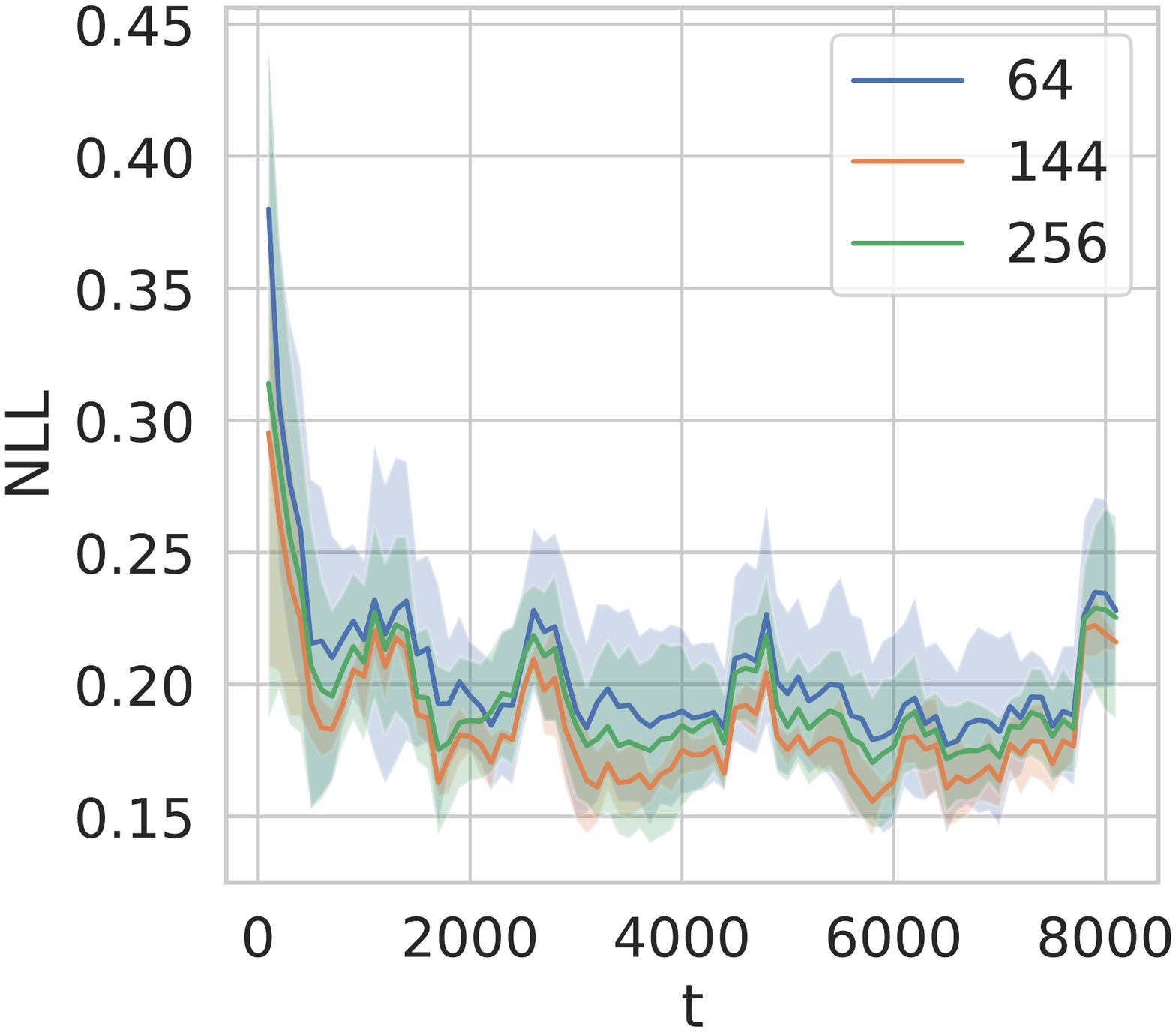}
		\caption{O-SVGP, Powerplant}
		\label{supp:fig:svgp_powerplant_num_inducing_ablation}
	\end{subfigure}
	\caption{Here we ablate the number of inducing points for both WISKI and O-SVGP. We find that
		WISKI is not very sensitive to the number of inducing points, but always improves if more inducing points are added. O-SVGP sometimes performs better with fewer inducing points, a phenomenon we attribute to either 1) poor optimization of the GVI objective or 2) overfitting due to the downweighted KL terms in the GVI objective. In theory adding inducing points should only improve the performance of an SVGP. This observation highlights the difficulties O-SVGP often encounters in practice.}
	\label{supp:fig:num_inducing_ablation}
\end{figure*}

\begin{figure}[t]
	\begin{subfigure}{0.48\textwidth}
		\centering
		\includegraphics[width=\textwidth,clip,clip,trim=0cm 10cm 0cm 10cm]{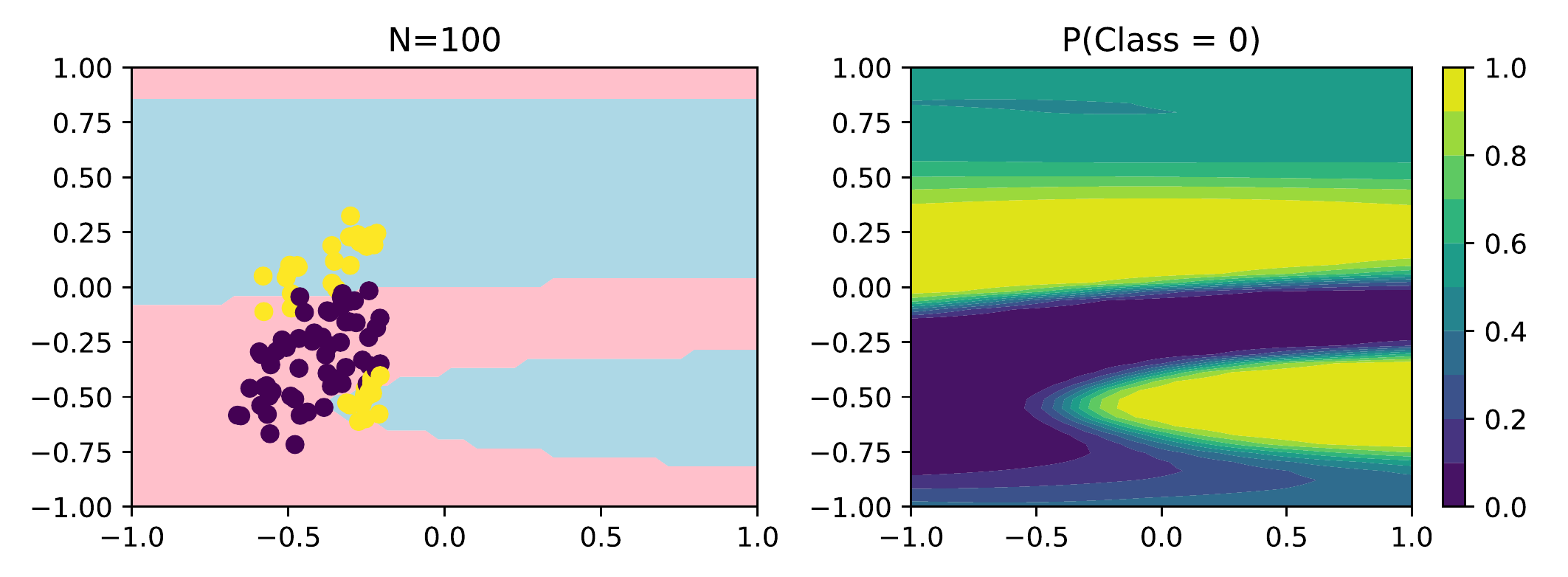}
		\caption{Test Accuracy - 70\%}
		\label{supp:fig:banana_viz_100}
	\end{subfigure}
	\hfill
	\begin{subfigure}{0.48\textwidth}
		\centering
		\includegraphics[width=\textwidth,clip,clip,trim=0cm 10cm 0cm 10cm]{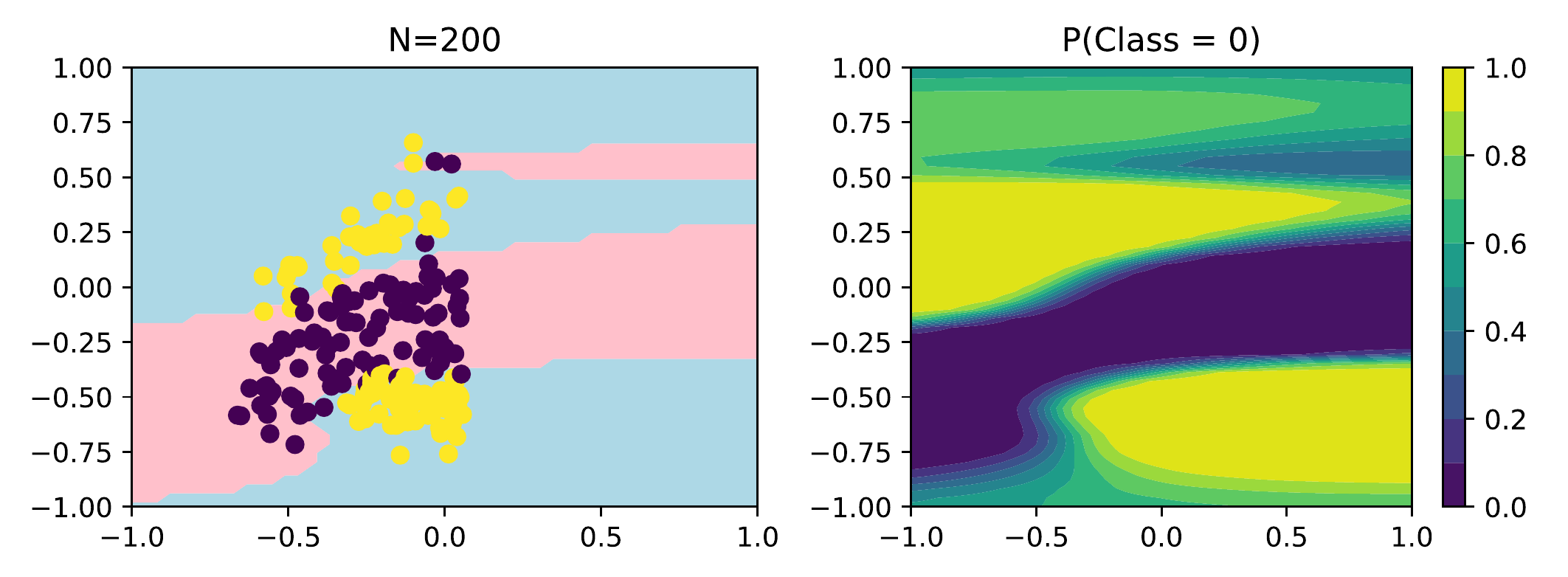}
		\caption{Test Accuracy - 70\%}
		\label{supp:fig:banana_viz_200}
	\end{subfigure}
	\hfill
	\begin{subfigure}{0.48\textwidth}
		\centering
		\includegraphics[width=\textwidth,clip,clip,trim=0cm 10cm 0cm 10cm]{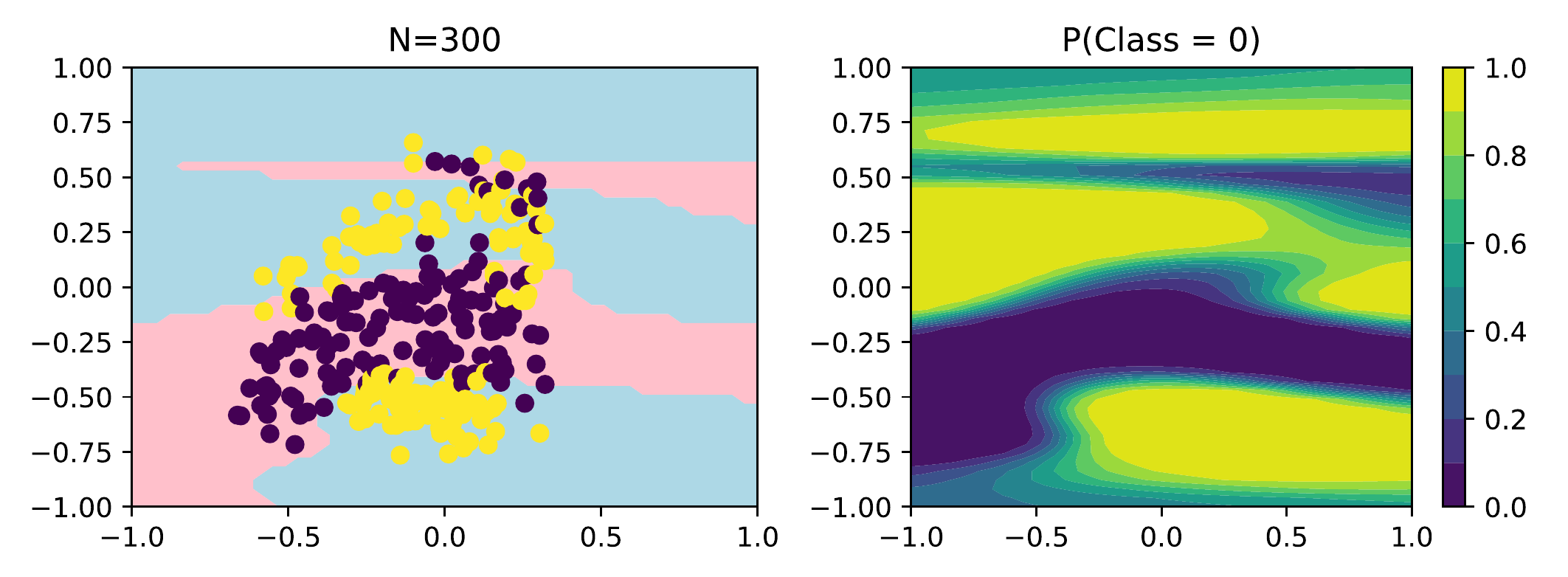}
		\caption{Test Accuracy - 77\%}
		\label{supp:fig:banana_viz_300}
	\end{subfigure}
	\hfill
	\begin{subfigure}{0.48\textwidth}
		\centering
		\includegraphics[width=\textwidth,clip,clip,trim=0cm 10cm 0cm 10cm]{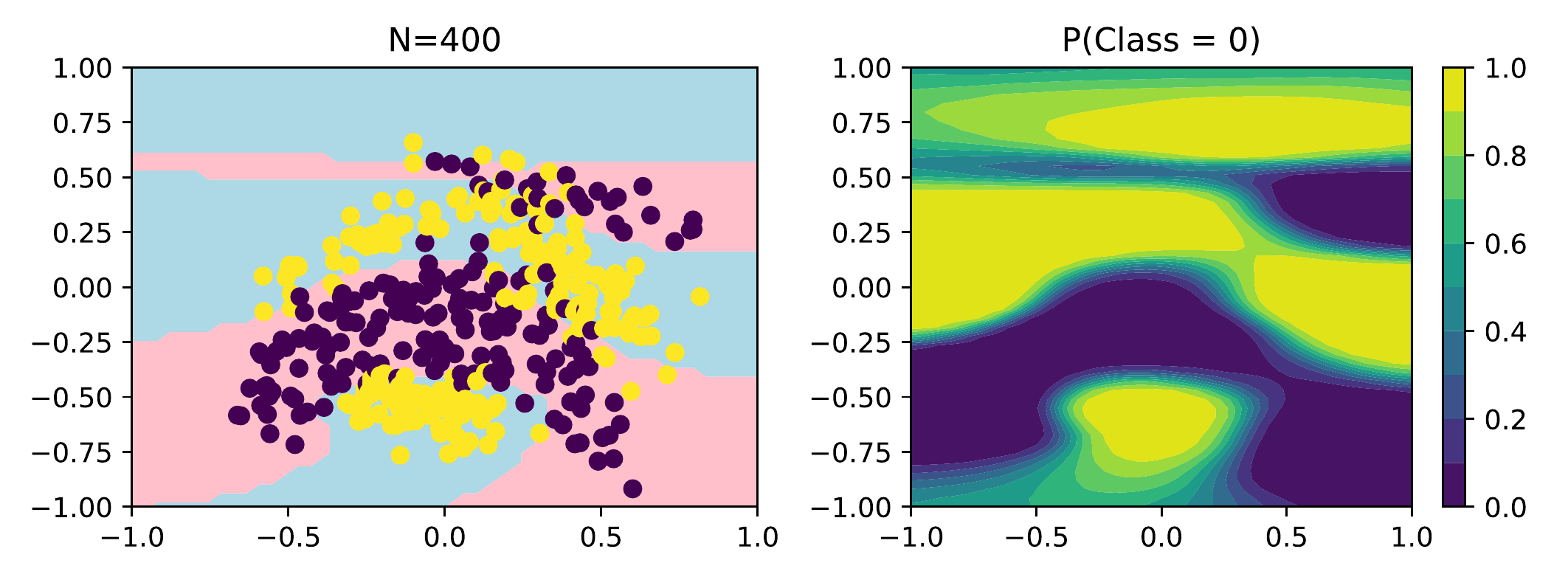}
		\caption{Test Accuracy - 88\%}
		\label{supp:fig:banana_viz_400}
	\end{subfigure}
	\caption{Online Gaussian Process Dirichlet classification with WISKI on observations from the banana dataset arriving in non-i.i.d. fashion (shown). The WISKI classifier is updated with a single gradient step after each individual observation.}
	\label{supp:fig:banana_viz}
\end{figure}

Algorithm \ref{supp:alg:online_learning} summarizes online learning with WISKI. If an input projection is not learned, then $h(\vec x; \phi)$ can be taken to be the identity map, and the projection parameter update is consequently a no-op. In the rest of this section we provide additional experimental results in the regression and classification setting.  In Figure \ref{supp:fig:regression_rmse_comparison} we report the RMSE for each of the UCI regression tasks. Note that the RMSE is computed on the standardized labels. The qualitative behavior is identical to that of the NLL plots in the main text (Figure \ref{main:fig:regression_nll_comparison}). Figure \ref{supp:fig:banana_viz} is a visualization of a WISKI classifier on non-i.i.d. data. Figures \ref{supp:fig:num_inducing_ablation} and \ref{supp:fig:svgp_beta_ablation} report the results of our ablations on $m$ and $\beta$, respectively. This section provides all necessary implementation details to reproduce our results.

\paragraph{Data Preparation}
For all datasets, we scaled input data to lie in $[-1, 1]^d$. For regression datasets we standardized the targets to have zero mean and unit variance. If the raw dataset did not have a train/test split, we randomly selected 10\% of the observations to form a test dataset. From the remainding 90\% we removed an additional 5\% of the observations for pretraining.

\paragraph{Hyperparameters}
We pretrained all models for $T_{\mathrm{batch}}$ epochs, with learning rates $\eta_{\mathrm{batch}}$. If we learned a projection of the inputs, we used a lower learning rate for the projection parameters. While a small learning rate worked well for all tasks, for the best performance we used a higher learning rate for easier tasks (Table \ref{supp:table:hyperparameters}).

\begin{center}
	\begin{tabular}{|c|c|c|c|c|c|c|c|} \label{supp:table:hyperparameters}
		Task & $m$ & $T_{\mathrm{batch}}$ & $\eta_{\mathrm{batch}}(\theta)$ & $\eta_{\mathrm{batch}}(\phi)$ & $\eta_{\mathrm{online}}(\theta)$ & $\eta_{\mathrm{online}}(\phi)$ & $\beta$ \\
		\hline Banana & 256 & 200 & 5e-2 & - & 5e-3 & - & 1e-3 \\
		\hline SVM Guide 1 & 256 & 200 & 5e-2 & - & 5e-3 & - & 1e-3 \\
		\hline Skillcraft & 256 & 200 & 5e-2 & 5e-3 & 5e-3 & 5e-4 & 1e-3 \\
		\hline Powerplant & 256 & 200 & 5e-2 & 5e-3 & 5e-3 & 5e-4 & 1e-3 \\
		\hline Elevators & 256 & 200 & 1e-2 & 1e-3 & 1e-3 & 1e-4 & 1e-3 \\
		\hline Protein & 256 & 200 & 1e-2 & 1e-3 & 1e-3 & 1e-4 & 1e-3 \\
		\hline 3DRoad & 1600 & 800 & 1e-2 & - & 1e-3 & - & 1e-3 \\
		\hline
	\end{tabular}
\end{center}

\begin{table}
	\centering
	\begin{tabular}{l c c c c c c}
		\toprule
		$m$  & $r$ & NLL \\
		\hline 256 & 128 & $8.2\mathrm{e}{+6} \pm 9.8\mathrm{e}{+6}$  \\
		\hline 256 & 192 & $1.000 \pm 0.010$ \\
		\hline 256 & 256 & $1.007 \pm 0.015$ \\
		\hline 1024 & 256 & $2.9\mathrm{e}{+7}\pm 9.2{e}{+7}$ \\
		\hline 1024 & 512 & $1.050 \pm 0.082$ \\
		\hline 1024 & 768 & $0.995 \pm 0.007$ \\
		\hline 1024 & 1024 & $1.007 \pm 0.008$ \\
		\bottomrule
	\end{tabular}
	\caption{Root rank ($r$) ablation by NLL across both $m=256$ and $m=1024$ inducing points on skillcraft. Too small of a rank fails to converge; however, once $r$ is large enough (about $m/2$), the performance is unchanged.}
	\label{tab:k_ablation}
\end{table}

\subsection{Bayesian Optimization Experimental Details and Further Results}\label{supp:bo}
For the Bayesian optimization experiments, we considered noisy three dimensional versions of the Bayesian optimization test functions available from BoTorch\footnote{\url{https://botorch.org/api/test_functions.html}}. We sed the BoTorch implementation of the test functions with the \textit{qUCB} acquisition function with $q = 3,$ randomly choosing five points to initialize with and running $1500$ BO steps, so that we end up with $4505$ data points acquired from the models.
We then followed BoTorch standard optimization of the acquisition functions by optimizing with LBFGS-B with $10$ random restarts, $512$ samples to initialize the optimization with, a batch limit of $5$ and $200$ iterations of LBFGS-B.
We fit the model to convergence at each iteration as model fits are very important in BO using LBFGS-B for exact and WISKI while using Adam for OSVGP because the variational parameters are much higher dimensional so LBFGS-B is prohibitively slow.
The timing results take into account the model re-fitting stage, the acquisition optimization stage, and the expense of adding a new datapoint into the model.
We used a single AWS instance with eight Nvidia Tesla V100s for these experiments, running each experiment four times, except for StyblinskiTang, which we ran three times (as the exact GP ran out of memory during a bayes opt step on one of the seeds).
We measure time per iteration by adding both the model fitting time and the acquisition function optimization time.
While a single training step is somewhat faster for O-SVGP than for WISKI, we found that it tended to take longer to optimize acquisition functions in BO loops.
  
\begin{figure*}[h!]
	\includegraphics[angle=90,width=\textwidth,clip,clip,trim=4cm 0cm 4cm 0cm]{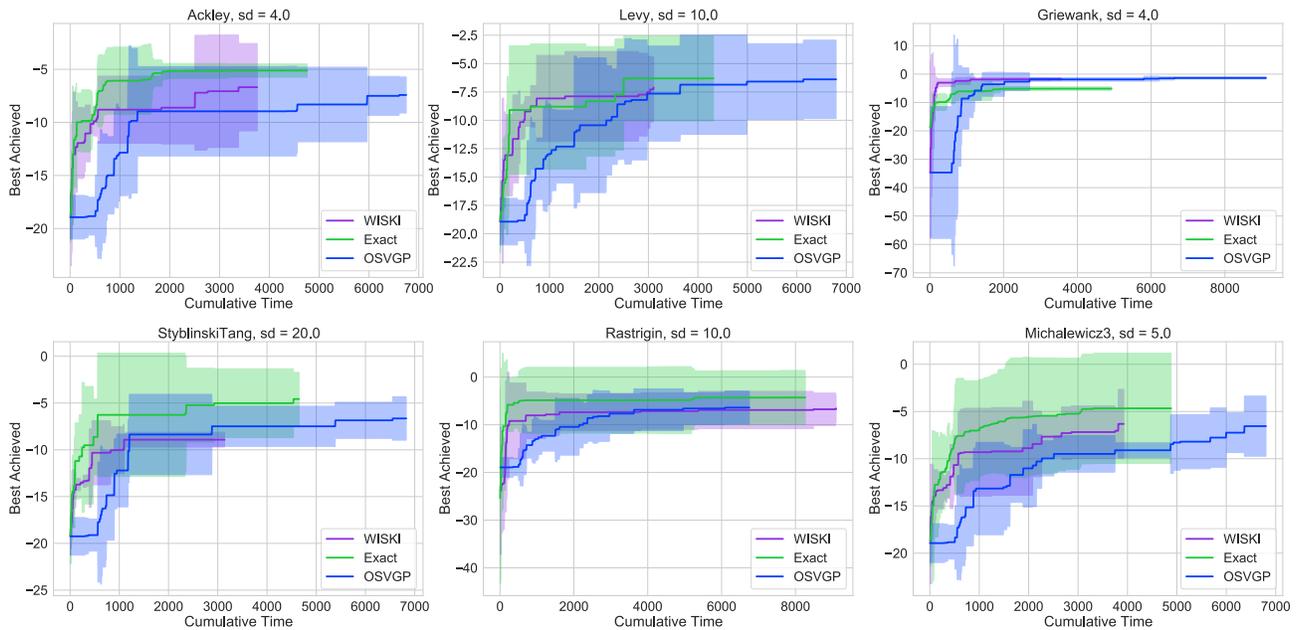}
	\caption{Bayesian optimization results in terms of total optimization time. Throughout, WISKI is generally the fastest, except on Griewank, while reaching similar optimization performance to the exact GPs across the board. WISKI is somewhat better but significantly faster than the other methods on Griewank, but with similar performance to O-SVGP on StyblinskiTang and Michalewicz.} \label{supp:fig:full_bo}
\end{figure*}

\begin{figure*}[ht]
	\includegraphics[angle=90,width=\textwidth,clip,clip,trim=4cm 0cm 4cm 0cm]{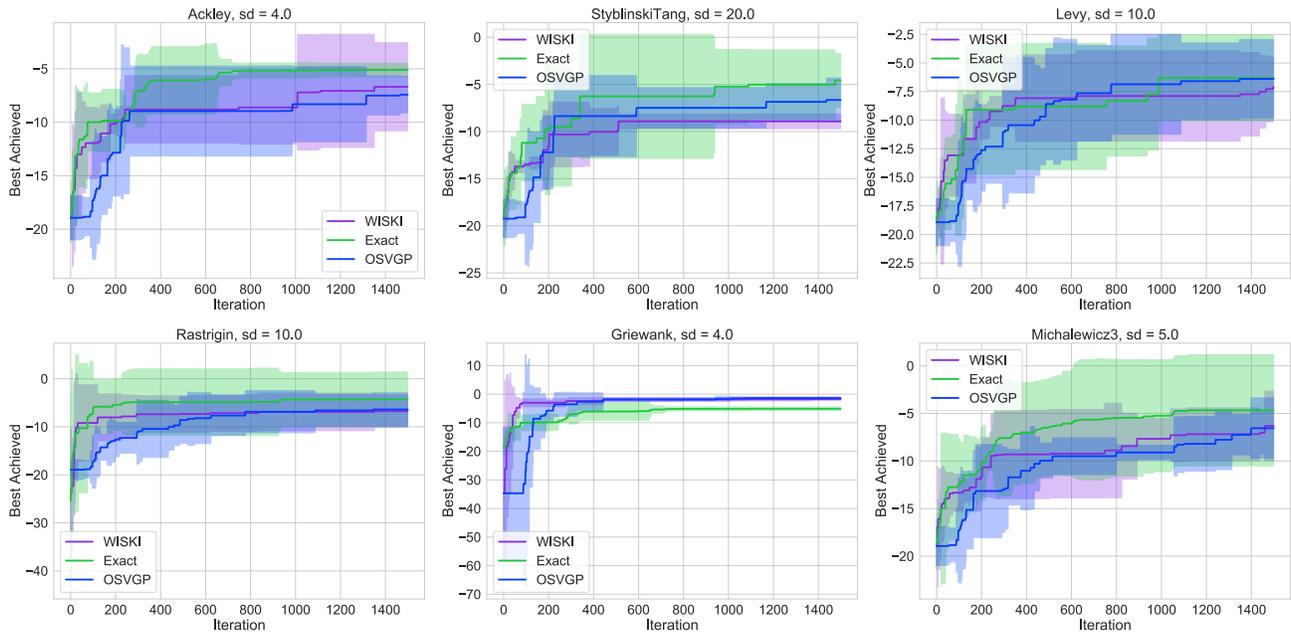}
	\caption{Bayesian optimization results in terms of iteration complexity for noisy $3D$ test functions. Throughout, WISKI performs comparably to the exact GP.} \label{supp:fig:iteration_complexity}
\end{figure*}

\begin{figure*}[h!]
	\includegraphics[angle=90,width=\textwidth,clip,clip,trim=4cm 0cm 4cm 0cm]{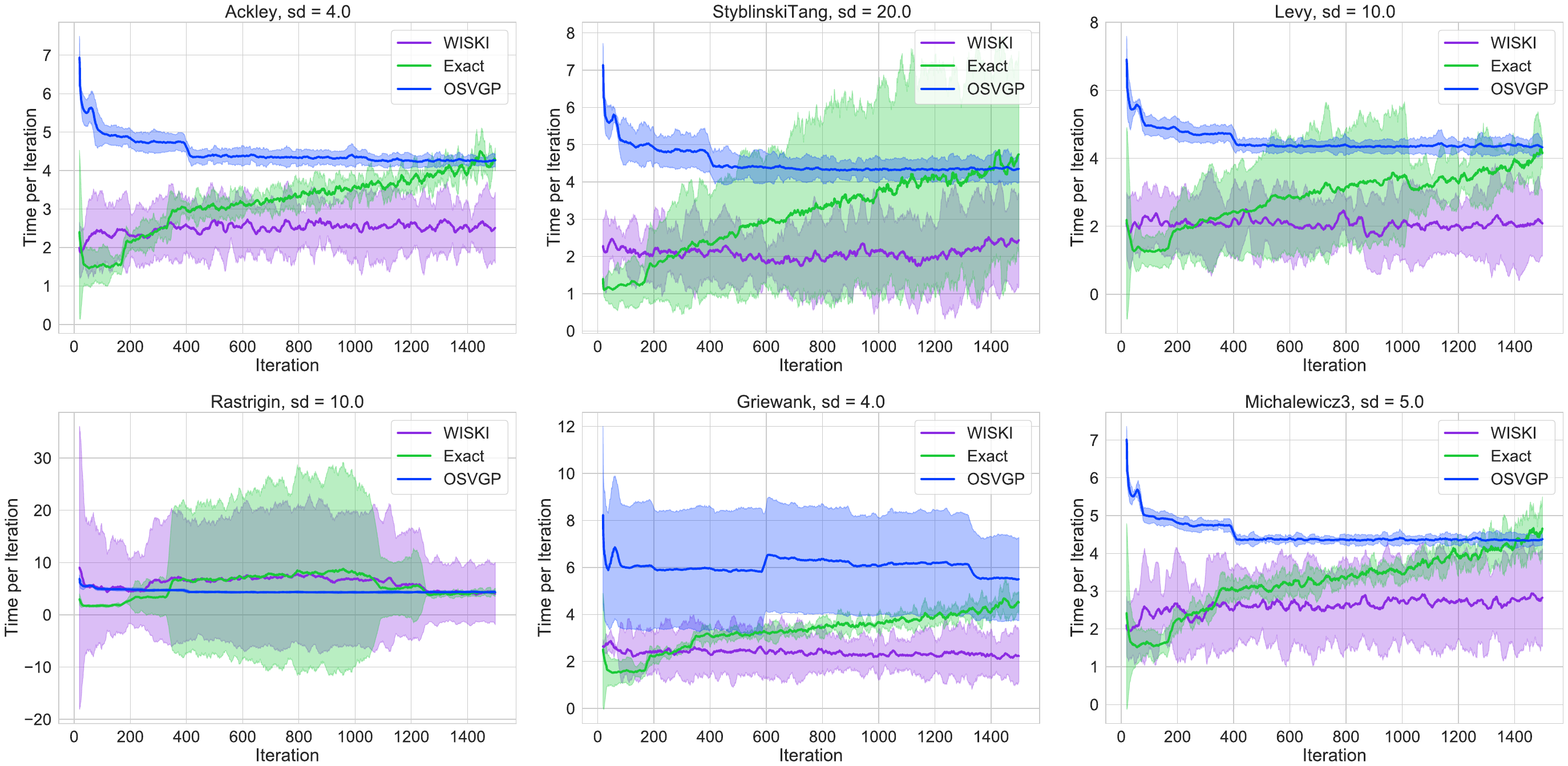}
	\caption{Bayesian optimization results in terms of time complexity for the noisy $3D$ test functions. WISKI is the fastest method on the problems, while the exact GPs increase time per iteration at least linearly (they use LOVE predictive variances internally). While OSVGP is constant time, it typically is somewhat slower due to larger constants with respect to $n,$ $m^3$ versus $m^2$ for WISKI.} \label{supp:fig:time_per_iteration}\vspace{-0.5cm}
\end{figure*}

Results over time per iteration and maximum achieved value by time for the rest of the test suite are shown in Figure \ref{supp:fig:full_bo}.
Overall WISKI performs comparably in terms of maximum achieved value to the exact GP reaches that value in terms of quicker wallclock time.
In Figure \ref{supp:fig:iteration_complexity}, we show the maximum value achieved by iteration for each problem, finding that the exact GPs typically converge to their optimum first, while WISKI converges afterwards with OSVGP slightly after that.
In Figure \ref{supp:fig:time_per_iteration}, we show the time per iteration for all three methods, finding that WISKI is constant time throughout as is OSVGP, while the exact approach scales broadly quadratically (as expected given that the BoTorch default for sampling uses LOVE predictive variances and sampling).
Digging deeper into the results, we found that the speed difference between OSVGP and WISKI is attributable to the increased predictive variance and sampling speed for OSVGP ($\mathcal{O}(m^3)$ compared to $\mathcal{O}(m^2)$).

\begin{table}[h!]
	\centering
\begin{tabular}{| c | c | c| c| c| c|}
	\hline Levy & Ackley & StyblinskiTang & Rastrigin & Griewank & Michalewicz \\ \hline
	$10.0$ & $4.0$ & $20.0$ & $10.0$ & $4.0$ & $5.0$ \\ \hline
\end{tabular}
	\caption{Noise standard deviations used for the Bayesian optimization experiments.}
\label{tab:problem_sd}
\end{table}

\subsection{Active Learning Experimental Details}
For the active learning problem, we used a batch size of $6$ for all models, with a base kernel that was a scaled ARD Matern-$0.5$ kernel, and lengthscale priors of Gamma($3$, $6$) and outputscale priors of Gamma($2$, $0.15$).
For both OSVGP and WISKI, we used a grid size of $900$ ($30$ per dimension); for OSVGP, we trained the inducing points, finding fixed inducing points did not reduce the RMSE. 
For O-SVGP, we used $\beta = 0.001$ and a learning rate of $1e-4$ with the Adam optimizer, while for WISKI and the exact GPs, we used a learning rate of $0.1.$
Here, we re-fit the models until the training loss stopped decaying, analogous to the BO experiments.
The dataset can be downloaded at \url{https://wjmaddox.github.io/assets/data/malaria_df.hdf5}.

\end{document}